\documentclass[sigconf]{acmart}

\usepackage{amsmath}
\usepackage{amsfonts}
\usepackage{tcolorbox}
\usepackage{multicol}
\usepackage{algorithm}
\usepackage{algorithmic}
\usepackage{newfloat}
\usepackage{listings}
\usepackage{subcaption}
\usepackage{tabularx}
\usepackage{graphicx}
\usepackage{supertabular}
\usepackage{placeins}
\usepackage{float}
\usepackage{subcaption}
\usepackage{multirow}
\usepackage[font=small,labelfont=bf]{caption}
\usepackage{dblfloatfix}          %

\DeclareCaptionStyle{ruled}{labelfont=normalfont,labelsep=colon,strut=off} 
\floatstyle{ruled}
\newfloat{listing}{tb}{lst}{}
\floatname{listing}{Listing}

\AtBeginDocument{%
  }

\acmConference[Conference acronym 'XX]{}{}{}

\renewcommand\footnotetextcopyrightpermission[1]{} 
\newcounter{tightlistcnt}
\newenvironment{tightitemize}{%
	\begin{list}{\textbullet}{\usecounter{tightlistcnt}%
			\topsep 0in
			\partopsep 0in
			\itemsep 0in
			\parsep 0in
			\leftmargin .75em
			\rightmargin 0in
			\listparindent 0em
			\itemindent 0em
			\labelsep .75em
			\labelwidth 0.75em
		}%
	}{%
	\end{list}%
}

\newenvironment{tightenumerate}{%
	\begin{list}{\arabic{tightlistcnt}.}{\usecounter{tightlistcnt}
			\topsep 0in
			\partopsep 0in
			\itemsep .2em
			\parsep 0in
			\leftmargin 0.0em
			\rightmargin 0in
			\listparindent 0em
			\itemindent .75em
			\labelsep .75em
			\labelwidth 0in
		}%
	}{%
	\end{list}%
}

\begin{document}

\title[IMAGEO-Bench]{From Pixels to Places: A Systematic Benchmark for Evaluating Image Geolocalization Ability in Large Language Models}


\settopmatter{authorsperrow=4}
\author{Lingyao Li}
\authornote{Both authors contributed equally to this research, and both are corresponding authors.}
\affiliation{%
  \institution{University of South Florida}
  \city{Tampa}
  \country{USA}
}

\author{Runlong Yu}
\authornotemark[1]
\affiliation{%
  \institution{University of Alabama}
  \city{Tuscaloosa}
  \country{USA}
}

\author{Qikai Hu}
\affiliation{%
  \institution{University of Michigan}
  \city{Ann Arbor}
  \country{USA}}

\author{Bowei Li}
\affiliation{%
  \institution{Texas Tech University}
  \city{Lubbock}
  \country{USA}
}

\author{Min Deng}
\affiliation{%
 \institution{Texas Tech University}
 \city{Lubbock}
 \country{USA}}

\author{Yang Zhou}
\affiliation{%
  \institution{Texas A \& M University}
  \city{College Station}
  \country{USA}}

\author{Xiaowei Jia}
\affiliation{%
  \institution{University of Pittsburgh}
  \city{Pittsburgh}
  \country{USA}}



\begin{abstract}
Image geolocalization, the task of identifying the geographic location depicted in an image, is important for applications in crisis response, digital forensics, and location-based intelligence. While recent advances in large language models (LLMs) offer new opportunities for visual reasoning, their ability to perform image geolocalization remains underexplored. In this study, we introduce a benchmark called \textbf{IMAGEO-Bench} that systematically evaluates accuracy, distance error, geospatial bias, and reasoning process. Our benchmark includes three diverse datasets covering global street scenes, points of interest (POIs) in the United States, and a private collection of unseen images. Through experiments on 10 state-of-the-art LLMs, including both open- and closed-source models, we reveal clear performance disparities, with closed-source models generally showing stronger reasoning. Importantly, we uncover geospatial biases as LLMs tend to perform better in high-resource regions (e.g., North America, Western Europe, and California) while exhibiting degraded performance in underrepresented areas. Regression diagnostics demonstrate that successful geolocalization is primarily dependent on recognizing urban settings, outdoor environments, street-level imagery, and identifiable landmarks. Overall, \textbf{IMAGEO-Bench} provides a rigorous lens into the spatial reasoning capabilities of LLMs and offers implications for building geolocation-aware AI systems.

\vspace{0.1cm}
\noindent \textbf{Datasets:} \url{https://doi.org/10.5281/zenodo.16670471} \\
\noindent \textbf{Code:} \url{https://github.com/Lingyao1219/benchmark-reverse-geocoding}

\end{abstract}






\maketitle

\section{Introduction}

Image geolocalization, the process of inferring geographic coordinates or addresses from visuals, presents a complex challenge at the intersection of computer vision and geographic information systems (GIS) \cite{lin2013cross, workman2015wide, jin2017learned, zamir2014image}. Image geolocalization requires holistic scene interpretation and reasoning, encompassing both explicit indicators (e.g., landmarks, signage) and contextual cues (e.g., architectural styles, vegetation, spatial layout) to infer a precise location \cite{wu2022im2city, lin2013cross}. The accurate identification of location from images has important implications for real-world scenarios, including digital forensics \cite{kadha2025unravelling}, urban analytics \cite{lin2013cross} and crisis management \cite{li2025cross, yin2025llm}. 


Conventionally, this task has been investigated with computer vision models, using either supervised classification \cite{pramanick2022world, seo2018cplanet} or retrieval-based methods \cite{vivanco2023geoclip, zhu2022transgeo, vo2017revisiting, shatwell2025gt}. These models often leverage metadata embedded in image files \cite{samano2020you} or rely on extensive GIS databases \cite{hays2014large, brejcha2017state} to recognize landmark or visual cues. However, early methods often fail to generalize to previously unseen environments and perform inadequately in the absence of distinctive landmarks. Their effectiveness is further constrained by data coverage and resolution, especially when the metadata is missing. 

Recent advances in large language models (LLMs), such as GPT and Gemini, have transformed the landscape of image understanding and spatial reasoning \cite{chen2024spatialvlm, wu2024mind, li2025llms, tian2025nuscenes}. These models can describe intricate details, understand spatial relationships between objects, and infer context from subtle visual cues within an image \cite{chen2024spatialvlm, yang2025thinking}. By leveraging extensive world knowledge and visual-textual alignment, LLMs offer a promising avenue for overcoming the limits of conventional geolocalization, even when confronted with out-of-distribution data \cite{zhou2024img2loc, li2025recognition, xu2024addressclip}. Recent research further demonstrates that targeted strategies, such as supervised fine-tuning \cite{yi2025geolocsft} and retrieval-augmented inference \cite{li2025recognition}, can substantially enhance LLMs’ geolocalization capabilities.


In parallel with LLM advances, there has been rapid progress in developing benchmarks to evaluate their visual and linguistic capabilities \cite{lacoste2023geo, hua2025disentangling, fan2024nphardeval4v}. A representative example is GEO-Bench, which evaluates LLMs on classification and segmentation tasks in geoscience \cite{lacoste2023geo}. For image geolocalization, benchmarks like Img2Loc \cite{zhou2024img2loc} and LLMGeo \cite{wang2024llmgeo} have assessed LLMs' ability to localize images. However, these efforts often rely on established datasets and lag in incorporating the recent model advances. Moreover, key dimensions like LLMs' spatial reasoning abilities and potential geospatial biases remain underexplored. To address these gaps, we introduce \textbf{IMAGEO-Bench}, a benchmark that systematically evaluates LLMs' image geolocalization abilities. Our key contributions are listed as follows.

\begin{itemize}
    \item \textbf{Comprehensive benchmark datasets}. We compile three datasets with diverse geographic regions, urban/rural scenes, and location indicators, constructed via stratified sampling and augmented with privately collected images to ensure broad coverage.
    \item \textbf{Multi-dimensional evaluation framework}. We propose an evaluation framework that looks beyond standard accuracy to assess spatial precision and computational cost, providing a nuanced understanding of model performance across various dimensions.
    \item \textbf{Geographic bias quantification}. We quantify geographic biases in LLMs by revealing performance disparities across different regions at both global and local scales.
    \item \textbf{Reasoning process analysis}. We employ statistical regression to analyze the reasoning processes behind LLM geolocalization and to identify the most influential visual features.
\end{itemize}

\section{Related Work}




\subsection{Image Geolocalization}
\label{subsec:imageo_cv}

Image geolocalization in computer vision has focused on two main approaches: classification-based and retrieval-based models~\cite{weyand2016planet,seo2018cplanet,muller2018geolocation,pramanick2022world,clark2023we}. Classification methods (e.g., PlaNet~\cite{weyand2016planet}) treat geolocation as a discrete prediction task by assigning an image to a predefined spatial grid. These methods can benefit from global coverage and efficient inference but rely on rigid spatial discretization, which can constrain fine-grained accuracy and interpretability~\cite{seo2018cplanet}. Retrieval-based methods (e.g., IM2GPS~\cite{hays2008im2gps}) localize an image by matching it to a geotagged reference set based on visual or semantic similarity. Although enhancements like deep feature learning and probabilistic fusion~\cite{vo2017revisiting} have improved accuracy, these methods remain reliant on dense reference imagery and struggle in sparsely covered areas. Metric learning models like NetVLAD~\cite{arandjelovic2016netvlad} and VIGOR~\cite{zhu2021vigor} further boost retrieval performance using cross-view embeddings and attention mechanisms, particularly effective in visually rich urban settings.

However, these methods often face several limitations: (i) they rely on large labeled datasets and generalize inadequately to unseen or underrepresented regions; (ii) their predictions are often opaque, lacking interpretability or insight into the reasoning process; and (iii) they struggle to incorporate multimodal semantic cues (e.g., textual signage, architectural styles, or cultural markers) that humans intuitively use for geolocation. These limitations call for an approach that can leverages world knowledge and multimodal cues to generalize more effectively. In this regard, LLMs offer a promising solution due to their reasoning abilities, cross-modal integration, and greater interpretability.

\vspace{-0.1cm}
\subsection{LLMs for Spatial Reasoning}
\label{subsec:llm_spatial}

Recent progress in vision-language pre-training has given rise to multimodal LLMs with strong visual reasoning capabilities. Foundation models (e.g., CLIP~\cite{radford2021learning} and ALIGN~\cite{jia2021scaling}) have established powerful image-text alignment at web scale, enabling zero-shot classification and retrieval. BLIP-2~\cite{li2023blip} further advances this by connecting frozen language models with visual encoders, supporting open-ended visual description and optical character recognition (OCR). Modern multimodal LLMs (e.g., GPT, Gemini, and Qwen) have been demonstrated effective on tasks such as visual question answering~\cite{guo2023images, kuang2025natural}, document analysis~\cite{li2024enhancing}, and diagram understanding~\cite{hu2024mplug}. However, their spatial reasoning abilities, particularly for fine-grained geolocation, remain underexplored.

A few recent studies have begun to close this gap by evaluating modern foundation models' abilities to localize images~\cite{wang2024llmgeo, zhou2024img2loc}. For example, LLMGeo~\cite{wang2024llmgeo} benchmarks GPT-4V and LLaVA on challenging in-the-wild images. It reveals limited accuracy of investigated models and highlights room for improvement. Img2Loc~\cite{zhou2024img2loc} introduces a retrieval-augmented generation (RAG) pipeline that combines vision-language models with large-scale image search to refine coordinate estimates. A more recent study presents a benchmark built on Google Street View images~\cite{jay2025evaluating}, showing that VLMs can achieve sub-300km median errors but suffer from notable failure modes and reasoning transparency.

\begin{figure*}[htbp]
  \centering
  \includegraphics[width=1\textwidth]{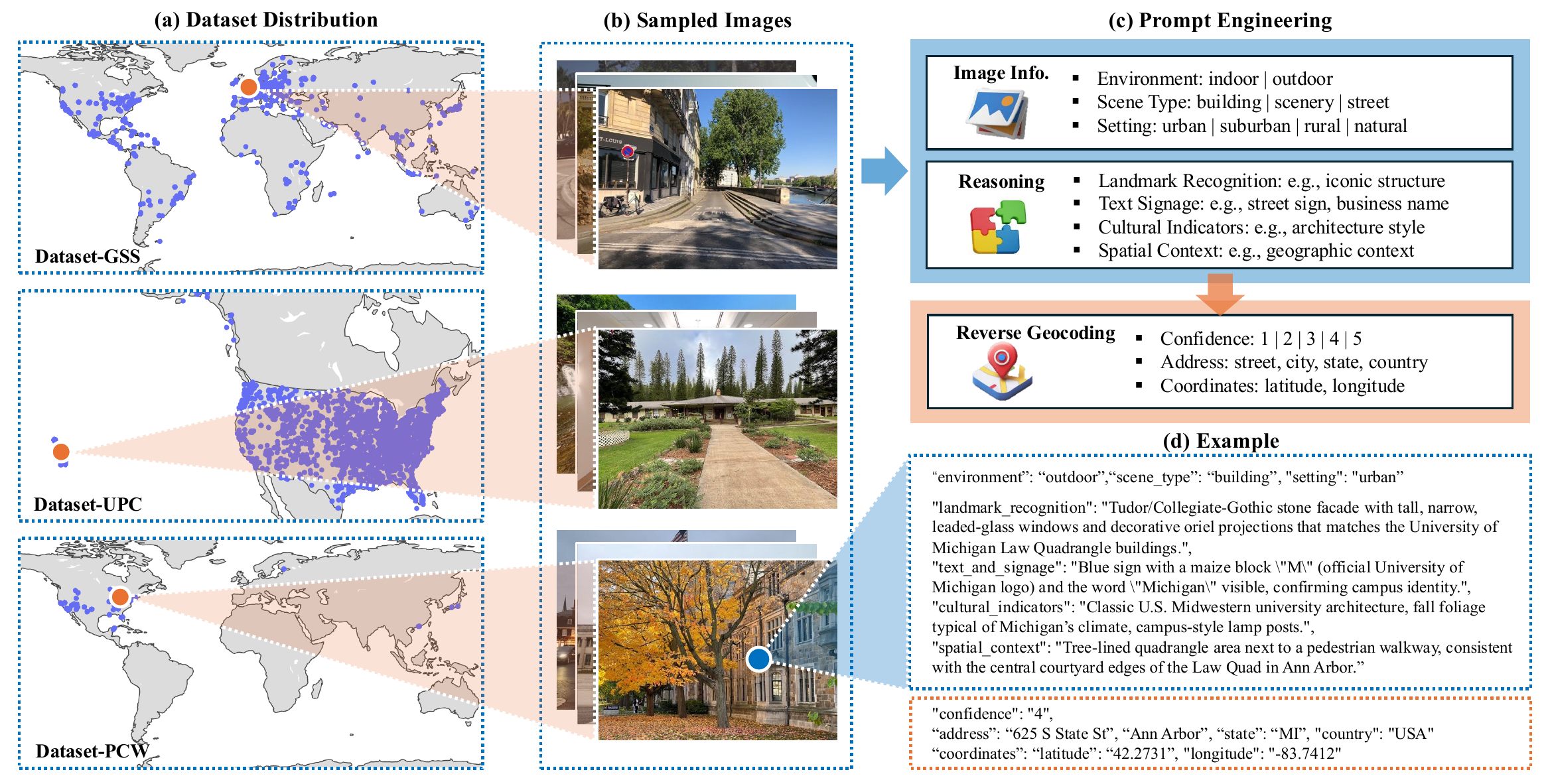}
  \caption{The illustrative framework to implement this study. (a) Data distribution for each benchmark dataset. (b) Sampled images from each dataset. (c) Prompt design. (d) An example of output by the o3 model.}
  \label{fig:framework}
\end{figure*}

While these studies represent important progress, they are constrained by task-specific designs, limited geographic scope, and the absence of standardized evaluation metrics. These underscore the need for a systematic framework to assess LLMs’ geolocation capabilities across diverse regions and real-world conditions. To address these gaps, we introduce \textbf{IMAGEO-Bench}, a benchmark for systematically evaluating LLMs on location accuracy, computational cost, geographic fairness, and reasoning process based on three newly compiled datasets. The framework of this benchmark is illustrated in Figure~\ref{fig:framework}.





\vspace{-0.1cm}
\section{Benchmark Dataset Preparation}

We construct three benchmark datasets from distinct sources to enable comprehensive evaluation, with their geographic distribution illustrated in Figure~\ref{fig:framework}(a). We opt to build new benchmark datasets for two main reasons. First, this helps mitigate data contamination, as widely used datasets like IM2GPS3k and YFCC4k \cite{hays2008im2gps} may have been incorporated in LLM training corpora. Second, our datasets are designed to span diverse visual contexts (e.g., indoor vs. outdoor) and geospatial scales (e.g., global vs. U.S. specific), allowing for a more rigorous assessment of LLM capabilities. Sample images of these benchmark datasets are presented in \textbf{Appendix~\ref{appendix:dataset}}. Please note that the distribution of some visual features, namely environment, scene type, and setting, are extracted from the outputs of gemini-2.5-pro, as our objective is not to evaluate the impact of specific image attributes on LLM geolocalization performance, but rather to summarize the model-identified features.

\vspace{-0.1cm}
\subsection{Dataset-GSS: Global Streetscape Set}

Dataset-GSS is derived from the NUS Global Streetscapes dataset \cite{hou2024global}, a collection of approximately 10 million streetscape images collected from Mapillary and KartaView across 688 cities worldwide. Each image is geo-tagged with precise latitude and longitude coordinates and has city-level labels. Then, the authors of the NUS dataset manually annotate a subset of over 10,000 images with eight contextual attributes: panoramic status, lighting condition, glare, view direction, image quality, reflection, weather, and platform \cite{hou2024global}.

We build Dataset-GSS by refining this manually labeled subset through a multi-stage filtering process. First, we filter the images with all eight contextual labels, which results in 8,568 images. Second, we only retain the images with the annotations---\textit{quality: good}, \textit{reflection: no}, and \textit{glare: no}---to ensure image quality, narrowing the dataset to 6,178 images. Last, we manually filter to exclude images with insufficient geographic information (e.g., no signs or clear visual cues). After filtering, we obtain Dataset-GSS, which comprises 6,152 high-quality street-level images from 396 cities across 123 countries. This carefully sampled dataset provides a rich variety of scenes captured under diverse geographies, environmental conditions, architectural styles, and cultural contexts. The data distribution of Dataset-GSS is presented in \textbf{Appendix~\ref{appendix:distribution}}.

\subsection{Dataset-UPC: U.S. POIs Crowdsourced Set}

Dataset-UPC is constructed from crowdsourced images from Google Maps Points of Interest (POIs) to capture a diverse range of regions and location types. We first sample POIs from the Google Maps dataset released by UC San Diego~\cite{yan2023personalized}, which contains 4,963,111 unique POIs across the United States collected up to September 2021. This dataset includes extensive attributes for each POI, including addresses (street, city, state, ZIP code), geolocation coordinates (latitude and longitude), business descriptions, and categorical labels, which makes it ideal for geolocalization evaluation.

To ensure geographic coverage in Dataset-UPC, we employ a multi-stage stratified sampling strategy. We begin with compiling a list of 17 distinct POI categories, including ``restaurant,'' ``hotel,'' ``park,'' and ``scenic spot,'' to guide the sampling. For each state, we randomly select three POIs from each category, resulting in a pool of 4,325 POIs. We then refine this pool through a two-step filtering process to ensure data usability and integrity. First, we discard entries with non-functional image URLs. Second, we manually remove advertisement images and those containing identifiable human faces. After this curation, Dataset-UPC comprises 2,929 images spanning all 50 states and the District of Columbia.

\vspace{-0.1cm}
\subsection{Dataset-PCW: Privately Collected Wild Set}

Dataset-PCW is constructed as a held-out validation set to prevent potential data leakage. Unlike Datasets-GSS or -UPC, it does not draw from publicly available sources. Instead, it consists of 220 original photo–address pairs collected by the authors. The images span a wide variety of scene types (e.g., natural, rural, urban), weather conditions (e.g., sunny, cloudy, snowy), and temporal contexts (e.g., daytime, nighttime). Since none of the images overlap with existing datasets or known LLM training sources, Dataset-PCW is well suited to evaluate LLMs’ geolocation performance on unseen data.


\vspace{-0.1cm}
\section{Benchmark Development}

\textbf{IMAGEO-Bench} defines a unified end-to-end task for image geolocation. It includes a structured output format, reasoning template, evaluation metrics, and post-hoc feature-importance analysis tools, allowing for clear and consistent comparisons across LLMs.

\vspace{-0.1cm}
\subsection{Task Formalization}
\label{subsec:task_formal}

Given an input image $\mathbf{x}$, the task requires models to predict a structured geographic output tuple: $G = (C,\;A,\;(\phi,\lambda))$, where $(\phi,\lambda)$ are the predicted latitude and longitude in decimal degrees, $A$ is a structured textual address, and $C \in \{1,2,3,4,5\}$ denotes the model's self-reported confidence score. The objective is to minimize the great-circle distance error $d((\phi,\lambda),(\phi^{\star},\lambda^{\star}))$ between the predicted and ground-truth coordinates. Predictions are provided in a structured JSON format for standardized parsing and evaluation. During inference, models are strictly restricted from accessing external APIs or GPS metadata.

\vspace{-0.1cm}
\subsection{Output Schema and Reasoning Template}

To evaluate LLMs’ geolocalization capabilities, we require models to explicitly reason about the given image by extracting and interpreting salient visual features, as illustrated in Figure~\ref{fig:framework}(c). These features are collected based on prior research identifying layout, signage, text, architecture, road, infrastructure, and buildings as critical cues for image geolocation \cite{li2025recognition, jay2025evaluating, wu2022im2city, tian2017cross}. Such structured output allows us to systematically analyze models' reasoning process. Since image geolocalization often depends on recognizing features tied to these specific places, prompting models to surface these elements can facilitate a step-by-step reasoning process similar to that of human geographers. Specifically, the structured model output comprises three key components:

\begin{itemize}
    \item \textbf{Image information:} High-level descriptors characterizing the environment (indoor, outdoor), scene type (building, scenery, street, mixed), and setting (urban, suburban, rural, natural).
    
    \item \textbf{Reasoning:} A detailed justification for location inference, organized into four distinct categories of visual evidence: (i) Landmark recognition: Unique natural or human-made structures serving as distinctive geographic markers. (ii) Text and signage: Textual clues from visible signs, license plates, storefronts, or other textual indicators. (iii) Cultural indicators: Region-specific architectural styles, cultural symbols, or prevalent societal patterns. (iv)~Spatial context: Environmental layout, vegetation patterns, road networks, infrastructure types, or building density.
    
    \item \textbf{Geolocalization}: The final structured geographic prediction tuple $(C, A, (\phi, \lambda))$, as formally defined before.
\end{itemize}

\vspace{-0.1cm}
\subsection{Evaluation Metrics}

We establish a set of metrics to measure LLMs' performance:

\begin{itemize}
    \item \textbf{Identification rate:} It measures the model’s ability to produce a valid output, defined as the percentage of input images for which a complete and correctly formatted JSON is returned. Instances where the model fails to produce a usable output are treated as identification failures.

    \item \textbf{Geographic accuracy:} It evaluates the semantic correctness of the predicted address $A_{pred}$ against the ground truth $A_{true}$ at three administrative levels, including (i) Country accuracy ($Acc_{\text{country}}$), (ii) State accuracy ($Acc_{\text{state}}$), and (iii) City accuracy ($Acc_{\text{city}}$). This metric quantifies the proportion of predictions that align with the ground-truth labels.

    \item \textbf{Distance error:} It calculates the great-circle distance $d$ (kilometers) between the predicted coordinates $(\phi_{\text{p}}, \lambda_{\text{p}})$ and the true coordinates $(\phi_{\text{t}}, \lambda_{\text{t}})$ using the Haversine formula:

    \vspace{-2mm}
    \begin{equation}
        d = 2r \arcsin(\sqrt{\sin^2(\Delta\phi/2) + \cos(\phi_{\text{p}})\cos(\phi_{\text{t}})\sin^2(\Delta\lambda/2)})
    \end{equation}
    
    where $\Delta\phi = \phi_{\text{t}} - \phi_{\text{p}}$, $\Delta\lambda = \lambda_{\text{t}} - \lambda_{\text{p}}$, and $r$ represents the Earth's mean radius (6,371 km).

    \item \textbf{Computational cost:} It assesses the model's efficiency by tracking the computational costs, including the averaged cost $Avg_{\text{Cost}}$ and the tokens per inference $Avg_{\text{Tokens}}$ (i.e., the average number of input and output tokens consumed for a single image query).
\end{itemize}


\vspace{-0.1cm}
\subsection{Post-hoc Feature-Importance Diagnostics}
\label{subsec:feature_diag}

\textbf{IMAGEO-Bench} includes an interpretability module to assess how specific structured output fields correlate with geolocalization performance. This diagnostic suite supports model-agnostic evaluation, identifying which visual attributes are most predictive of successful localization. Specifically, we construct a diagnostic set from each model’s structured JSON outputs. Key categorical predictors—such as environment type, scene category, setting, and landmark presence—are extracted from the predefined schema and one-hot encoded relative to baseline categories. We further define two metrics to represent the geolocation outcome, including (i) a binary indicator representing city-level prediction correctness, and (ii) a continuous log-transformed distance error. Let $\mathbf{x} \in \mathbb{R}^d$ denote this encoded feature vector, and let $Y$ represent the geolocation outcome (e.g., city-level correctness or log-distance error). To quantify feature importance, we fit the following two regression models.

    

\begin{itemize}
  \item \textbf{Logistic regression.} 
        For the city-level correctness indicator $Y \in \{0,1\}$, we estimate \vspace{-0.1cm}
    \begin{equation} 
        \mathbb{P}(Y = 1 \mid \mathbf{x}) = \sigma(\mathbf{w}^\top \mathbf{x} + b),
    \end{equation}
        where $\sigma(z) = 1 / (1 + e^{-z})$ is the sigmoid function.  
        We report~$\exp(w_j)$ as the odds ratio of feature $j$ relative to its reference category.

  \item \textbf{Ridge regression.}  
        For the log-distance error, we fit: \vspace{-0.05cm}
    \begin{equation} 
        \log (d) = \mathbf{w}^\top \mathbf{x} + b + \varepsilon,
    \end{equation}  
        where $\varepsilon \sim \mathcal{N}(0, \sigma^2)$,  
        by minimizing the penalized squared error $\sum_i (y_i - \hat{y}_i)^2 + \lambda \lVert \mathbf{w} \rVert_2^2$ with $\lambda = 1.0$.
\end{itemize}

All predictors are one-hot encoded binary variables, so additional normalization is unnecessary.  
Because all features share the same scale, the ridge coefficients $\mathbf{w}$ are directly comparable. We aggregate $|w_j|$ within each semantic field to obtain field-level importance.

The importance of the field level is calculated by aggregating the absolute regression coefficients associated with each predictor group. Coefficients are interpreted relative to their baseline categories, with careful consideration given to potential collinearity effects. These diagnostic results provide valuable information on the semantic and visual cues that most strongly influence model geolocalization
performance.

\begin{figure*}[htbp]
    \centering
    \begin{subfigure}{1\linewidth}
        \includegraphics[width=\linewidth]{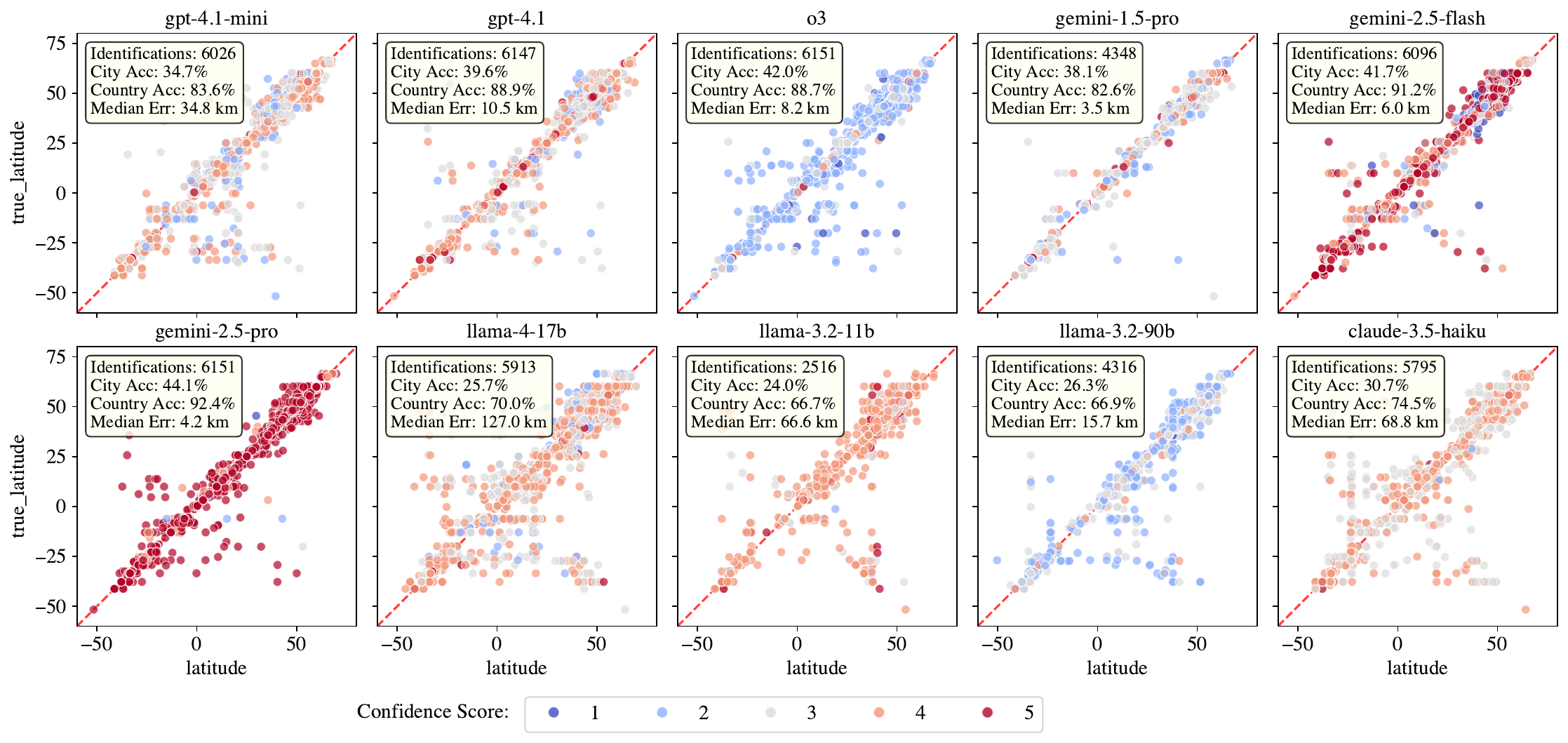}
        \hspace*{-55em}\raisebox{1em}{\small\textbf{(a)}}
    \end{subfigure}
    \begin{subfigure}{1\linewidth}
        \includegraphics[width=\linewidth]{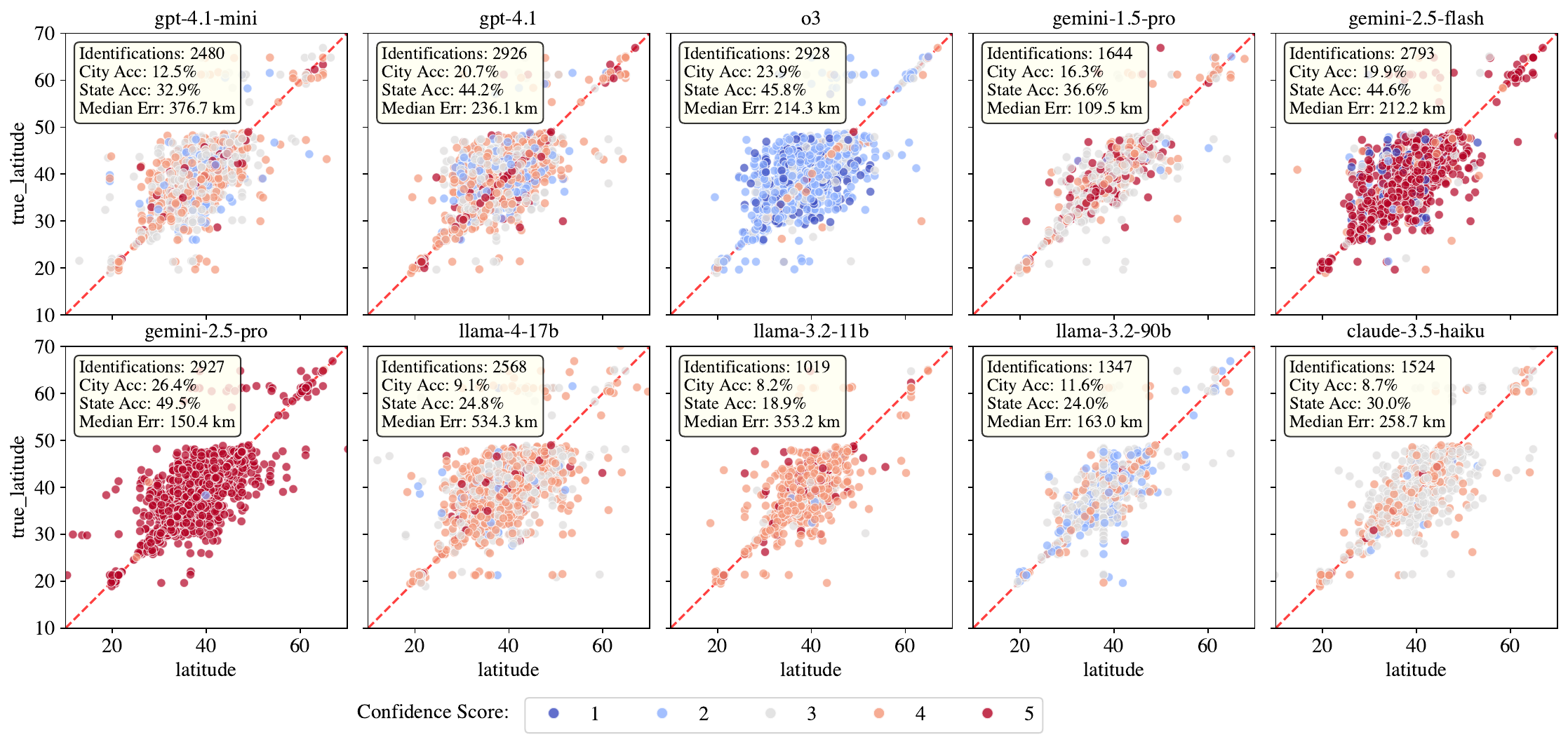}
        \hspace*{-55em}\raisebox{1em}{\small\textbf{(b)}}
    \end{subfigure}
    \vspace{-0.6cm}
    \caption{Benchmark performance based on latitude prediction. (a) Dataset-GSS; (b) Dataset-UPC. Perfect predictions lie on the red dashed diagonal line. The proximity of points to this line indicates the model's accuracy, with the color of each point representing the model's self-reported confidence score on a scale of 1 to 5.}
    \vspace{-0.4cm}
    \label{fig:latitude_bench}
\end{figure*}

\section{Experimental Setup}

\subsection{Prompt Design}

We employ a structured prompting methodology to ensure consistent and parsable outputs. Our approach combines a system prompt, which defines the model's role as a ``geocoding analysis'' expert, with a main prompt that directs the model to use a step-by-step, chain-of-thought reasoning process. To further guide the model's inference, we include a one-shot example within the prompt. We mandate that the response be a valid JSON object partitioned into three sections: (i) Image information for high-level scene classification, (ii) reasoning for a detailed analysis of visual evidence, and (iii) image geolocalization for the final coordinate estimation. This standardized design facilitates a direct and objective comparison of the feature extraction and inference capabilities across all models. Specific prompt design is presented in \textbf{Appendix~\ref{appendix:prompt}}.

\begin{table*}[h]
\centering
\vspace{-0.2cm}
\caption{LLM performance benchmark based on Dataset-GSS and Dataset-UPC.}
\vspace{-0.2cm}
\small
\setlength{\tabcolsep}{4pt}
\begin{tabular}{lcccccccccccc}
\toprule
 & \multicolumn{6}{c}{\textbf{Dataset-GSS}} & \multicolumn{6}{c}{\textbf{Dataset-UPC}} \\
\hline
\textbf{Model} & \textbf{City} & \textbf{Country} & \textbf{Mean} & \textbf{Median} & \textbf{Avg} & \textbf{Avg} & \textbf{City} & \textbf{State} & \textbf{Mean} & \textbf{Median} & \textbf{Avg} & \textbf{Avg} \\
 & \textbf{Acc (\%)} & \textbf{Acc (\%)} & \textbf{Dist (km)} & \textbf{Dist (km)} & \textbf{Tokens} & \textbf{Cost (\$)} & \textbf{Acc (\%)} & \textbf{Acc (\%)} & \textbf{Dist (km)} & \textbf{Dist (km)} & \textbf{Tokens} & \textbf{Cost (\$)} \\
\midrule
gpt-4.1-mini & 34.7\% & 83.6\% & 381.1 & 34.8 & 3245 & 0.00164 & 12.5\% & 32.9\% & 768.8 & 376.7 & 1496 & 0.00092 \\
gpt-4.1 & 39.6\% & 88.9\% & 296.1 & 10.5 & 1842 & 0.00536 & 20.7\% & 44.2\% & 685.8 & 236.1 & 1465 & 0.00455 \\
o3 & 42.0\% & 88.7\% & 288.1 & 8.2 & 2912 & 0.01460 & 23.9\% & 45.8\% & 662.7 & 214.3 & 2281 & 0.01150 \\
gemini-1.5-pro & 38.1\% & 82.6\% & \textbf{161.4} & \textbf{3.5} & 1178 & 0.00243 & 16.3\% & 36.6\% & \textbf{406.5} & \textbf{109.5} & 1165 & 0.00237 \\
gemini-2.5-flash & 41.7\% & 91.2\% & 227.0 & 6.0 & 1379 & 0.00141 & 19.9\% & 44.6\% & 587.7 & 212.2 & 1326 & 0.00128 \\
gemini-2.5-pro & \textbf{44.1\%} & \textbf{92.4\%} & 216.3 & 4.2 & 1367 & 0.00556 & \textbf{26.4\%} & \textbf{49.5\%} & 544.4 & 150.4 & 1328 & 0.00517 \\
llama-4-17b & 25.7\% & 70.0\% & 840.3 & 127.0 & 2961 & 0.00093 & 9.1\% & 24.8\% & 1217.5 & 534.3 & 1767 & 0.00061 \\
llama-3.2-11b & 24.0\% & 66.7\% & 797.4 & 66.6 & \textbf{823} & \textbf{0.00015} & 8.2\% & 18.9\% & 955.5 & 353.2 & \textbf{822} & \textbf{0.00015} \\
llama-3.2-90b & 26.3\% & 66.9\% & 382.9 & 15.7 & 7211 & 0.00865 & 11.6\% & 24.0\% & 706.8 & 163.0 & 5206 & 0.00625 \\
claude-3.5-haiku & 30.7\% & 74.5\% & 568.2 & 68.8 & 2478 & 0.00281 & 8.7\% & 30.0\% & 697.1 & 258.7 & 1421 & 0.00190 \\
\bottomrule
\vspace{-0.5cm}
\label{table:performance}
\end{tabular}
\end{table*}

\subsection{Model Selection and Experiment Settings}

For our benchmark analysis, we select a diverse set of candidate LLMs with vision capabilities from several prominent providers. The models under evaluation include OpenAI's o3, gpt-4.1, and gpt-4.1-mini; Anthropic's claude-3.5-haiku; Google's gemini-1.5-pro, gemini-2.5-flash, and gemini-2.5-pro; and various Llama models, specifically llama-3.2-11b, llama-3.2-90b, and llama-4-maverick-17b. We interface with each model through its designated API. 

In our experimental setup, we query each model with a temperature setting of 0.7, which encourages a balance between factual recall and inferential creativity. We log all outputs for each image, including the model's raw JSON response and error states, to a .jsonl file for subsequent analysis. At the conclusion of each experimental run, we also compute and record a summary of the total operational cost and token consumption. Representative model outputs of image geolocalization are provided in \textbf{Appendix~\ref{appendix:output}}.

\section{Results and Analysis}

\subsection{Benchmark Performance}
\label{subsec:bench_perf}

To benchmark LLMs' performance, we generate two sets of scatter plots using the introduced metrics. Table~\ref{table:performance} presents a quantitative summary, including accuracy at various geospatial scales, median distance error, and computational costs. For visual analysis, Figure~\ref{fig:latitude_bench} displays scatter plots for latitude prediction on (a) Dataset-GSS and (b)~Dataset-UPC, while corresponding longitude results are provided in \textbf{Appendix~\ref{appendix:longitude}}. The benchmark performance for Dataset-PCW is included in \textbf{Appendix~\ref{appendix:pcw}}.

Performance across Dataset-GSS and Dataset-UPC shows clear differences, with most LLMs exhibiting higher prediction accuracy on Dataset-GSS. This discrepancy is likely due to the inherent characteristics of the compiled datasets. Dataset-GSS primarily consists of publicly available global street-level images captured in outdoor environments, which provide clear geographic indicators. In contrast, Dataset-UPC comprises crowdsourced images contributed by Google Maps users, many of which are taken indoors, lack explicit geographic cues or distinctive architectural details. These characteristics significantly complicate accurate identification of geolocation. As a result, most LLMs achieve higher accuracy with Dataset-GSS across all evaluated levels, including city, country, and state, and report lower median distance errors.


Closed-source models consistently outperform their open-source counterparts across both datasets. Specifically, gemini-2.5-pro achieves remarkable results in Dataset-GSS, with a median error of 4.2 km and country-level accuracy of 92.4\%. Other top performers like gpt-4.1 and o3 maintain high performance with median errors of 10.5 km and 8.2 km, respectively, in Dataset-GSS. Conversely, open-source models such as llama-4-17b and llama-3.2-11b exhibit substantially higher median errors (e.g., 127.0 km for llama-4-17b in Dataset-GSS and 534.3 km in Dataset-UPC). This performance gap likely originates from proprietary advantages of closed-source providers, including larger training datasets, richer geospatial corpora, and more refined model architectures.

We also observe a clear scaling trend, where larger models generally exhibit stronger capabilities. For example, gpt-4.1 surpasses its smaller counterpart, gpt-4.1-mini, with a city-level accuracy of 39.6\% versus 34.7\% and a much lower median distance error (10.5 km versus 34.8 km). Similarly, llama-3.2-90b improves upon llama-3.2-11b with a city-level accuracy of 26.3\% versus 24.0\%, and a median distance error of 15.7 km versus 66.6 km. However, performance differences are not solely size-dependent. Smaller models like gemini-2.5-flash (city accuracy 41.7\%, median error 6.0 km) outperform many larger models, suggesting that other factors (e.g., model architecture, multimodal fusion techniques, and training quality) can contribute to improved performance. This reinforces that scaling is helpful but not sufficient on its own.

Confidence scores vary significantly across the models, among which gemini-2.5-pro exhibits the highest confidence, while o3 reports the lowest. Even within a single provider, confidence levels differ: for instance, gpt-4.1 is substantially more confident than gpt-4.1-mini, and gemini-2.5-pro scores higher than gemini-1.5-pro. For models like gpt-4.1, o3, and claude-3.5-haiku, higher confidence scores generally align with better prediction accuracy within the same model. However, higher confidence does not always imply greater performance across models. For instance, gemini-2.5-pro shows a higher confidence distribution than o3, yet their performance in Dataset-GSS is comparable (median errors of 4.2 km versus 8.2 km). These findings indicate that confidence alone is not a reliable proxy for model accuracy and should be interpreted in conjunction with performance metrics.

In addition, the computational cost reveal significant variations across models. llama-3.2-11b emerges as the most efficient option, using only $\sim$823 tokens at \$0.00015 per request, while o3 is the most expensive model at \$0.01460 per request in Dataset-GSS. However, this cost savings of open-sourced models comes with substantially lower performance. Among closed-source models, gemini-2.5-flash offers an optimal balance, achieving 41.7\% city accuracy at just \$0.00141 per request. Interestingly, model size doesn't always correlate to resource consumption. For example, gpt-4.1-mini consumes 3,245 tokens despite its ``mini'' label, nearly 3$\times$ of gemini-1.5-pro's 1,178 tokens. These findings highlight a trade-off between cost-efficiency and performance for effective image geolocalization.

\begin{figure}[t]
  \centering
  \includegraphics[width=0.5\textwidth]{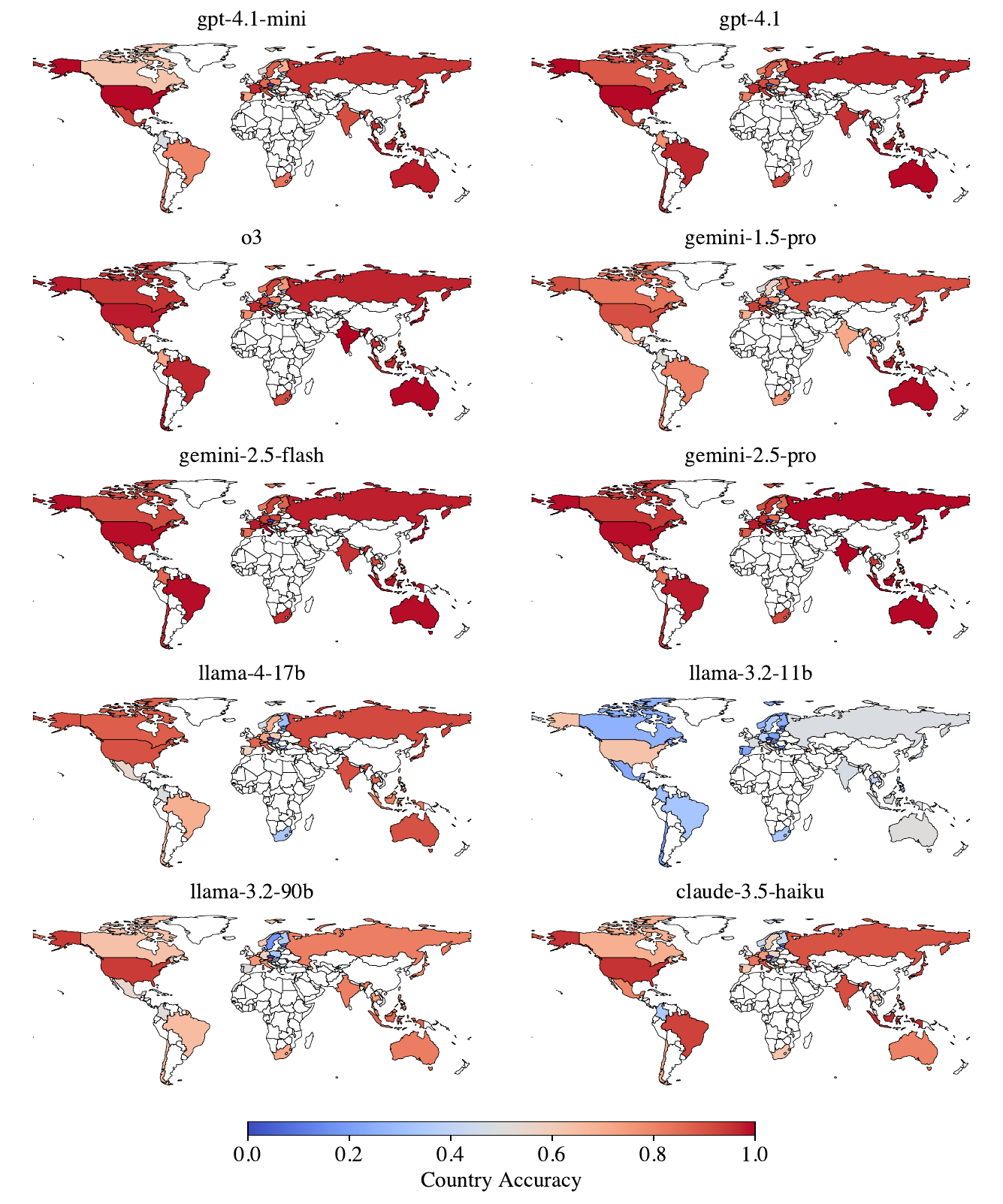}
  \vspace{-0.4cm}
  \caption{Country-level averaged accuracy across models on Dataset-GSS (a threshold of 20 sampled images).}
  \vspace{-0.4cm}
  \label{fig:dataset1_map}
\end{figure}

\begin{figure}[t]
  \centering
  \includegraphics[width=0.44\textwidth]{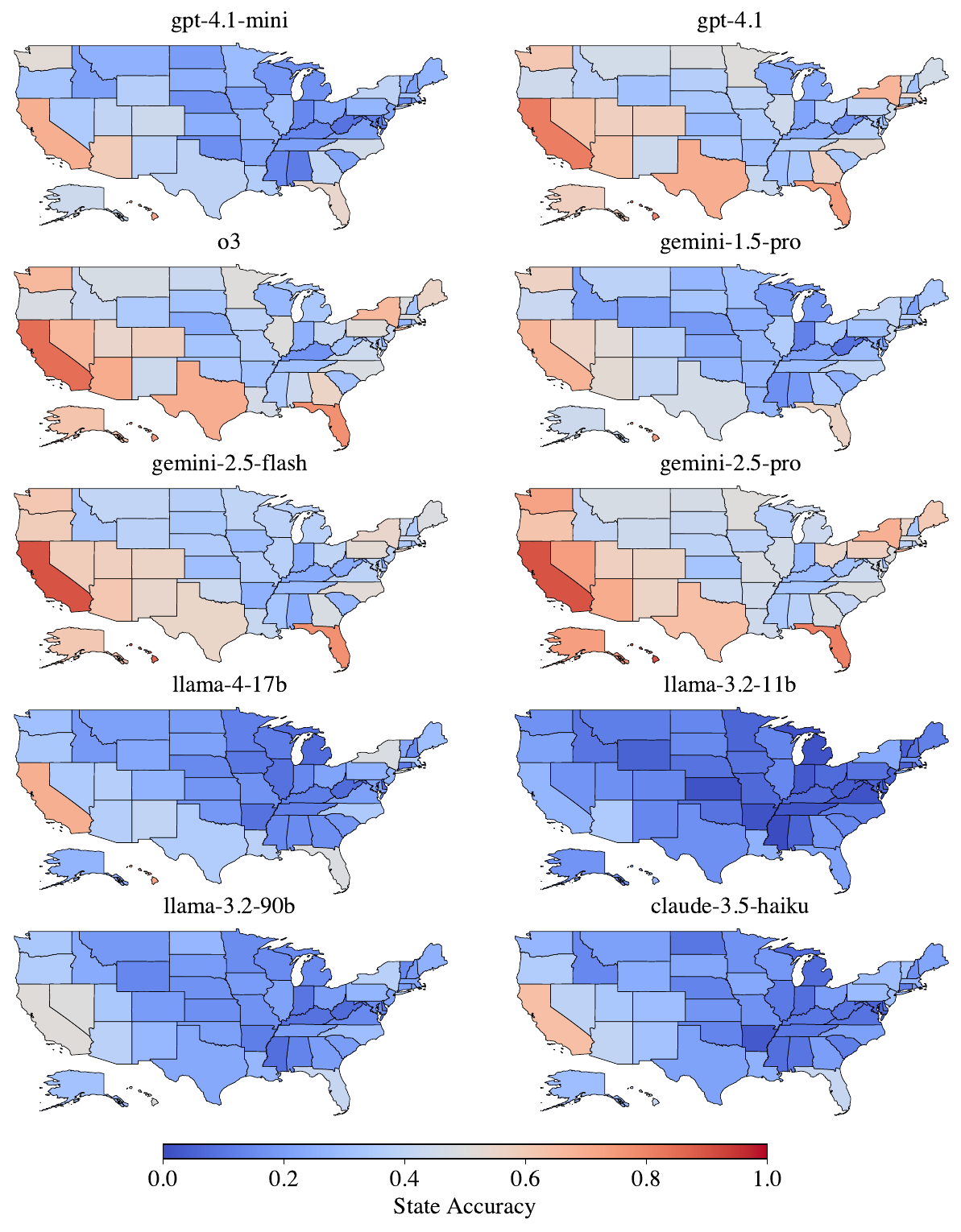}
  \vspace{-0.3cm}
  \caption{U.S. state-level averaged accuracy across models on Dataset-UPC (a threshold of 20 sampled images).}
  \vspace{-0.4cm}
  \label{fig:dataset2_map}
\end{figure}

\subsection{Geospatial Analysis}
\label{subsec:geospatial}

We conduct geospatial analyses on Dataset-GSS and Dataset-UPC to evaluate model performance across geographic regions and scales. In Dataset-GSS (Figure~\ref{fig:dataset1_map}), we calculate country-level accuracy for nations with over 20 samples to ensure statistical reliability. Results reveal pronounced geographic bias and model variability. Models such as o3 and gemini-2.5-pro achieve near-perfect accuracy (approaching 1.0) in regions like North America, Russia, and Australia. In contrast, llama-3.2-11b consistently underperforms, with accuracy often below 0.6 and approaching zero in several Latin European countries. Models like llama-3.2-90b and claude-3.5-haiku show intermediate accuracy (0.6–0.9). It is noting that the United States, Australia, and West European Countries maintain high accuracy across all models, suggesting strong model generalization in these well-represented regions.

In Dataset-UPC (Figure~\ref{fig:dataset2_map}), we again observe distinct geographic disparities in model performance across U.S. states. Specifically, gemini-2.5-flash, gemini-2.5-pro, and o3 consistently perform well, particularly in California, Hawaii, and Florida. In contrast, llama-3.2-11b falls below 0.5 accuracy in nearly all states. gpt-4.1 shows moderate performance, with accuracy typically between 0.6 and 0.8 in coastal regions. California and Florida emerge as consistent high-performing states across models, likely due to their dense urban environments and distinctive landmarks that can facilitate image geolocalization. 

Together, the results from both datasets illustrate potential geospatial biases in model behavior. High performance in regions like the United States, Australia, and parts of Western Europe points to the models’ strong alignment with densely sampled, visually distinctive, and data-rich environments. In contrast, accuracy declines sharply in countries and states with fewer training representations or more ambiguous visual features. These patterns reflect a potential limitation that LLMs are not yet equipped to generalize robustly across diverse global geographies.


\subsection{Reasoning Analysis}
\label{subsec:feat_attr}

\begin{figure}[t]
  \centering
  \includegraphics[width=\linewidth]{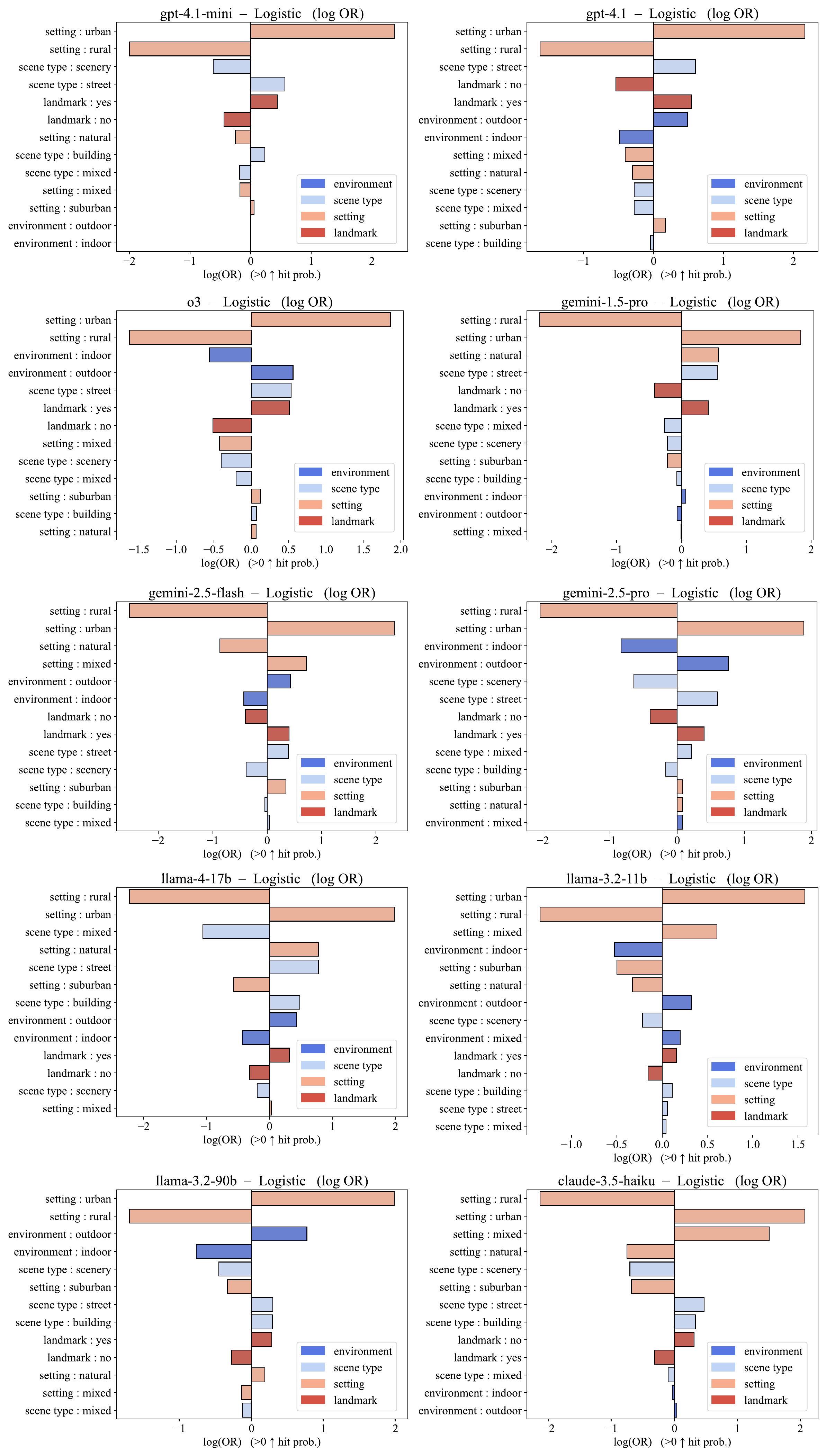}
   \vspace{-0.4cm}
  \caption{Log-odds coefficients from diagnostic logistic regression on Dataset-GSS. Positive values increase the likelihood of correct city prediction; negative values decrease it.}
  \vspace{-0.4cm}
  \label{fig:reason_city}
\end{figure}

\subsubsection{Feature attribution via reasoning}

We apply logistic regression diagnostics to examine the features for the LLM reasoning process. The summarized result for Dataset-GSS is illustrated in Figure~\ref{fig:reason_city}. Image elements are categorized into four semantic groups: environment, scene type, setting, and landmark. Across the evaluated LLMs, we observe consistent trends as well as model-specific variations. The main findings are as follows. 

First, nearly all models strongly favor urban and outdoor environments while consistently penalizing rural, natural, and indoor scenes. One possible explanation is that dense urban contexts, featuring detailed façades, prominent signage, and infrastructure elements (e.g., roads and street furniture), provide reliable visual cues for accurate geolocalization. Conversely, natural scenes, which often lack distinctive visual anchors, pose significant localization challenges, as reflected by consistently negative model associations.

Second, regarding scene type, street-level imagery tends to provide stronger localization signals than building façades or general scenery. This is likely because street scenes often contain rich contextual cues (e.g., textual signage, road markings, and distinctive urban infrastructure) that are both visually salient and semantically interpretable by LLMs. In contrast, scenic images without clear architectural landmarks can negatively impact prediction accuracy.

Third, the presence of distinctive landmarks consistently improves the geolocation performance across nearly all models. Landmark cues are typically assigned strong positive weights, indicating their critical role in guiding correct predictions. However, the claude-3.5-haiku model is a notable outlier: it assigns marginal or even negative weights to the landmark cue. A post-hoc audit reveals that it labels ``landmark = yes'' on 99.67\% of images, compared to 41.77\% for all other models. This over-identification of landmarks suggests that the model frequently misclassifies ordinary structures as distinctive, thereby diluting the predictive value of this feature.

Overall, current LLMs perform best when presented with clear, distinct visual elements, including textual signage, iconic architecture, and densely structured urban layouts. They struggle significantly, however, in contexts that lack easily verbalizable landmarks or signage. Additional analyses and visualizations, including distance error evaluations and results on the UPC and PCW datasets, are provided in \textbf{Appendix~\ref{app:extended_attr}}.


\begin{figure}[t]
  \centering
  \includegraphics[width=1\linewidth]{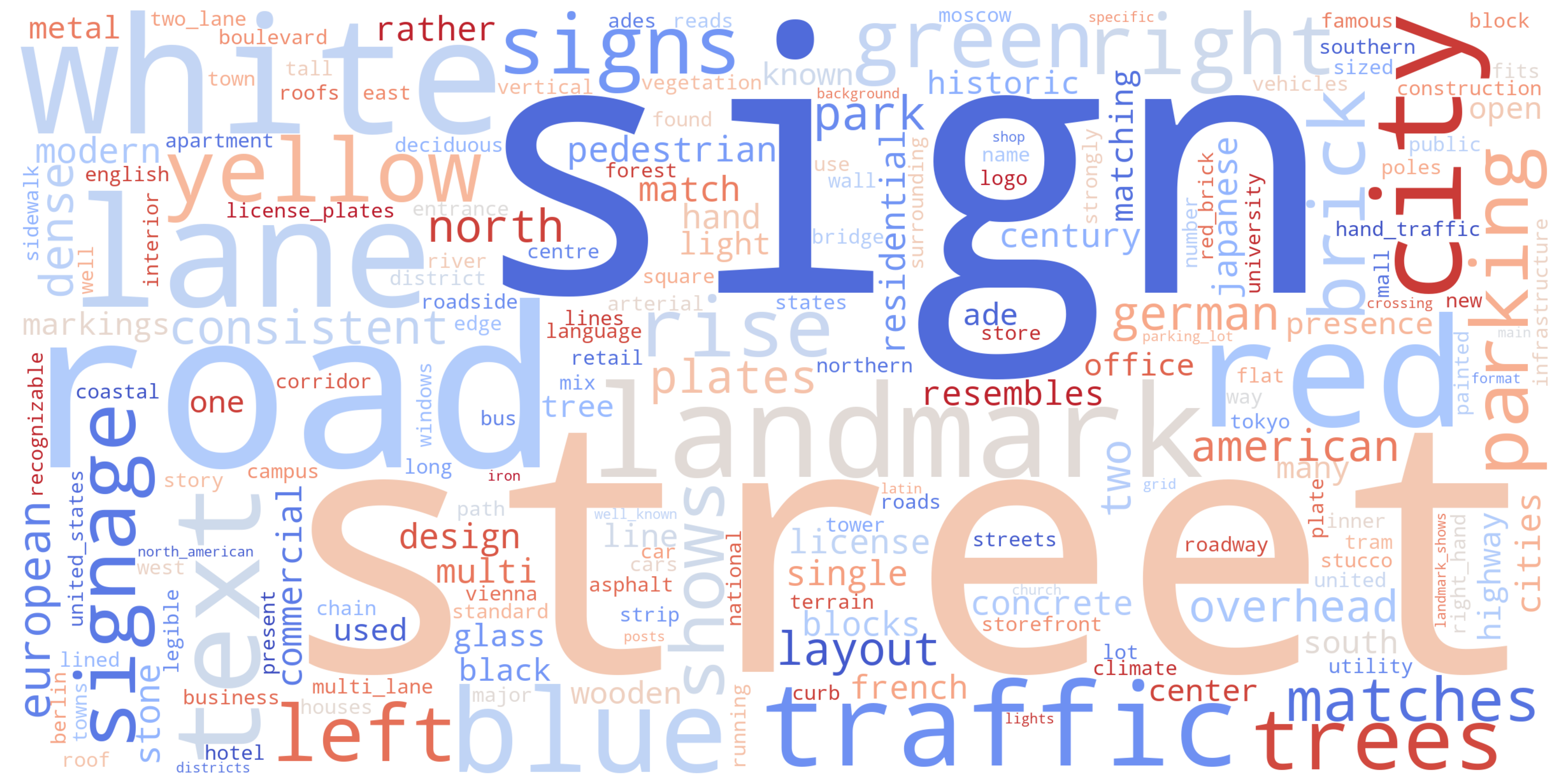}
   \vspace{-0.4cm}
  \caption{Mixed unigram–bigram word cloud derived from the reasoning fields of the o3 model. Font size reflects token frequency.}%
  \vspace{-0.5cm}
\label{fig:wordcloud_o3}
\end{figure}

\subsubsection{Word-Cloud Analysis}

In addition, we perform a word cloud analysis to provide a lexical overview of the cues used by the o3 model in three datasets, as shown in Figure~\ref{fig:wordcloud_o3}. First, the terms of the traffic infrastructure dominate the model outputs. Terms such as ``sign,'' ``street,'' ``road,'' ``lane,'' and ``traffic light'' appear more frequently, reflecting the strong reliance on OCR-visible elements and road geometry. Second, color terms, such as ``red,'' ``white,'' ``yellow,'' ``blue,'' ``green'', function as indirect geographic signals. These cues are often tied to traffic regulations (e.g., yellow center lines in the United States vs. white in Europe), enabling the model to infer regional priors based on color patterns. Third, regional adjectives are sparse but informative. Although generic terms like ``brick'' or ``trees'' are common, less frequent modifiers such as ``German,'' ``Japanese,'' or ``American'' suggest that o3 rarely states a location outright, instead embedding regional cues in compound phrases like ``license plates'' or ``right hand'' (driving side). In general, the lexical profile indicates that o3 is highly based on discrete, regulative, and text-based anchors, while underutilizing cues related to vegetation, climate, or terrain. A broader comparison across nine additional models is provided in \textbf{Appendix~\ref{app:wordclouds}}.


\section{Discussion and Conclusions}

Our study introduces \textbf{IMAGEO-Bench}, a comprehensive benchmark designed to evaluate the image geolocalization capabilities of LLMs across accuracy, computational cost, geographic fairness, and reasoning process. Our key findings are summarized below.

\textbf{Closed-source models lead in performance but at a higher cost}. Models like gemini-2.5-pro and o3 consistently outperform open-source alternatives in accuracy and distance error. However, this comes with significantly higher computational costs. gemini-2.5-flash offers a compelling balance between cost-efficiency and accuracy. This suggest that model architecture, training data quality, and modality alignment can outweigh parameter count in determining performance.

\textbf{Geographic performance disparities reflect potential training data and reasoning biases}. Models consistently perform better in high-resource, visually distinctive regions like North America, Western Europe, and Australia, while struggling in underrepresented or ambiguous areas. States like California and Florida see higher accuracy, likely due to richer training data and clearer visual cues. These disparities highlight limitations in global generalization and raise concerns about geographic fairness.

\textbf{Models rely on visual anchors, not holistic understanding}. LLMs tend to perform better in outdoor urban scenes with clear cues like signage and landmarks, while rural, natural, and indoor images lead to higher errors due to limited identifiable features. Models can infer broad regional context (e.g., country-level), but struggle with fine-grained localization in the absence of strong visual anchors. Moreover, spatial layout or image composition appears to have little impact on prediction performance. 

This study opens several avenues for future research. One direction could conduct a more comprehensive study covering diverse geographies and scene types to assess model generalization under real-world variability. Another future work could explore how language-based cues contribute to geolocation predictions, especially in scenes lacking visual anchors. Lastly, future work could explore strategies such as data augmentation or region-adaptive prompting to reduce geographic bias—an essential step toward building more equitable and globally robust geolocation systems.

\bibliographystyle{ACM-Reference-Format}
\bibliography{main}

@inproceedings{lin2013cross,
  title={Cross-view image geolocalization},
  author={Lin, Tsung-Yi and Belongie, Serge and Hays, James},
  booktitle={Proceedings of the IEEE Conference on Computer Vision and Pattern Recognition},
  pages={891--898},
  year={2013}
}

@inproceedings{workman2015wide,
  title={Wide-area image geolocalization with aerial reference imagery},
  author={Workman, Scott and Souvenir, Richard and Jacobs, Nathan},
  booktitle={Proceedings of the IEEE International Conference on Computer Vision},
  pages={3961--3969},
  year={2015}
}

@inproceedings{jin2017learned,
  title={Learned contextual feature reweighting for image geo-localization},
  author={Jin Kim, Hyo and Dunn, Enrique and Frahm, Jan-Michael},
  booktitle={Proceedings of the IEEE Conference on Computer Vision and Pattern Recognition},
  pages={2136--2145},
  year={2017}
}

@article{zamir2014image,
  title={Image geo-localization based on multiplenearest neighbor feature matching usinggeneralized graphs},
  author={Zamir, Amir Roshan and Shah, Mubarak},
  journal={IEEE transactions on pattern analysis and machine intelligence},
  volume={36},
  number={8},
  pages={1546--1558},
  year={2014},
  publisher={IEEE}
}

@inproceedings{wu2022im2city,
  title={Im2city: image geo-localization via multi-modal learning},
  author={Wu, Meiliu and Huang, Qunying},
  booktitle={Proceedings of the 5th ACM SIGSPATIAL International Workshop on AI for Geographic Knowledge Discovery},
  pages={50--61},
  year={2022}
}

@article{kadha2025unravelling,
  title={Unravelling Digital Forgeries: A Systematic Survey on Image Manipulation Detection and Localization},
  author={Kadha, VijayaKumar and Bakshi, Sambit and Das, Santos Kumar},
  journal={ACM Computing Surveys},
  volume={57},
  number={12},
  pages={1--36},
  year={2025},
  publisher={ACM New York, NY}
}

@article{li2025cross,
  title={Cross-view geolocalization and disaster mapping with street-view and VHR satellite imagery: A case study of Hurricane IAN},
  author={Li, Hao and Deuser, Fabian and Yin, Wenping and Luo, Xuanshu and Walther, Paul and Mai, Gengchen and Huang, Wei and Werner, Martin},
  journal={ISPRS Journal of Photogrammetry and Remote Sensing},
  volume={220},
  pages={841--854},
  year={2025},
  publisher={Elsevier}
}

@inproceedings{samano2020you,
  title={You are here: Geolocation by embedding maps and images},
  author={Samano, Noe and Zhou, Mengjie and Calway, Andrew},
  booktitle={European Conference on Computer Vision},
  pages={502--518},
  year={2020},
  organization={Springer}
}

@article{yin2025llm,
  title={LLM-enhanced disaster geolocalization using implicit geoinformation from multimodal data: A case study of Hurricane Harvey},
  author={Yin, Wenping and Xue, Yong and Liu, Ziqi and Li, Hao and Werner, Martin},
  journal={International Journal of Applied Earth Observation and Geoinformation},
  volume={137},
  pages={104423},
  year={2025},
  publisher={Elsevier}
}

@incollection{hays2014large,
  title={Large-scale image geolocalization},
  author={Hays, James and Efros, Alexei A},
  booktitle={Multimodal location estimation of videos and images},
  pages={41--62},
  year={2014},
  publisher={Springer}
}

@article{brejcha2017state,
  title={State-of-the-art in visual geo-localization},
  author={Brejcha, Jan and {\v{C}}ad{\'\i}k, Martin},
  journal={Pattern Analysis and Applications},
  volume={20},
  number={3},
  pages={613--637},
  year={2017},
  publisher={Springer}
}

@article{fan2024nphardeval4v,
  title={Nphardeval4v: A dynamic reasoning benchmark of multimodal large language models},
  author={Fan, Lizhou and Hua, Wenyue and Li, Xiang and Zhu, Kaijie and Jin, Mingyu and Li, Lingyao and Ling, Haoyang and Chi, Jinkui and Wang, Jindong and Ma, Xin and others},
  journal={arXiv preprint arXiv:2403.01777},
  year={2024}
}

@inproceedings{hua2025disentangling,
  title={Disentangling logic: The role of context in large language model reasoning capabilities},
  author={Hua, Wenyue and Zhu, Kaijie and Li, Lingyao and Fan, Lizhou and Jin, Mingyu and Lin, Shuhang and Xue, Haochen and Li, Zelong and Wang, JinDong and Zhang, Yongfeng},
  booktitle={Findings of the Association for Computational Linguistics: ACL 2025},
  pages={19219--19242},
  year={2025}
}

@article{wu2024mind,
  title={Mind's eye of LLMs: visualization-of-thought elicits spatial reasoning in large language models},
  author={Wu, Wenshan and Mao, Shaoguang and Zhang, Yadong and Xia, Yan and Dong, Li and Cui, Lei and Wei, Furu},
  journal={Advances in Neural Information Processing Systems},
  volume={37},
  pages={90277--90317},
  year={2024}
}

@inproceedings{chen2024spatialvlm,
  title={Spatialvlm: Endowing vision-language models with spatial reasoning capabilities},
  author={Chen, Boyuan and Xu, Zhuo and Kirmani, Sean and Ichter, Brain and Sadigh, Dorsa and Guibas, Leonidas and Xia, Fei},
  booktitle={Proceedings of the IEEE/CVF Conference on Computer Vision and Pattern Recognition},
  pages={14455--14465},
  year={2024}
}

@inproceedings{yang2025thinking,
  title={Thinking in space: How multimodal large language models see, remember, and recall spaces},
  author={Yang, Jihan and Yang, Shusheng and Gupta, Anjali W and Han, Rilyn and Fei-Fei, Li and Xie, Saining},
  booktitle={Proceedings of the Computer Vision and Pattern Recognition Conference},
  pages={10632--10643},
  year={2025}
}

@article{li2025recognition,
  title={Recognition through Reasoning: Reinforcing Image Geo-localization with Large Vision-Language Models},
  author={Li, Ling and Zhou, Yao and Liang, Yuxuan and Tsung, Fugee and Wei, Jiaheng},
  journal={arXiv preprint arXiv:2506.14674},
  year={2025}
}

@article{li2025llms,
  title={LLMs as World Models: Data-Driven and Human-Centered Pre-Event Simulation for Disaster Impact Assessment},
  author={Li, Lingyao and Li, Dawei and Ou, Zhenhui and Xu, Xiaoran and Liu, Jingxiao and Ma, Zihui and Yu, Runlong and Deng, Min},
  journal={arXiv preprint arXiv:2506.06355},
  year={2025}
}

@inproceedings{zhou2024img2loc,
  title={Img2Loc: Revisiting image geolocalization using multi-modality foundation models and image-based retrieval-augmented generation},
  author={Zhou, Zhongliang and Zhang, Jielu and Guan, Zihan and Hu, Mengxuan and Lao, Ni and Mu, Lan and Li, Sheng and Mai, Gengchen},
  booktitle={Proceedings of the 47th international acm sigir conference on research and development in information retrieval},
  pages={2749--2754},
  year={2024}
}

@inproceedings{xu2024addressclip,
  title={Addressclip: Empowering vision-language models for city-wide image address localization},
  author={Xu, Shixiong and Zhang, Chenghao and Fan, Lubin and Meng, Gaofeng and Xiang, Shiming and Ye, Jieping},
  booktitle={European Conference on Computer Vision},
  pages={76--92},
  year={2024},
  organization={Springer}
}

@article{wang2024llmgeo,
  title={Llmgeo: Benchmarking large language models on image geolocation in-the-wild},
  author={Wang, Zhiqiang and Xu, Dejia and Khan, Rana Muhammad Shahroz and Lin, Yanbin and Fan, Zhiwen and Zhu, Xingquan},
  journal={arXiv preprint arXiv:2405.20363},
  year={2024}
}

@article{lacoste2023geo,
  title={Geo-bench: Toward foundation models for earth monitoring},
  author={Lacoste, Alexandre and Lehmann, Nils and Rodriguez, Pau and Sherwin, Evan and Kerner, Hannah and L{\"u}tjens, Bj{\"o}rn and Irvin, Jeremy and Dao, David and Alemohammad, Hamed and Drouin, Alexandre and others},
  journal={Advances in Neural Information Processing Systems},
  volume={36},
  pages={51080--51093},
  year={2023}
}

@inproceedings{pramanick2022world,
  title={Where in the world is this image? transformer-based geo-localization in the wild},
  author={Pramanick, Shraman and Nowara, Ewa M and Gleason, Joshua and Castillo, Carlos D and Chellappa, Rama},
  booktitle={European Conference on Computer Vision},
  pages={196--215},
  year={2022},
  organization={Springer}
}

@inproceedings{zhu2022transgeo,
  title={Transgeo: Transformer is all you need for cross-view image geo-localization},
  author={Zhu, Sijie and Shah, Mubarak and Chen, Chen},
  booktitle={Proceedings of the IEEE/CVF Conference on Computer Vision and Pattern Recognition},
  pages={1162--1171},
  year={2022}
}

@article{yi2025geolocsft,
  title={Geolocsft: Efficient visual geolocation via supervised fine-tuning of multimodal foundation models},
  author={Yi, Qiang and Shan, Lianlei},
  journal={arXiv preprint arXiv:2506.01277},
  year={2025}
}

@inproceedings{hays2008im2gps,
  title={IM2GPS: estimating geographic information from a single image},
  author={Hays, James and Efros, Alexei A},
  booktitle={2008 IEEE Conference on Computer Vision and Pattern Recognition},
  pages={1--8},
  year={2008},
  organization={IEEE},
  doi={10.1109/CVPR.2008.4587784}
}

@inproceedings{weyand2016planet,
  title={PlaNet - Photo Geolocation with Convolutional Neural Networks},
  author={Weyand, Tobias and Kostrikov, Ilya and Philbin, James},
  booktitle={European Conference on Computer Vision (ECCV)},
  pages={37--55},
  year={2016},
  publisher={Springer},
  doi={10.1007/978-3-319-46484-8_3}
}

@inproceedings{vo2017revisiting,
  title={Revisiting im2gps in the deep learning era},
  author={Vo, Nam and Jacobs, Nathan and Hays, James},
  booktitle={Proceedings of the IEEE international conference on computer vision},
  pages={2621--2630},
  year={2017}
}

@inproceedings{guo2023images,
  title={From images to textual prompts: Zero-shot visual question answering with frozen large language models},
  author={Guo, Jiaxian and Li, Junnan and Li, Dongxu and Tiong, Anthony Meng Huat and Li, Boyang and Tao, Dacheng and Hoi, Steven},
  booktitle={Proceedings of the IEEE/CVF conference on computer vision and pattern recognition},
  pages={10867--10877},
  year={2023}
}

@article{kuang2025natural,
  title={Natural language understanding and inference with mllm in visual question answering: A survey},
  author={Kuang, Jiayi and Shen, Ying and Xie, Jingyou and Luo, Haohao and Xu, Zhe and Li, Ronghao and Li, Yinghui and Cheng, Xianfeng and Lin, Xika and Han, Yu},
  journal={ACM Computing Surveys},
  volume={57},
  number={8},
  pages={1--36},
  year={2025},
  publisher={ACM New York, NY}
}

@inproceedings{li2024enhancing,
  title={Enhancing visual document understanding with contrastive learning in large visual-language models},
  author={Li, Xin and Wu, Yunfei and Jiang, Xinghua and Guo, Zhihao and Gong, Mingming and Cao, Haoyu and Liu, Yinsong and Jiang, Deqiang and Sun, Xing},
  booktitle={Proceedings of the IEEE/CVF Conference on Computer Vision and Pattern Recognition},
  pages={15546--15555},
  year={2024}
}

@inproceedings{hu2024mplug,
  title={mplug-paperowl: Scientific diagram analysis with the multimodal large language model},
  author={Hu, Anwen and Shi, Yaya and Xu, Haiyang and Ye, Jiabo and Ye, Qinghao and Yan, Ming and Li, Chenliang and Qian, Qi and Zhang, Ji and Huang, Fei},
  booktitle={Proceedings of the 32nd ACM International Conference on Multimedia},
  pages={6929--6938},
  year={2024}
}

@inproceedings{arandjelovic2016netvlad,
  title={NetVLAD: CNN architecture for weakly supervised place recognition},
  author={Arandjelovic, Relja and Gronat, Petr and Torii, Akihiko and Pajdla, Tomas and Sivic, Josef},
  booktitle={Proceedings of the IEEE conference on computer vision and pattern recognition},
  pages={5297--5307},
  year={2016}
}

@inproceedings{zhu2021vigor,
  title={Vigor: Cross-view image geo-localization beyond one-to-one retrieval},
  author={Zhu, Sijie and Yang, Taojiannan and Chen, Chen},
  booktitle={Proceedings of the IEEE/CVF Conference on Computer Vision and Pattern Recognition},
  pages={3640--3649},
  year={2021}
}

@inproceedings{radford2021learning,
  title={Learning transferable visual models from natural language supervision},
  author={Radford, Alec and Kim, Jong Wook and Hallacy, Chris and Ramesh, Aditya and Goh, Gabriel and Agarwal, Sandhini and Sastry, Girish and Askell, Amanda and Mishkin, Pamela and Clark, Jack and others},
  booktitle={International conference on machine learning},
  pages={8748--8763},
  year={2021},
  organization={PmLR}
}

@inproceedings{jia2021scaling,
  title={Scaling up visual and vision-language representation learning with noisy text supervision},
  author={Jia, Chao and Yang, Yinfei and Xia, Ye and Chen, Yi-Ting and Parekh, Zarana and Pham, Hieu and Le, Quoc and Sung, Yun-Hsuan and Li, Zhen and Duerig, Tom},
  booktitle={International conference on machine learning},
  pages={4904--4916},
  year={2021},
  organization={PMLR}
}

@inproceedings{li2023blip,
  title={Blip-2: Bootstrapping language-image pre-training with frozen image encoders and large language models},
  author={Li, Junnan and Li, Dongxu and Savarese, Silvio and Hoi, Steven},
  booktitle={International conference on machine learning},
  pages={19730--19742},
  year={2023},
  organization={PMLR}
}

@article{vivanco2023geoclip,
  title={Geoclip: Clip-inspired alignment between locations and images for effective worldwide geo-localization},
  author={Vivanco Cepeda, Vicente and Nayak, Gaurav Kumar and Shah, Mubarak},
  journal={Advances in Neural Information Processing Systems},
  volume={36},
  pages={8690--8701},
  year={2023}
}

@article{jay2025evaluating,
  title={Evaluating precise geolocation inference capabilities of vision language models},
  author={Jay, Neel and Nguyen, Hieu Minh and Hoang, Trung Dung and Haimes, Jacob},
  journal={arXiv preprint arXiv:2502.14412},
  year={2025}
}

@inproceedings{seo2018cplanet,
  title={Cplanet: Enhancing image geolocalization by combinatorial partitioning of maps},
  author={Seo, Paul Hongsuck and Weyand, Tobias and Sim, Jack and Han, Bohyung},
  booktitle={Proceedings of the European Conference on Computer Vision (ECCV)},
  pages={536--551},
  year={2018}
}

@article{tian2025nuscenes,
  title={Nuscenes-spatialqa: A spatial understanding and reasoning benchmark for vision-language models in autonomous driving},
  author={Tian, Kexin and Mao, Jingrui and Zhang, Yunlong and Jiang, Jiwan and Zhou, Yang and Tu, Zhengzhong},
  journal={arXiv preprint arXiv:2504.03164},
  year={2025}
}

@inproceedings{muller2018geolocation,
  title={Geolocation estimation of photos using a hierarchical model and scene classification},
  author={Muller-Budack, Eric and Pustu-Iren, Kader and Ewerth, Ralph},
  booktitle={Proceedings of the European conference on computer vision (ECCV)},
  pages={563--579},
  year={2018}
}

@inproceedings{clark2023we,
  title={Where we are and what we're looking at: Query based worldwide image geo-localization using hierarchies and scenes},
  author={Clark, Brandon and Kerrigan, Alec and Kulkarni, Parth Parag and Cepeda, Vicente Vivanco and Shah, Mubarak},
  booktitle={Proceedings of the IEEE/CVF Conference on Computer Vision and Pattern Recognition},
  pages={23182--23190},
  year={2023}
}

@article{shatwell2025gt,
  title={GT-Loc: Unifying When and Where in Images Through a Joint Embedding Space},
  author={Shatwell, David G and Dave, Ishan Rajendrakumar and Swetha, Sirnam and Shah, Mubarak},
  journal={arXiv preprint arXiv:2507.10473},
  year={2025}
}

@article{hou2024global,
  title={Global Streetscapes—A comprehensive dataset of 10 million street-level images across 688 cities for urban science and analytics},
  author={Hou, Yujun and Quintana, Matias and Khomiakov, Maxim and Yap, Winston and Ouyang, Jiani and Ito, Koichi and Wang, Zeyu and Zhao, Tianhong and Biljecki, Filip},
  journal={ISPRS Journal of Photogrammetry and Remote Sensing},
  volume={215},
  pages={216--238},
  year={2024},
  publisher={Elsevier}
}

@inproceedings{yan2023personalized,
  title={Personalized showcases: Generating multi-modal explanations for recommendations},
  author={Yan, An and He, Zhankui and Li, Jiacheng and Zhang, Tianyang and McAuley, Julian},
  booktitle={Proceedings of the 46th International ACM SIGIR Conference on Research and Development in Information Retrieval},
  pages={2251--2255},
  year={2023}
}

@inproceedings{tian2017cross,
  title={Cross-view image matching for geo-localization in urban environments},
  author={Tian, Yicong and Chen, Chen and Shah, Mubarak},
  booktitle={Proceedings of the IEEE Conference on Computer Vision and Pattern Recognition},
  pages={3608--3616},
  year={2017}
}

@article{wei2022chain,
  title={Chain-of-thought prompting elicits reasoning in large language models},
  author={Wei, Jason and Wang, Xuezhi and Schuurmans, Dale and Bosma, Maarten and Xia, Fei and Chi, Ed and Le, Quoc V and Zhou, Denny and others},
  journal={Advances in neural information processing systems},
  volume={35},
  pages={24824--24837},
  year={2022}
}

@article{brown2020language,
  title={Language models are few-shot learners},
  author={Brown, Tom and Mann, Benjamin and Ryder, Nick and Subbiah, Melanie and Kaplan, Jared D and Dhariwal, Prafulla and Neelakantan, Arvind and Shyam, Pranav and Sastry, Girish and Askell, Amanda and others},
  journal={Advances in neural information processing systems},
  volume={33},
  pages={1877--1901},
  year={2020}
}

\appendix

\section{Ethical Considerations}

Image geolocalization poses ethical risks related to privacy, surveillance, and misuse, as inferring locations from images can enable intrusive tracking or unintended disclosure of sensitive places. Although this study relies on publicly available or author-collected images and restricts access to external metadata, the demonstrated capabilities of LLMs underscore the need for responsible deployment. In addition, the observed geographic performance disparities raise concerns about fairness, as errors may disproportionately affect underrepresented regions. These considerations highlight the importance of bias auditing, uncertainty-aware reporting, and clear usage guidelines when applying geolocation systems in real-world settings.

\section{Statement of AI Assistant Use}

During the development of this paper, we use LLM Assistants (e.g., ChatGPT, Claude) for the following two main aspects: (i) Reference discovery: use the deep research tools from major providers to explore relevant work and literature. (ii) Code assistance: use coding agents to assist developing the code base of the current work. (iii) Grammar check: use LLMs to detect grammar errors in the drafty version of the paper, for better displaying our results.

\section{Sample Images from Benchmark Datasets}
\label{appendix:dataset}

\begin{figure*}[htbp]
  \centering
  \includegraphics[width=1\textwidth]{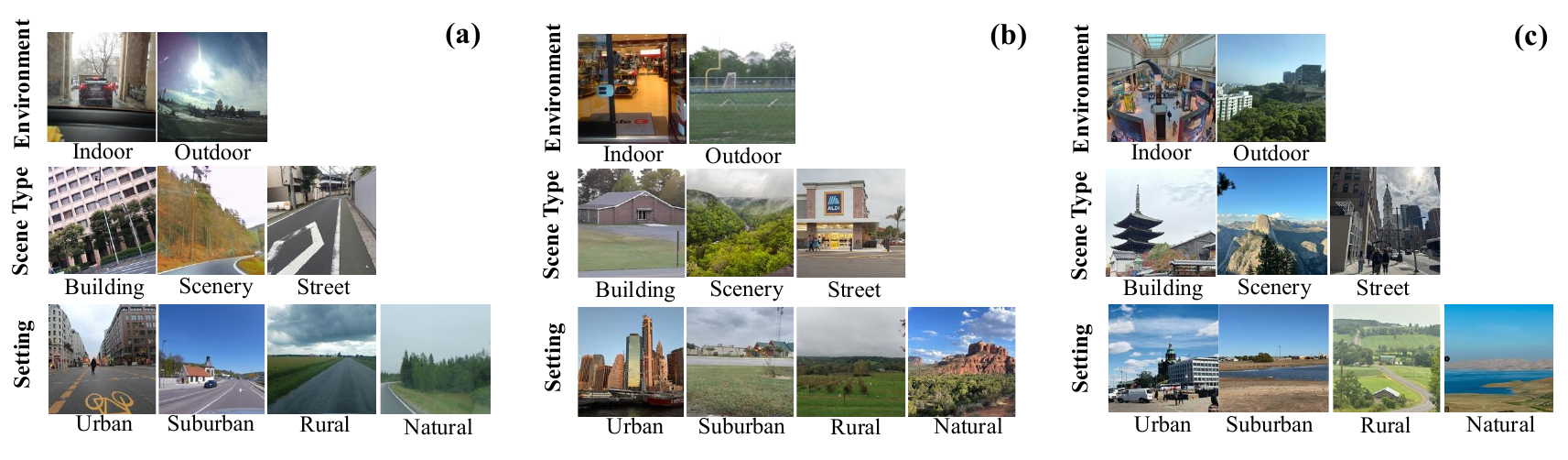}
  \vspace{-0.2cm}
  \caption{Sample images from the benchmark datasets: (a) Dataset-GSS; (b) Dataset-UPC; (c) Dataset-PCW.}
\vspace{-0.4cm}
  \label{fig:dataset-samples}
\end{figure*}

Sample images from our three datasets in the benchmark are presented in Figure~\ref{fig:dataset-samples}. 

\section{Image Data Distribution}
\label{appendix:distribution}

The sample distribution based on annotated features provided by the NUS team \cite{hou2024global} for Dataset-GSS is shown in Table~\ref{tab:gss_distribution}. Table~\ref{tab:feature_distribution} presents the distribution of features extracted from model outputs by gemini-2.5-pro across our benchmark datasets. Please note that these features are not human-annotated, as our goal is not to evaluate the influence of specific image attributes on LLMs’ geolocalization performance.

\begin{table}[h]
\caption{Sample distribution of Dataset-GSS.}
\label{tab:gss_distribution}
\centering
\footnotesize
\begin{tabular}{lcc}
\toprule
\textbf{Feature} & \textbf{Value} & \textbf{Count} \\
\midrule
lighting\_condition & day & 5601 \\
lighting\_condition & dusk/dawn & 319 \\
lighting\_condition & night & 232 \\
view\_direction & front/back & 5451 \\
view\_direction & side & 701 \\
weather & clear & 4638 \\
weather & cloudy & 1189 \\
weather & snowy & 175 \\
weather & rainy & 147 \\
weather & foggy & 3 \\
\bottomrule
\end{tabular}
\end{table}

\begin{table}[h]
\centering
\caption{Sample distribution based on gemini-pro-2.5.}
\vspace{-0.2cm}
\label{tab:feature_distribution}
\footnotesize
\begin{tabular}{lcccc}
\toprule
\textbf{Feature} & \textbf{Value} & \textbf{GSS} & \textbf{UPC} & \textbf{PCW} \\
\midrule
\multirow{3}{*}{environment} & outdoor & 6144 & 2400 & 264 \\
 & indoor & 2 & 522 & 4 \\
 & mixed & 5 & 5 & 2 \\
\midrule
\multirow{4}{*}{scene\_type} & street & 4636 & 120 & 33 \\
 & mixed & 1072 & 685 & 141 \\
 & scenery & 414 & 403 & 56 \\
 & building & 29 & 1719 & 40 \\
\midrule
\multirow{5}{*}{setting} & urban & 4350 & 595 & 149 \\
 & suburban & 998 & 1675 & 81 \\
 & rural & 755 & 373 & 8 \\
 & natural & 48 & 283 & 31 \\
 & mixed & 0 & 1 & 1 \\
\bottomrule
\end{tabular}
\end{table}

In addition, Dataset-GSS is constructed using stratified sampling to improve geographic diversity but its composition still reflects the availability patterns of large-scale street-level imagery platforms such as Mapillary. Consequently, some regions, particularly North America and Western Europe, remain more densely represented than parts of Africa and some areas of Asia. This imbalance is not unique to IMAGEO-Bench, but instead mirrors broader disparities in publicly available geo-tagged street-level imagery. To make the dataset composition transparent, we report its continent-level distribution and the most represented countries in Table~\ref{tab:gss_continent_distribution} and Table~\ref{tab:gss_country_distribution}.

\begin{table}[t]
\caption{Continent-level distribution of Dataset-GSS.}
\vspace{-0.2cm}
\label{tab:gss_continent_distribution}
\centering
\footnotesize
\begin{tabular}{lrr}
\toprule
Continent & Count & Percent \\
\midrule
Europe & 3201 & 52.0\% \\
Asia & 1229 & 20.0\% \\
North America & 1104 & 17.9\% \\
South America & 259 & 4.2\% \\
Oceania & 211 & 3.4\% \\
Africa & 148 & 2.4\% \\
\midrule
Total & 6152 & 100.0\% \\
\bottomrule
\end{tabular}
\end{table}

\begin{table}[t]
\caption{Top country-level distribution of Dataset-GSS. Percentages are reported relative to the full dataset ($N=6152$).}
\vspace{-0.2cm}
\label{tab:gss_country_distribution}
\centering
\footnotesize
\setlength{\tabcolsep}{4pt}
\begin{tabular}{lr@{\hspace{8pt}}lr}
\toprule
Country & \% & Country & \% \\
\midrule
United States & 12.5\% & Switzerland & 2.6\% \\
Germany & 11.1\% & Canada & 2.2\% \\
Japan & 7.3\% & Spain & 2.0\% \\
France & 7.2\% & Hungary & 2.0\% \\
Finland & 4.0\% & Chile & 1.6\% \\
Austria & 3.9\% & Moldova & 1.6\% \\
Russia & 3.8\% & Brazil & 1.5\% \\
Taiwan & 3.4\% & Malaysia & 1.3\% \\
Netherlands & 3.3\% & Azerbaijan & 1.3\% \\
Indonesia & 3.0\% & Other places & 21.4\% \\
Australia & 3.0\% &  &  \\
\bottomrule
\end{tabular}
\end{table}

As shown in Table~\ref{tab:gss_continent_distribution}, more than half of Dataset-GSS is drawn from Europe, followed by Asia and North America. At the country level (Table~\ref{tab:gss_country_distribution}), the United States and Germany are the most represented countries, with several Western European and East Asian countries also contributing notable shares. These statistics provide additional context for interpreting model performance and geospatial disparities observed on Dataset-GSS.

\section{Prompt Design}
\label{appendix:prompt}

The prompt design shown below (Figure~\ref{fig:prompt_design}) enables LLMs to perform image geolocalization from a given image. The prompt design shown below is guided by two key strategies: (i) chain-of-thought reasoning \cite{wei2022chain} and (ii) few-shot learner guidance \cite{brown2020language}. Each model is instructed to follow a structured reasoning process: first, analyze the visual content (e.g., landmarks, textual signage, and cultural indicators); then, apply chain-of-thought reasoning to integrate these observations; and finally, produce geocoded predictions accompanied by confidence scores.

\begin{figure*}[t]
\centering
\begin{tcolorbox}[colback=gray!5,colframe=blue!40!black,title=Prompt design for image geolocalization.]
\begin{minipage}[t]{0.48\textwidth}
\lstset{basicstyle=\scriptsize,breaklines=true,frame=none,breakatwhitespace=true}
\begin{lstlisting}
SYSTEM_PROMPT = "You are an AI assistant specialized in 
geocoding analysis from images."

IMAGE_LOCATION_PROMPT = 
You are given an image. Use a step-by-step (chain-of-thought) reasoning process to infer the most likely location.

Structure your response in three sections:
1. **Image Information** - Describe the general setting of the image.
2. **Reasoning** - Provide detailed justifications using observable visual cues.
3. **image geolocalization** - Output your best-guess location with a confidence score.

Your response must be valid JSON in this exact format:
{ "image_information": {
    "environment": "indoor|outdoor",
    "scene_type": "building|scenery|street|mixed",
    "setting": "urban|suburban|rural|natural"},
  "reasoning": {
    "landmark_recognition": "e.g., iconic structures, 
        architectural landmarks, natural features",
    "text_and_signage": "e.g., street signs, business 
        names, license plates, visible text",
    "cultural_indicators": "e.g., architectural styles, 
        regional patterns, cultural elements",
    "spatial_context": "e.g., geographic relationships, 
        infrastructure patterns"},
  "reverse_geocoding": {
    "confidence": "1|2|3|4|5",
    "address": {
      "street": "street address",
      "city": "city name",
      "state": "state/province", 
      "country": "country name"},
    
\end{lstlisting}
\end{minipage}
\hfill
\begin{minipage}[t]{0.48\textwidth}
\lstset{basicstyle=\scriptsize,breaklines=true,frame=none,breakatwhitespace=true}
\begin{lstlisting}
  "coordinates": {
      "latitude": "decimal degrees",
      "longitude": "decimal degrees"}}}

Example:
{ "image_information": {
    "environment": "outdoor",
    "scene_type": "building",
    "setting": "urban"},
  "reasoning": {
    "landmark_recognition": "Empire State Building 
        with distinctive Art Deco spire clearly visible",
    "text_and_signage": "NYC taxi markings 
        and New York license plates",
    "cultural_indicators": "Dense Manhattan urban 
        layout with characteristic skyscraper arrangement",
    "spatial_context": "Midtown Manhattan street grid 
        and building density patterns"},
  "reverse_geocoding": {
    "confidence": "5",
    "address": {
      "street": "350 5th Ave",
      "city": "New York",
      "state": "NY",
      "country": "USA"},
    "coordinates": {
      "latitude": "40.7484",
      "longitude": "-73.9857"}}}

Guidance:
- Return only valid JSON, no additional text
- Return a confidence score between 1 (low) and 5 (high)
- For image geolocalization, return only one 
  location that you are mostly confident
\end{lstlisting}
\end{minipage}
    \end{tcolorbox}
    \vspace{-2mm}
    \caption{Prompt used to elicit structured geolocation reasoning and prediction from LLMs.}
    \label{fig:prompt_design}
\end{figure*}

\section{Longitude Analysis on Datasets-GSS \& -UPC}
\label{appendix:longitude}

\begin{figure*}[htbp]
    \centering
    \begin{subfigure}{1\linewidth}
        \includegraphics[width=\linewidth]{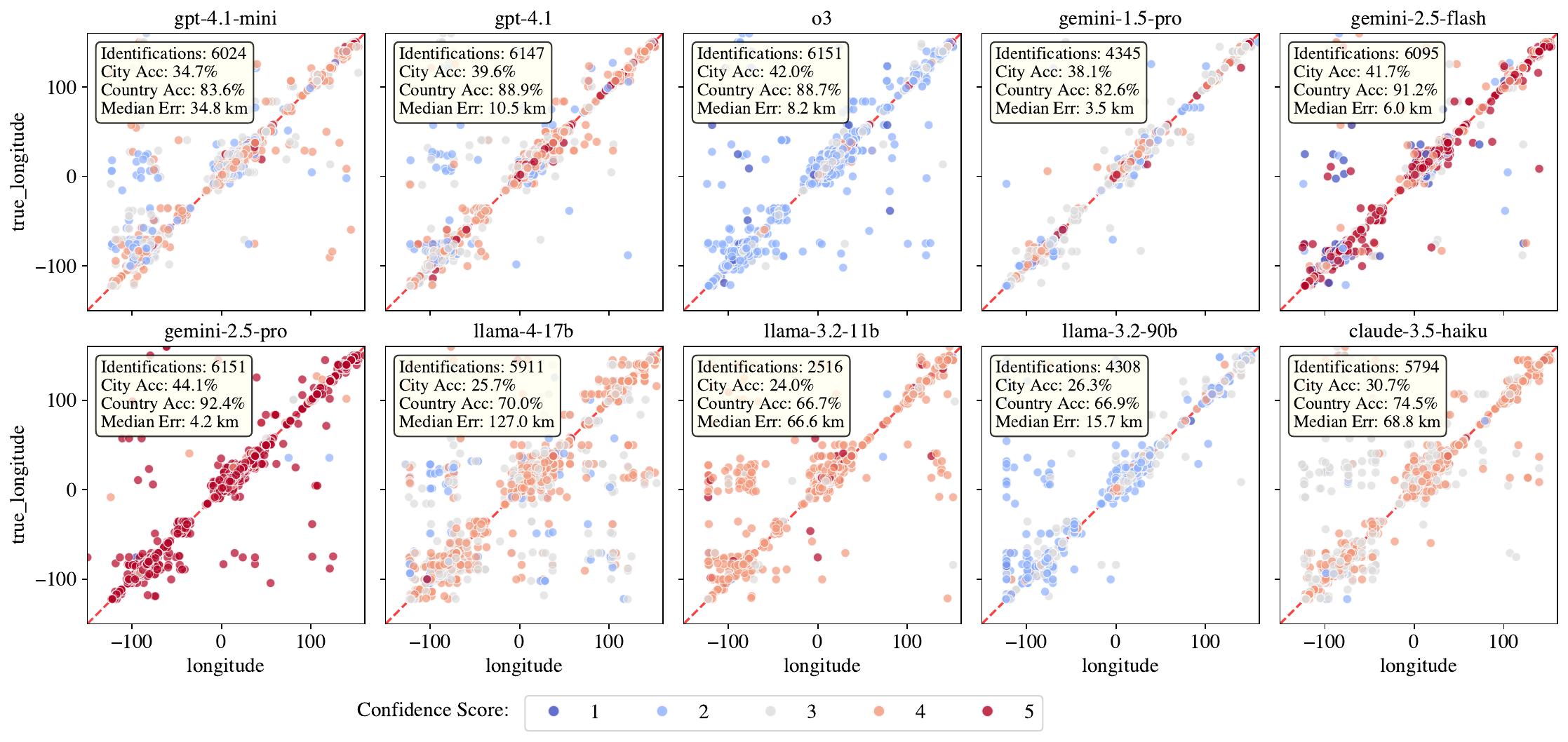}
        \hspace*{-55em}\raisebox{1em}{\small\textbf{(a)}}
    \end{subfigure}
    \begin{subfigure}{1\linewidth}
        \includegraphics[width=\linewidth]{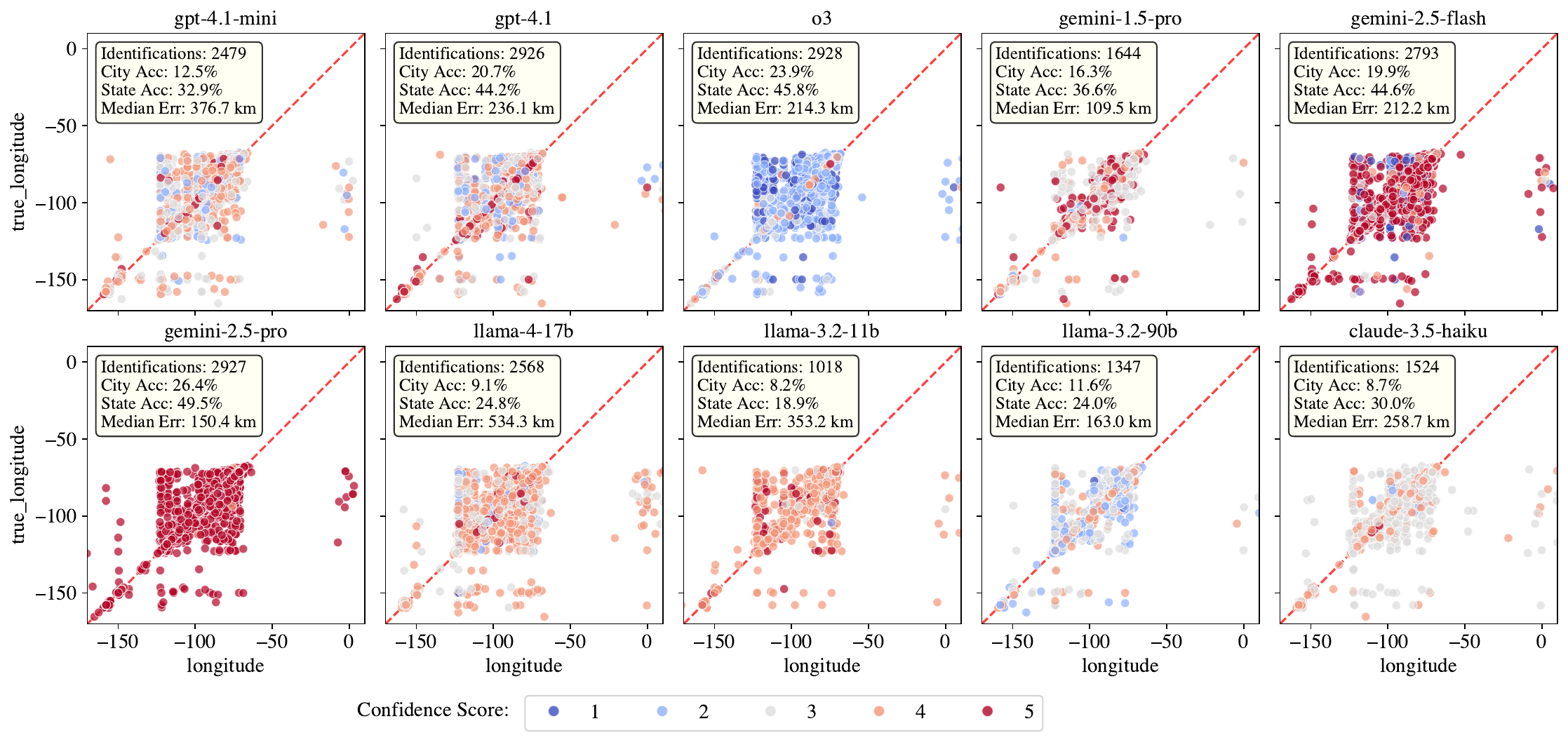}
        \hspace*{-55em}\raisebox{1em}{\small\textbf{(b)}}
    \end{subfigure}
    \vspace{-0.2cm}
    \caption{Benchmark performance based on longitude prediction. (a) Dataset-GSS, and (b) Dataset-UPC. Perfect predictions lie on the red dashed diagonal line. The proximity of points to this line indicates the model's accuracy, with the color of each point representing the model's self-reported confidence score on a scale of 1 to 5.}
    \label{fig:longitude_bench}
\end{figure*}

Figure~\ref{fig:longitude_bench} presents model performance on longitude prediction, complementing our earlier analysis of latitude. Overall trends are consistent: closed-source models generally achieve higher accuracy, while open-source models exhibit greater variance. In Dataset-GSS (Figure~\ref{fig:longitude_bench}(a)), most models produce predictions that align well with the ground-truth longitudes, as evidenced by tight clustering along the diagonal. Closed-source models such as gemini-2.5-pro, o3, and gpt-4.1 demonstrate strong alignment, whereas open-source models show more scattered predictions across the longitude range.

Despite the similarity with latitude analysis (Figure~\ref{fig:latitude_bench}, the longitude scatter plots reveal clear differences. For Dataset-UPC (Figure~\ref{fig:longitude_bench}(b)), predictions form a distinctive cube-like shape: both predicted (x-axis) and true (y-axis) longitudes fall between approximately $-120^\circ$ and $-66^\circ$, consistent with the full longitudinal extent of the U.S. continent. However, rather than aligning along the diagonal, predictions are densely concentrated in the middle of this range. This pattern is observed across most models, regardless of architecture or source. This pattern further suggests that, in the absence of clear visual cues, models often default to generic or centrally located longitude values. While such predictions typically fall within plausible U.S. bounds, they are frequently assigned to widely dispersed ground-truth locations. A likely explanation is that, unlike latitude—which can be inferred from environmental features such as vegetation, climate, or sun angle—longitude provides fewer visually distinctive signals across many U.S. scenes. Consequently, even high-performing models tend to produce centralized, non-specific longitude estimates when visual evidence is ambiguous.

\section{Benchmark Performance on Dataset-PCW}
\label{appendix:pcw}

Table~\ref{table:pcw_performance} shows the performance of models' geolocation prediction on unseen images from Dataset-PCW. The results reveal similar performance trends across models as observed with the other two datasets (see Table~\ref{table:performance}), with closed-source models maintaining an edge over open-source alternatives. This time, however, o3 achieves the highest city-level accuracy at 60.8\%, closely followed by gpt-4.1 at 60.0\%, while gemini-2.5-pro displays precision with the lowest median distance error of 0.7~km. In contrast, open-source models continue to lag behind: llama-4-17b and llama-3.2-11b reach only 40.8\% and 41.9\% city-level accuracy, respectively, with higher median errors of 34.3~km and 10.5~km.

Similarly, the computational cost analysis mirrors earlier patterns. While llama-3.2-11b offers the most cost-effective solution at \$0.00014 per request with only 783 average tokens, this efficiency comes at the expense of reduced accuracy. Conversely, o3 delivers superior performance but at a premium cost of \$0.01117 per request. Among the models, gemini-1.5-pro and gemini-2.5-flash show competitive performance (54.4\% city accuracy, 1.5~km median error) with relatively efficient resource use (1{,}180 tokens, \$0.00244 per request). These findings show that while larger models often perform better, the trade-off between model size, cost, and accuracy remains nuanced for practical use cases.

\begin{table}[ht]
\centering
\caption{LLM performance benchmark based on Dataset-PCW.}
\vspace{-0.2cm}
\small
\setlength{\tabcolsep}{5pt}
\begin{tabular}{lccccc}
    \toprule
    Model & City & Country & Median & Avg & Avg \\
     & Acc (\%) & Acc (\%) & Dist (km) & Tokens & Cost (\$) \\
    \midrule
    gpt-4.1-mini & 51.8\% & 95.3\% & 2.9 & 3041 & .00153 \\
    gpt-4.1 & 60.0\% & \textbf{97.7}\% & 1.4 & 1621 & .00476 \\
    o3 & \textbf{60.8\%} & 96.9\% & 0.9 & 2350 & .01117 \\
    gemini-1.5-pro & 54.4\% & 93.3\% & 1.5 & 1180 & .00244 \\
    gemini-2.5-flash & 50.2\% & 96.3\% & 1.9 & 1346 & .00133 \\
    gemini-2.5-pro & 58.7\% & 97.0\% & \textbf{0.7} & 1345 & .00534 \\
    llama-4-17b & 40.8\% & 93.4\% & 34.3 & 2823 & .00089 \\
    llama-3.2-11b & 41.9\% & 89.4\% & 10.5 & \textbf{783} & \textbf{.00014} \\
    llama-3.2-90b & 50.0\% & 89.7\% & 1.9 & 6848 & .00822 \\
    claude-3.5-haiku & 42.0\% & 94.1\% & 10.0 & 2350 & .00266 \\
    \bottomrule
\end{tabular}
\label{table:pcw_performance}
\end{table}





\section{Extended Feature Attribution Results}
\label{app:extended_attr}

We perform a post-hoc diagnostic to quantify how well each model’s own extracted semantic fields predict its success or failure. For each model, we regress its city-level prediction correctness (binary) and its log-transformed distance error (continuous) against these four fields using logistic and ridge regression. Results are reported on all three benchmark splits.

\begin{table}[ht]
  \centering
  \caption{Field-driven performance on \textbf{Dataset-GSS}.}
  \vspace{-0.2cm}
  \label{tab:d1-summary}
  \begin{tabular}{lrrrr}
    \toprule
    Model & Accuracy $\uparrow$ & AUC $\uparrow$ & MAE$_{\log}\,\downarrow$ & $R^2\,\uparrow$ \\
    \midrule
    gpt-4.1-mini      & 0.691 & 0.736 & 0.761 & 0.102 \\
    gpt-4.1           & 0.680 & 0.753 & 0.644 & 0.133 \\
    o3                & 0.707 & 0.777 & 0.633 & 0.141 \\
    gemini-1.5-pro    & 0.635 & 0.682 & 0.585 & 0.099 \\
    gemini-2.5-flash  & 0.699 & 0.760 & 0.575 & 0.165 \\
    gemini-2.5-pro    & 0.691 & 0.742 & 0.547 & 0.171 \\
    llama-4-17b       & 0.733 & 0.738 & 1.008 & 0.091 \\
    llama-3.2-11b     & 0.667 & 0.617 & 1.013 & 0.085 \\
    llama-3.2-90b     & 0.647 & 0.667 & 0.763 & 0.108 \\
    claude-3.5-haiku  & 0.676 & 0.747 & 0.836 & 0.140 \\
    \bottomrule
  \end{tabular}
\end{table}

\begin{table}[ht]
  \centering
  \caption{Field-driven performance on \textbf{Dataset-UPC}.}
  \vspace{-0.2cm}
  \label{tab:d2-summary}
  \begin{tabular}{lrrrr}
    \toprule
    Model & Accuracy $\uparrow$ & AUC $\uparrow$ & MAE$_{\log}\,\downarrow$ & $R^2\,\uparrow$ \\
    \midrule
    gpt-4.1-mini      & 0.852 & 0.695 & 0.926 & 0.083 \\
    gpt-4.1           & 0.793 & 0.739 & 0.908 & 0.177 \\
    o3                & 0.778 & 0.737 & 0.920 & 0.176 \\
    gemini-1.5-pro    & 0.755 & 0.681 & 0.834 & 0.030 \\
    gemini-2.5-flash  & 0.801 & 0.685 & 0.888 & 0.129 \\
    gemini-2.5-pro    & 0.759 & 0.687 & 0.874 & 0.159 \\
    llama-4-17b       & 0.897 & 0.686 & 1.030 & 0.053 \\
    llama-3.2-11b     & 0.858 & 0.666 & 1.015 & 0.071 \\
    llama-3.2-90b     & 0.759 & 0.689 & 1.029 & 0.087 \\
    claude-3.5-haiku  & 0.842 & 0.713 & 0.942 & 0.051 \\
    \bottomrule
  \end{tabular}
\end{table}


 \begin{figure}[ht]
  \centering
\includegraphics[width=\linewidth]{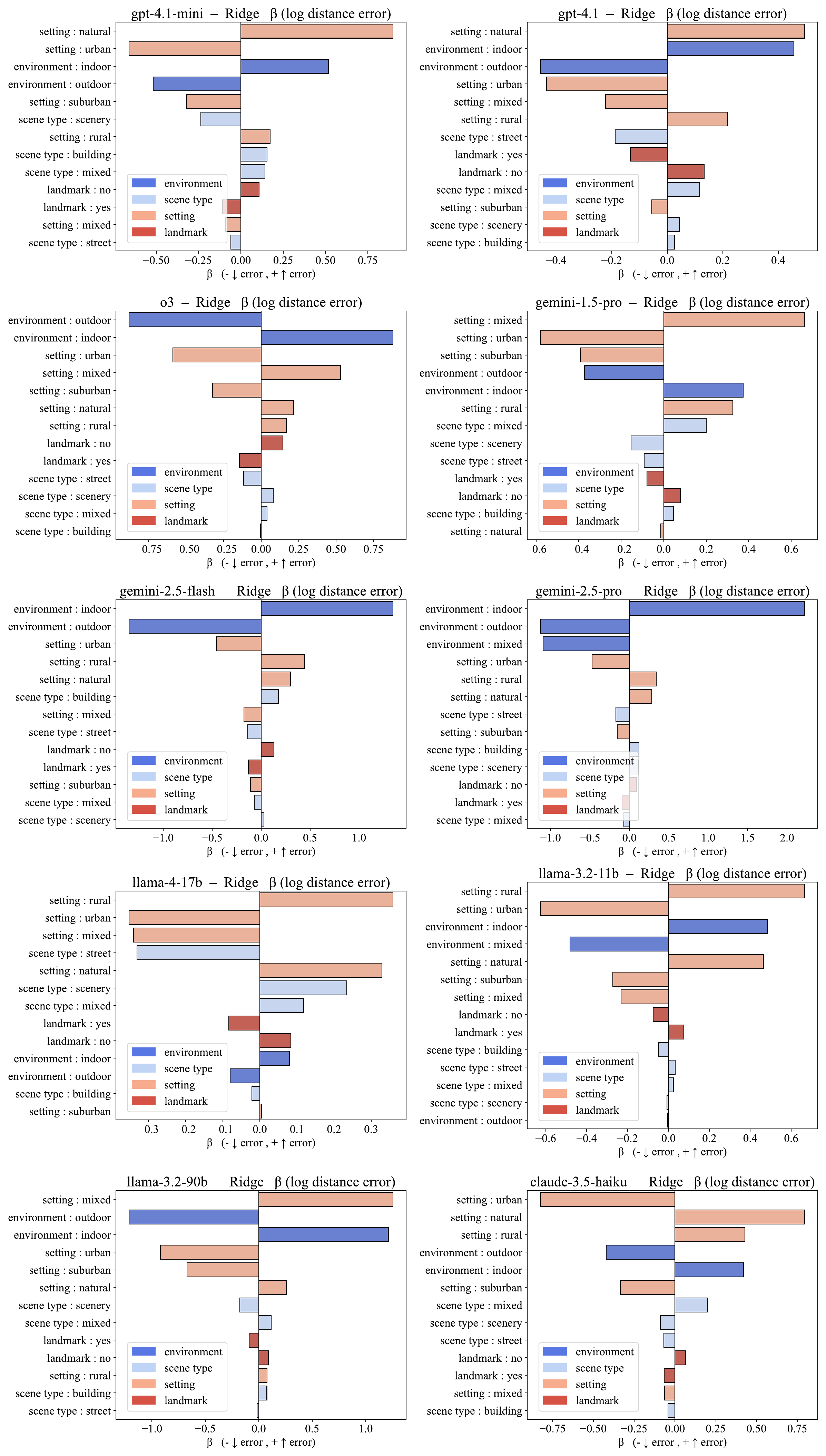}
  \caption{Ridge regression coefficients on Dataset-GSS, using log-distance error as the response. Negative values indicate cues that decrease error; positive values indicate cues that increase error.}
  \vspace{-0.2cm}
  \label{fig:dataset1 reasoning - dist-err}
\end{figure}

\begin{table}[ht]
  \centering
  \caption{Field-driven performance on \textbf{Dataset-PCW}.}
  \vspace{-0.2cm}
  \label{tab:d3-summary}
  \begin{tabular}{lrrrr}
    \toprule
    Model & Accuracy $\uparrow$ & AUC $\uparrow$ & MAE$_{\log}\,\downarrow$ & $R^2\,\uparrow$ \\
    \midrule
    gpt-4.1-mini      & 0.741 & 0.765 & 0.735 & 0.200 \\
    gpt-4.1           & 0.727 & 0.756 & 0.551 & 0.282 \\
    o3                & 0.742 & 0.770 & 0.541 & 0.271 \\
    gemini-1.5-pro    & 0.703 & 0.750 & 0.477 & 0.089 \\
    gemini-2.5-flash  & 0.757 & 0.790 & 0.630 & 0.244 \\
    gemini-2.5-pro    & 0.752 & 0.787 & 0.559 & 0.238 \\
    llama-4-17b       & 0.684 & 0.770 & 0.951 & 0.119 \\
    llama-3.2-11b     & 0.676 & 0.737 & 0.959 & 0.104 \\
    llama-3.2-90b     & 0.771 & 0.824 & 0.495 & 0.389 \\
    claude-3.5-haiku  & 0.766 & 0.833 & 0.783 & 0.217 \\
    \bottomrule
  \end{tabular}
\end{table}


This diagnostic analysis reveals how effectively current models ground their geolocation predictions in structured semantic cues, offering insight into their underlying visual-geospatial reasoning capabilities. On Dataset-GSS, which primarily features outdoor urban scenes rich in spatial context, models exhibit stronger alignment between their predicted semantic fields and their final performance. Regression models based on these four elements (environment, scene type, setting, and landmark) explain up to 17\% of the variance in log-distance error ($R^2$), with gemini-2.5-pro consistently leading in both interpretability and precision. Its ability to generate semantically rich, spatially aligned tags translates into the lowest MAE$_{\log}$ and the highest $R^2$, suggesting a high degree of internal consistency between scene understanding and coordinate prediction. In contrast, although llama-4-17b achieves the highest city-level classification accuracy on both Datasets-GSS and -UPC, its large distance errors indicate that these correct predictions may stem from over-reliance on salient, shortcut features---such as recognizable logos or architectural patterns---without engaging in deeper spatial reasoning or refinement.

\begin{figure}[t]
  \centering
\includegraphics[width=\linewidth]{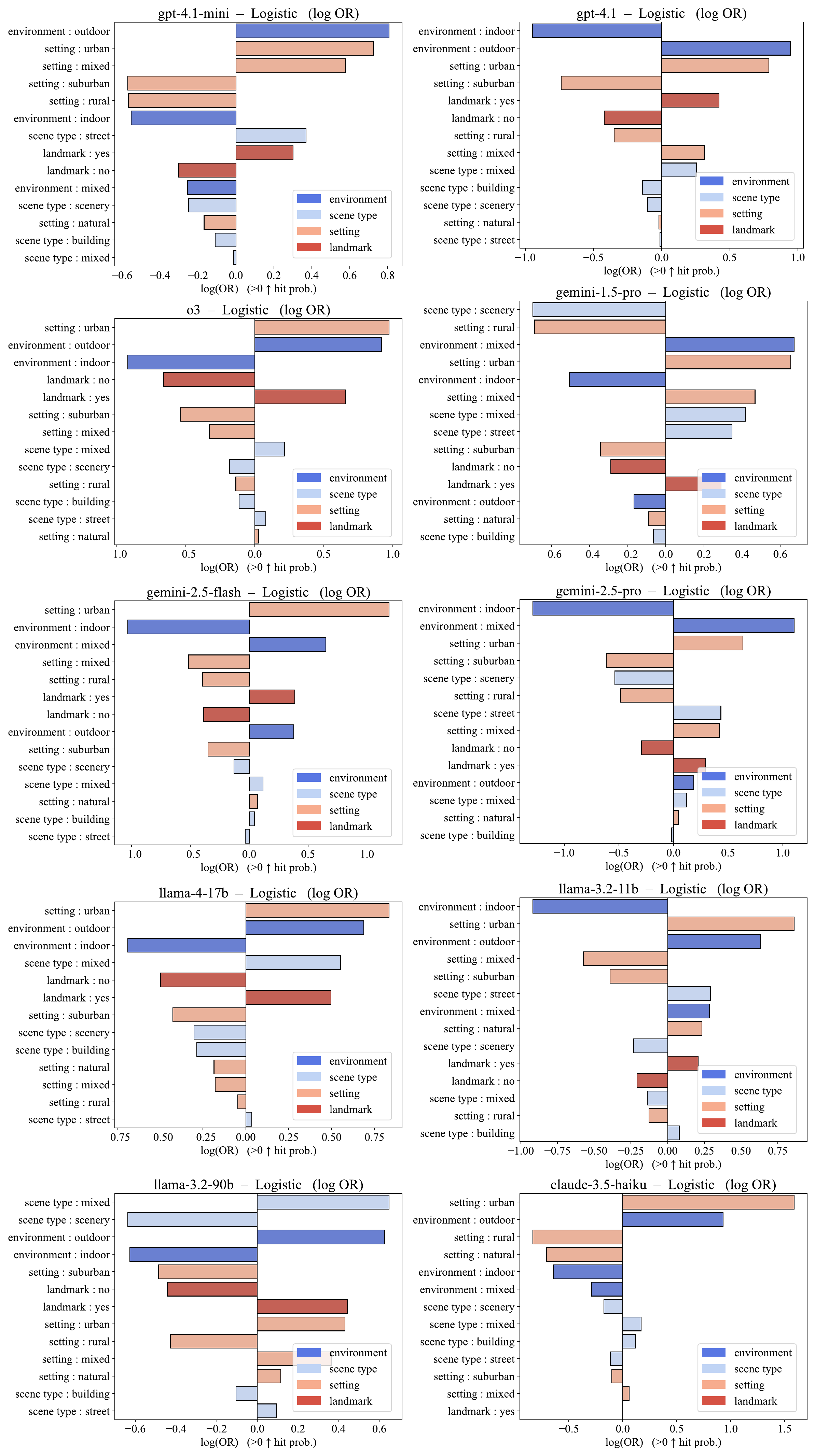}
\caption{Log-odds coefficients from diagnostic logistic regression on Dataset-UPC. Positive values increase the likelihood of correct city prediction; negative values decrease it. }
  \vspace{-0.2cm}
  \label{fig:dataset2 reasoning - city-level}
\end{figure}

The diagnostic further underscores the challenges posed by visually sparse or ambiguous environments. On Dataset-UPC, where scenes (e.g., indoors, signage fragments) often lack distinct geographic structures, field-based regressions degrade sharply across all models. In particular, llama-3.2-90b demonstrates that large open-weight models can internalize meaningful spatial priors when appropriately scaled and tuned, achieving high accuracy and strong $R^2$ in PCW. Still, the overall explanatory power of these four fields remains limited: Even in the best case, the variance in distance errors over 60\% remains unaccounted for. This gap suggests that high-level tags like ``rural'' or ``landmark'' only scratch the surface of the complex reasoning chain needed for accurate localization. Finer-grained semantics, such as regional architecture, vegetation type, textual cues, and spatial layout, are likely necessary to capture how LLMs reason about place fully.

\begin{figure}[t]
  \centering
  \includegraphics[width=\linewidth]{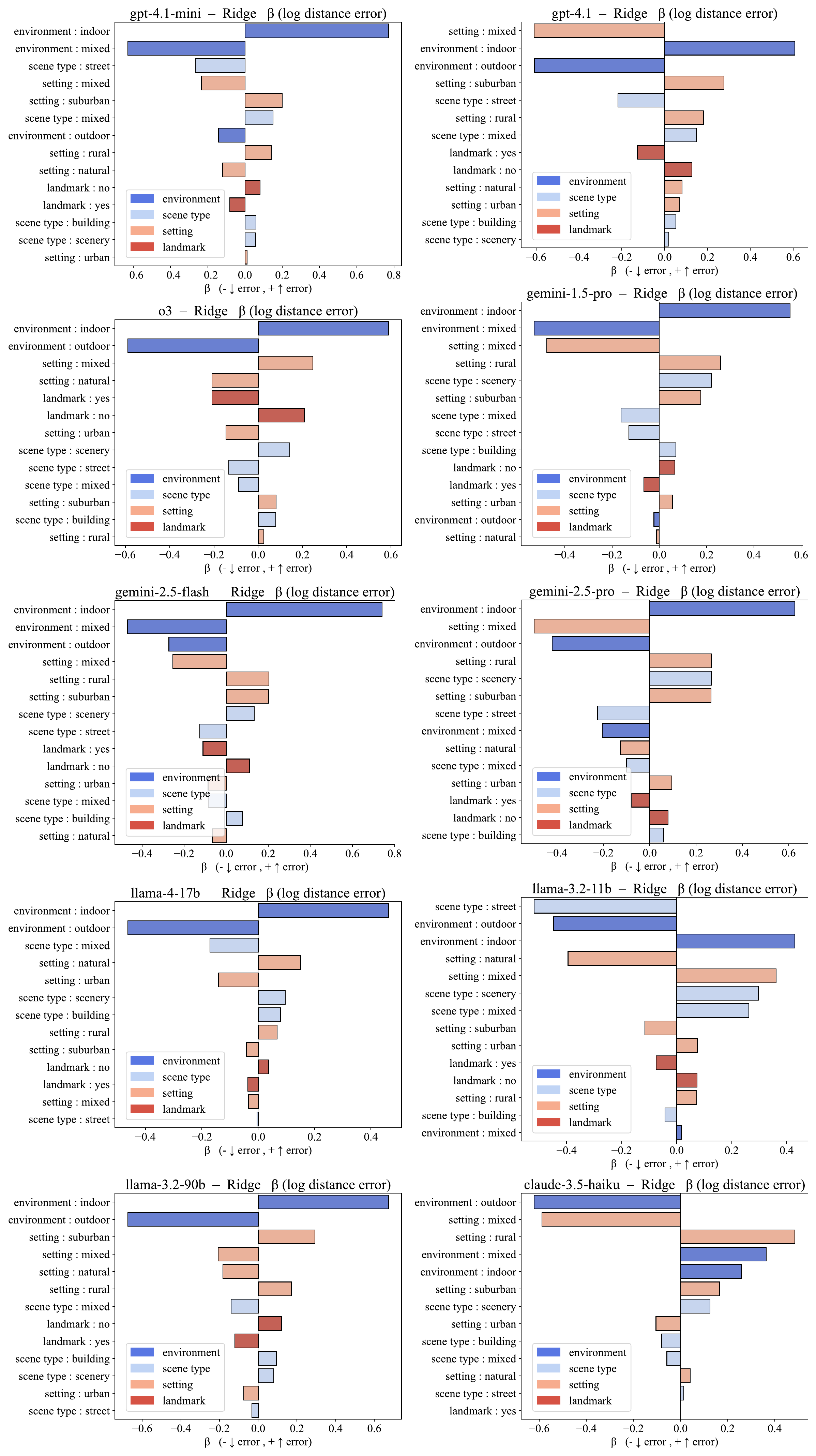}
   \caption{Ridge regression coefficients on Dataset-UPC, using log-distance error as the response. Negative values indicate cues that decrease error; positive values indicate cues that increase error.}
     \vspace{-0.2cm}
  \label{fig:dataset2 reasoning - dist-err}
\end{figure}

Figure \ref{fig:dataset1 reasoning - dist-err} displays the ridge-regression coefficients for the four semantic fields on Dataset-GSS, with the log-distance error as the dependent variable. Each bar shows the directional contribution of a cue: larger negative values (left of the axis) correspond to stronger error reduction, whereas positive values indicate error amplification. Among all cues, setting and environment are the most influential. Outdoor scenes receive the most negative weights, confirming that external spatial anchors (skylines, road geometry, and signage) can substantially improve geolocalization prediction. By contrast, indoor scenes attract large positive weights (notably for gemini-2.5-pro), underscoring how precision degrades once exterior context is lost. The setting field exhibits a similar polarity: urban imagery lowers error, whereas rural or natural settings raise it. This likely reflects biases in several models training data, which may overrepresent urban environments and their characteristic spatial patterns such as grid layouts or dense built form.

\begin{figure}[t]
  \centering
\includegraphics[width=\linewidth]{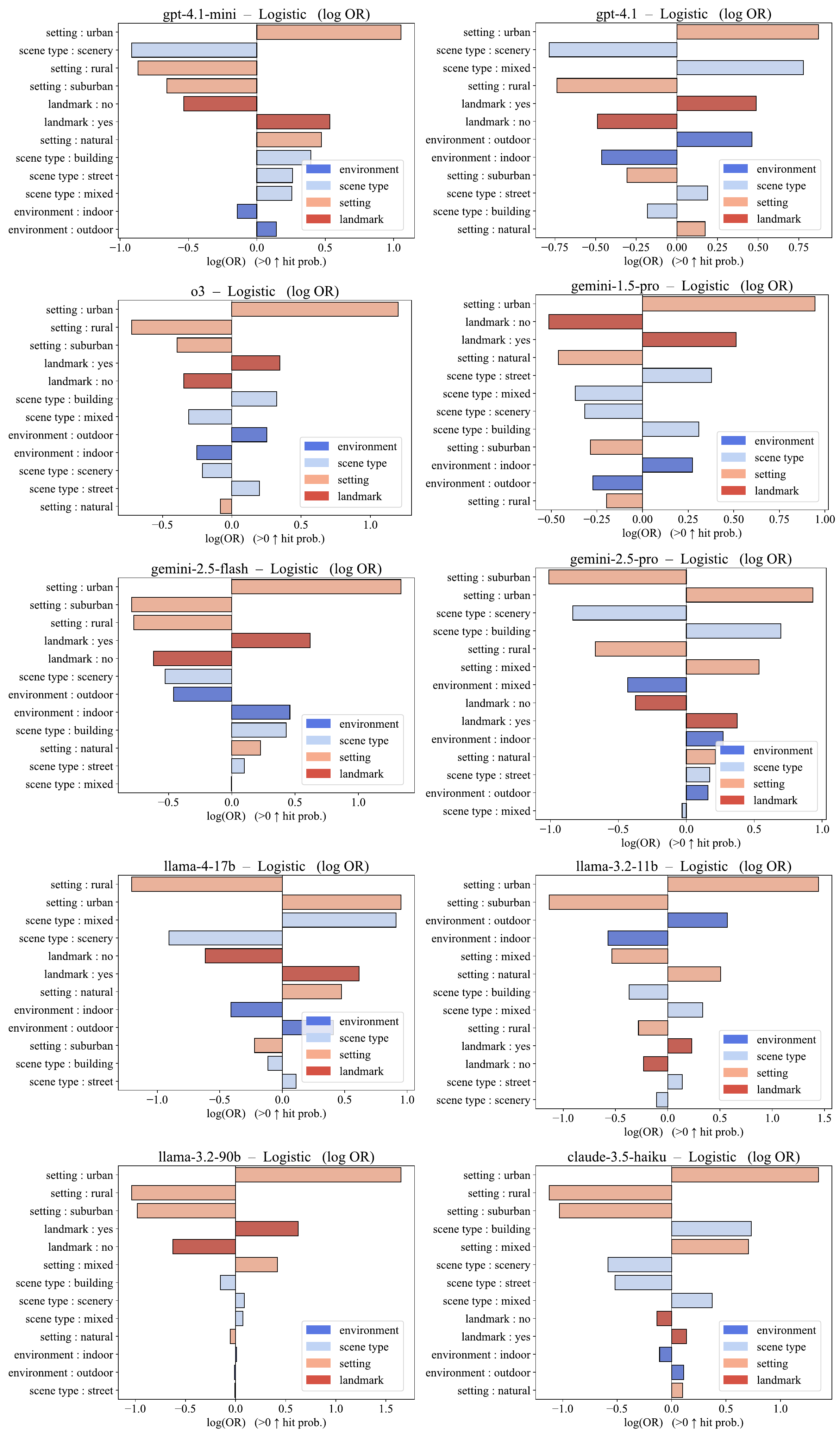}
   \caption{Log-odds coefficients from diagnostic logistic regression on Dataset-PCW. Positive values increase the likelihood of correct city prediction; negative values decrease it. }
    \vspace{-0.2cm}
  \label{fig:dataset3 reasoning - city-level}
\end{figure}

\begin{figure}[t]
  \centering
\includegraphics[width=\linewidth]{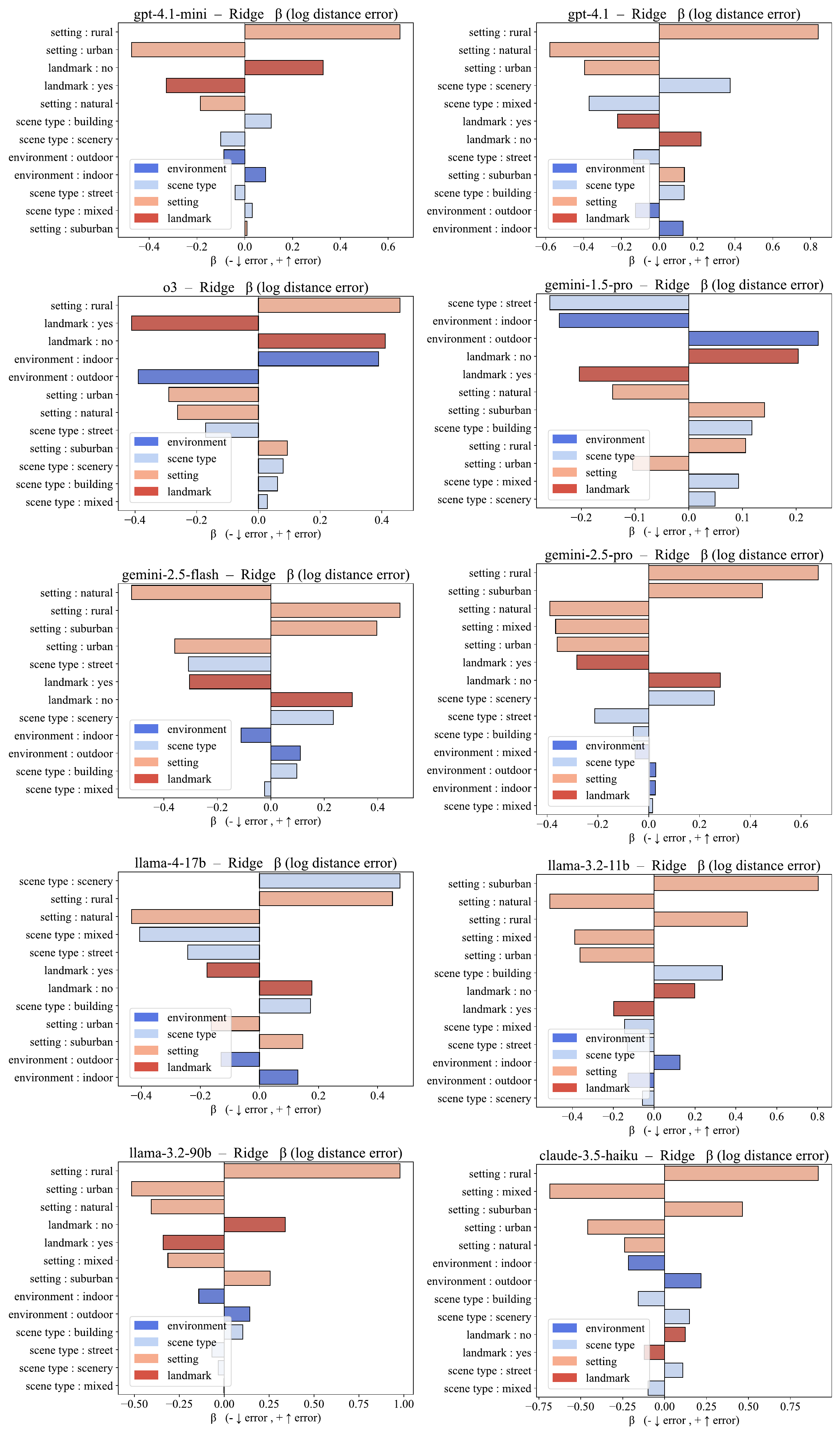}
    \caption{Ridge regression coefficients on Dataset-PCW, using log-distance error as the response. Negative values indicate cues that decrease error; positive values indicate cues that increase error.}
    \vspace{-0.4cm}
  \label{fig:dataset3 reasoning - dist-err}
\end{figure}

Figures~\ref{fig:dataset2 reasoning - city-level}–\ref{fig:dataset3 reasoning - dist-err} extend this analysis to Dataset-UPC and Dataset-PCW. On UPC, the predictive power of semantic fields declines sharply: $R^2$ values drop and landmark coefficients collapse toward zero, indicating reduced access to external anchors. On PCW, open-weight models like llama-3.2-90b show strong negative weights for urban and outdoor cues (corresponding to the highest $R^2$ on this split), but also greater variance, with large positive weights for ambiguous scene types such as buildings or scenery. In contrast, proprietary models like gemini-2.5-pro and o3 adopt flatter, more conservative profiles, yielding fewer high-confidence errors. Overall, model performance peaks when scenes are urban, outdoor, and landmark-rich, but degrades rapidly under distribution shift.

\begin{figure*}[t]
  \centering
\includegraphics[width=1\linewidth]{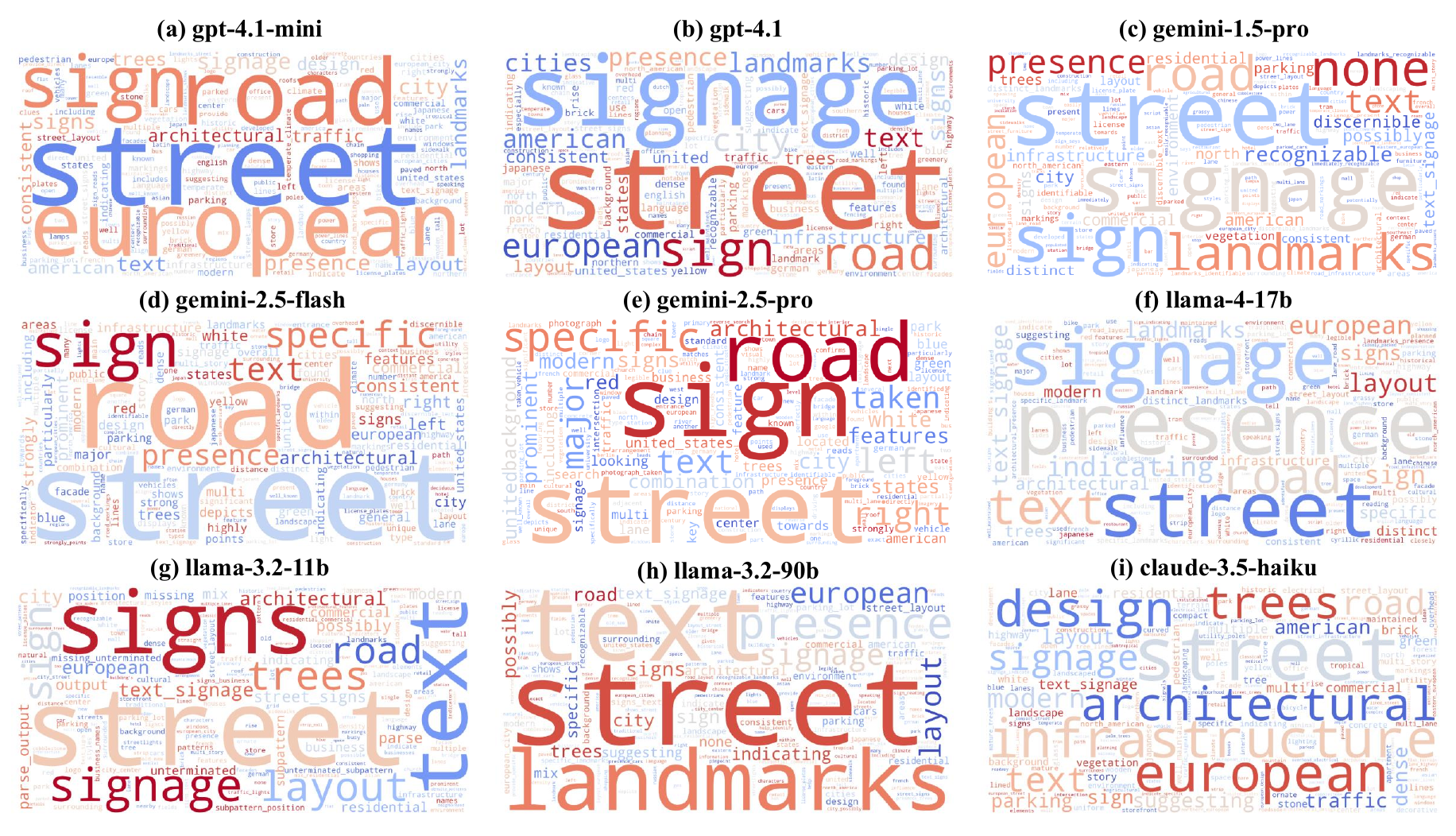}%
  \caption{Mixed unigram–bigram word cloud derived from the reasoning fields: (a) gpt-4.1-mini, (b) gpt-4.1, (c) gemini-1.5-pro, (d) gemini-2.5-flash, (e) gemini-2.5-pro, (f) llama-4-17b, (g) llama-3.2-11b, (h) llama-3.2-90b, (i) claude-3.5-haiku.}%
\label{fig:wordcloud}
\end{figure*}

\section{Extended Word-Cloud Analysis}
\label{app:wordclouds}

Figure~\ref{fig:wordcloud} presents the word clouds for the nine remaining models. First, most models rely on traffic-related terms such as ``sign,'' ``street,'' and ``road.'' This common vocabulary highlights a general model preference for OCR-visible and regulation-focused features, especially both gpt-4.1 and gpt-4.1-mini prominently feature these traffic cues. The gemini models similarly emphasize precise geolocation cues but differ notably in style: gemini-1.5-pro uniquely includes the placeholder ``none,'' explicitly signaling missing visual evidence, whereas gemini-2.5-flash and gemini-2.5-pro adopt a more concise, instruction-oriented vocabulary with words such as ``specific'' and ``taken.'' In contrast, the llama models predominantly rely on vague tokens including ``text,'' ``layout,'' and ``presence,'' with llama-3.2-11b occasionally producing debug-like outputs and llama-90b displaying some spatial awareness via the terms ``street'', ``landmarks.'' claude-3.5-haiku distinctly diverges from the other models by reducing its dependence on traffic terms, instead emphasizing scene descriptors such as ``architectural,'' ``trees,'' and ``design.'' Collectively, these lexical patterns align with the quantitative results presented earlier: most models rely heavily on abundant, discrete, and regulative visual cues.

\section{Text-Masking Ablation: Contextual Geospatial Reasoning}
\label{appendix:ablation}

To further examine the extent to which localization performance depends on explicit textual cues in the image, we conduct a controlled text-masking ablation. Specifically, we detect visible textual regions, including signage, storefront names, and license plates, and compare model performance on images with and without these cues. We report the resulting changes in city-level accuracy and median distance error for two representative models, o3 and gemini-2.5-pro, as shown in Table~\ref{tab:text_masking_ablation}.

\begin{table*}[ht] 
\caption{Text-masking ablation results showing the effect of removing visible textual cues on geolocalization performance.}
\label{tab:text_masking_ablation}
\centering
\footnotesize
\begin{tabular}{llcccccc}
\toprule
Dataset & Model & \begin{tabular}[c]{@{}c@{}}City Acc\\ (with text)\end{tabular} & \begin{tabular}[c]{@{}c@{}}City Acc\\ (w/o text)\end{tabular} & \begin{tabular}[c]{@{}c@{}}$\Delta$ City\\ Acc\end{tabular} & \begin{tabular}[c]{@{}c@{}}Median Dist\\ (with text)\end{tabular} & \begin{tabular}[c]{@{}c@{}}Median Dist\\ (w/o text)\end{tabular} & \begin{tabular}[c]{@{}c@{}}$\Delta$ Median\\ Dist\end{tabular} \\
\midrule
GSS & o3              & 46.6\% & 24.0\% & $-22.6$pp & 3.7 km   & 186.9 km & +183.2 km \\
GSS & gemini-2.5-pro  & 48.8\% & 21.0\% & $-27.8$pp & 2.6 km   & 110.6 km & +108.1 km \\
UPC & o3              & 26.1\% & 18.9\% & $-7.2$pp  & 192.3 km & 248.7 km & +56.3 km \\
UPC & gemini-2.5-pro  & 27.9\% & 21.4\% & $-6.4$pp  & 146.3 km & 167.1 km & +20.8 km \\
\bottomrule
\end{tabular}

\end{table*}

The results reveal a clear contrast across datasets. On Dataset-GSS, masking textual cues leads to a substantial decline in performance, with city-level accuracy dropping by 22--28\% and median distance error increasing by more than 100 km for both models. This pattern indicates that performance on global outdoor street imagery depends heavily on language-based anchors embedded in the visual scene. In contrast, the performance drop on Dataset-UPC is much smaller, with city-level accuracy decreasing by only about 6--7 percentage points and median distance error increasing more modestly. This suggests that when textual cues are weaker, less geographically distinctive, or less informative for fine-grained localization, model performance is less dominated by OCR-like recognition and remains constrained by broader contextual and geospatial ambiguity.

This contrast is also consistent with the characteristics of the two datasets. In Dataset-GSS, signage language, place names, and other visible text often directly encode city- or country-level information. By comparison, Dataset-UPC contains many indoor or brand-heavy POI images within a single country, where text is frequently generic, such as chain names or common business labels, and therefore contributes less incremental geographic signal.

\section{Sample Output by LLMs}
\label{appendix:output}

Tables~\ref{tab:sample_gss}--\ref{tab:sample_pcw} present sampled outputs from o3 and gemini-2.5-pro across our benchmark datasets. Each sample includes the model's geolocation prediction along with its corresponding reasoning, structured according to our standardized reasoning template. The ``ground truth'' column provides the correct location for comparison. The tables feature both correct and incorrect predictions to illustrate the strengths and failure cases of the models.

We observe consistent model behaviors. When images contain clear textual or architectural anchors, both o3 and gemini-2.5-pro achieve high accuracy. Notable examples include gemini-2.5-pro localizing Jakarta via an Indonesian traffic sign (Table~\ref{tab:sample_gss}, row~1), identifying a Seattle-area freeway from overhead signage (row~3), and pinpointing Riga from a Latvian facade (row~5). It also correctly resolves the ``American Clock \& Watch Museum'' in Bristol, CT (Table~\ref{tab:sample_upc}, row~2). Likewise, o3 infers the 7th arrondissement in Paris from French café signage (Table~\ref{tab:sample_gss}, row~2). However, this confidence often carries into near-misses: o3 misattributes Philadelphia’s City Hall to Honolulu (Table~\ref{tab:sample_upc}, row~1), while gemini-2.5-pro maps a Mexican coastal bar to New York City (Table~\ref{tab:sample_pcw}, row~6), both with confidence 4–5—highlighting a bias toward visually salient urban hubs. In text-sparse scenes, their strategies diverge: gemini-2.5-pro explicitly notes missing text (e.g., Table~\ref{tab:sample_pcw}, row~2), while o3 defaults to stylistic heuristics, referencing “Mediterranean tiles” in Fort Worth (Table~\ref{tab:sample_upc}, row~6) or “Midwest campus” traits in Duluth (Table~\ref{tab:sample_pcw}, row~4). Illumination also affects performance: gemini-2.5-pro correctly predicts Mackinaw City at night using bridge geometry and aurora cues (Table~\ref{tab:sample_pcw}, row~2), whereas o3 falters under low contrast in Lubbock (row~3). Both models struggle most in visually repetitive rural or coastal scenes, with o3 drifting from Luxembourg to Germany (Table~\ref{tab:sample_gss}, row~7) and gemini-2.5-pro misplacing a Mexican inlet to Florida (Table~\ref{tab:sample_pcw}, row~7), relying on vague terms like ``grassy embankment'' or ``sandy shoreline.'' 


\begin{table*}[ht]
\centering
\caption{Representative LLM image geolocalization output in Dataset-GSS.}
{\scriptsize
\begin{tabular}{|p{3cm}|p{2cm}|p{11cm}|}
\hline
\textbf{Image} & \textbf{Ground-Truth} & \textbf{LLM Output} \\
\hline

\raisebox{-\totalheight}{\includegraphics[width=0.15\textwidth]{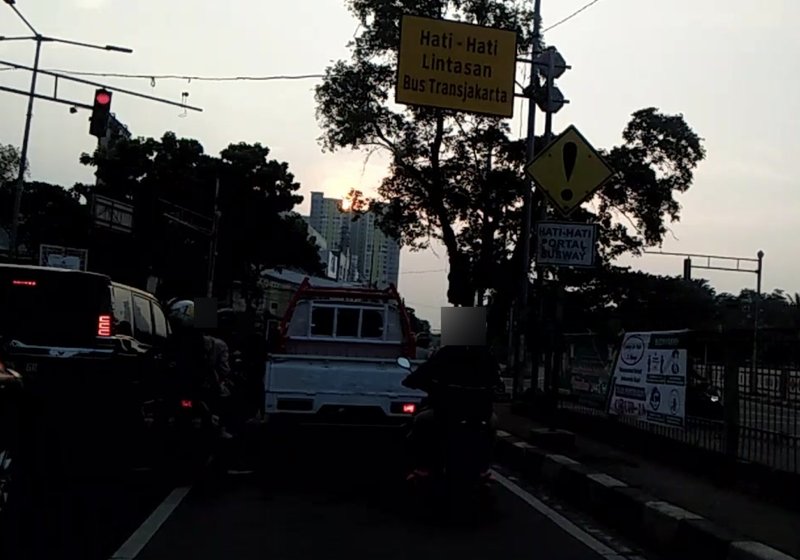}} & \textbf{Address}: Jakarta, Indonesia \textbf{Coordinates}: -6.2088, 106.8459 & \textbf{Model: gemini-2.5-pro}

\textbf{\textit{environment}}: outdoor 
\textbf{\textit{scene\_type}}: street 
\textbf{\textit{setting}}: urban

\textbf{\textit{landmark\_recognition}}: The tall, light-colored residential building visible in the background is the Permata Senayan apartment complex in Jakarta.
\textbf{\textit{text\_and\_signage}}: Multiple road signs are visible in Indonesian. The most prominent yellow sign reads "Hati - Hati Lintasan Bus Transjakarta," which translates to "Be Careful, Transjakarta Bus Crossing." Another sign below it reads "HATI-HATI PORTAL BUSWAY." Transjakarta is the Bus Rapid Transit (BRT) system exclusive to Jakarta, Indonesia. A vehicle license plate starting with the letter 'B' is also partially visible, which is the vehicle registration code for the Jakarta metropolitan area.
\textbf{\textit{cultural\_indicators}}: The language on the signs is Indonesian. The high density of traffic, including numerous motorcycles, is characteristic of major cities in Indonesia, particularly Jakarta.
\textbf{\textit{spatial\_context}}: The scene depicts a crossing for a dedicated busway. The combination of the specific sign configuration, the location of the Permata Senayan apartment building in the background, the fence on the right (belonging to the Gelora Bung Karno complex), and the orientation of the road points directly to Jalan Tentara Pelajar, near the Palmerah railway station in Central Jakarta. The camera is facing southeast along this road.

\textbf{\textit{Confidence}}: 5 
\textbf{\textit{Address}}: Jalan Tentara Pelajar, Jakarta, JAKARTA, Indonesia 
\textbf{\textit{Coordinates}}: -6.2117, 106.7977 \\
\hline

\raisebox{-\totalheight}{\includegraphics[width=0.15\textwidth]{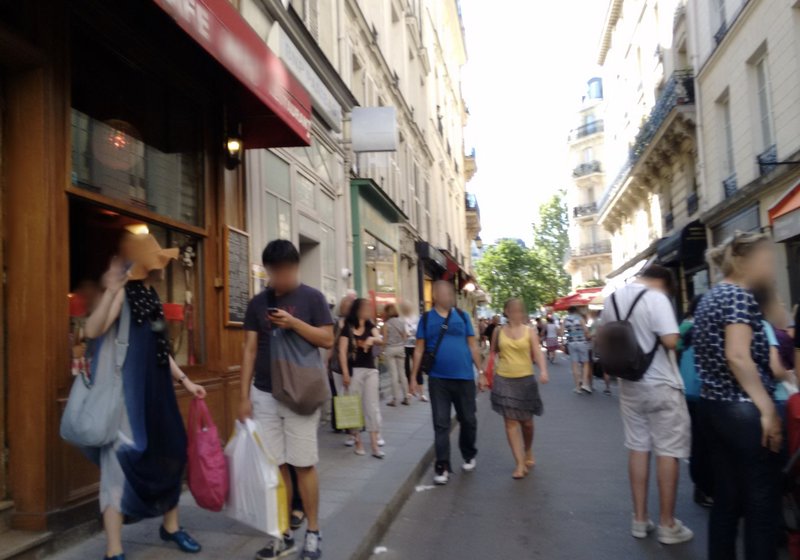}} & \textbf{Address}: Paris, France \textbf{Coordinates}: 48.8523, 2.3541 & \textbf{Model: o3}

\textbf{\textit{environment}}: outdoor 
\textbf{\textit{scene\_type}}: street 
\textbf{\textit{setting}}: urban

\textbf{\textit{landmark\_recognition}}: Haussmann-style cream stone buildings with wrought-iron balconies typical of central Paris streets; red café awnings common on well-known pedestrian market streets such as Rue Cler.
\textbf{\textit{text\_and\_signage}}: Word "CAFÉ" in French clearly visible on red awning; menu board beside wooden café frontage also in French.
\textbf{\textit{cultural\_indicators}}: Parisian sidewalk-café culture, narrow cobbled roadway shared by pedestrians, architectural moldings and window shutters characteristic of late-19th-century Paris buildings.
\textbf{\textit{spatial\_context}}: Pedestrian market street lined with small shops, cafés, and outdoor produce stalls matches the layout of the 7th-arrondissement market street Rue Cler located between Avenue de la Motte-Picquet and Rue de Grenelle.

\textbf{\textit{Confidence}}: 3 
\textbf{\textit{Address}}: Rue Cler, Paris, ÎLE-DE-FRANCE, France 
\textbf{\textit{Coordinates}}: 48.8569, 2.3062 \\
\hline

\raisebox{-\totalheight}{\includegraphics[width=0.15\textwidth]{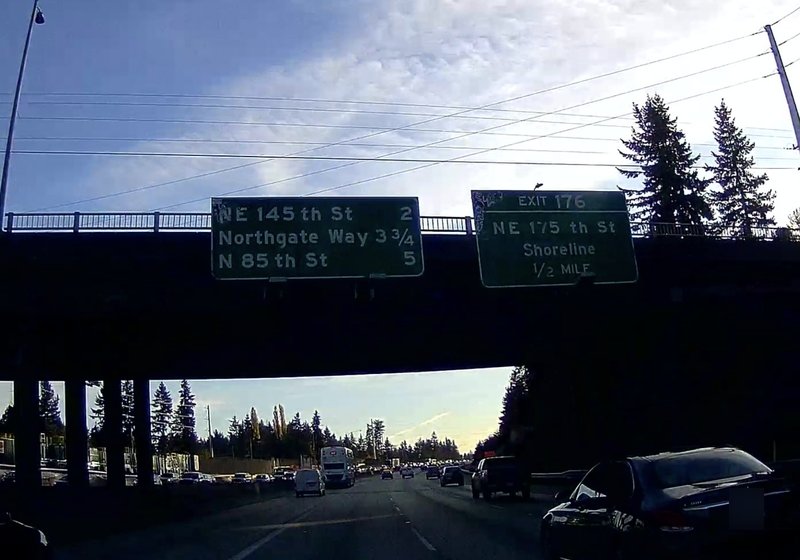}} & \textbf{Address}: Shoreline, USA \textbf{Coordinates}: 47.7633, -122.3239 & \textbf{Model: gemini-2.5-pro}

\textbf{\textit{environment}}: outdoor 
\textbf{\textit{scene\_type}}: street 
\textbf{\textit{setting}}: suburban

\textbf{\textit{landmark\_recognition}}: The primary landmarks are the highway infrastructure itself, including a multi-lane freeway, an overpass, and large overhead gantry signs. The presence of tall evergreen trees is characteristic of the Pacific Northwest region of the United States. 
\textbf{\textit{text\_and\_signage}}: Multiple highway signs are clearly visible. One sign indicates 'EXIT 176' for 'NE 175th St' and 'Shoreline' is 1/2 mile ahead. Another sign lists upcoming exits and their distances: 'NE 145th St' (2 miles), 'Northgate Way' (3 1/4 miles), and 'N 85th St' (5 miles). These exits, street names, and locations (Shoreline, Northgate) are all situated along Interstate 5 in the northern part of Seattle and its immediate suburbs in Washington State.
\textbf{\textit{cultural\_indicators}}: The style of the green highway signs with white text (Highway Gothic font) conforms to the Manual on Uniform Traffic Control Devices (MUTCD) used in the United States. The use of 'miles' for distance measurement is standard in the US. 
\textbf{\textit{spatial\_context}}: The image is taken from a vehicle traveling northbound on a major freeway. The exit numbers and destinations confirm the location is on Interstate 5 (I-5) North. Specifically, the signs for Exit 176 (NE 175th St) being 1/2 mile away, and being mounted on an overpass, place the camera's location just south of the NE 185th St overpass in Shoreline, Washington, looking north.

\textbf{\textit{Confidence}}: 5 
\textbf{\textit{Address}}: Interstate 5 N, Shoreline, WA, USA 
\textbf{\textit{Coordinates}}: 47.7592, -122.3293 \\
\hline

\raisebox{-\totalheight}{\includegraphics[width=0.15\textwidth]{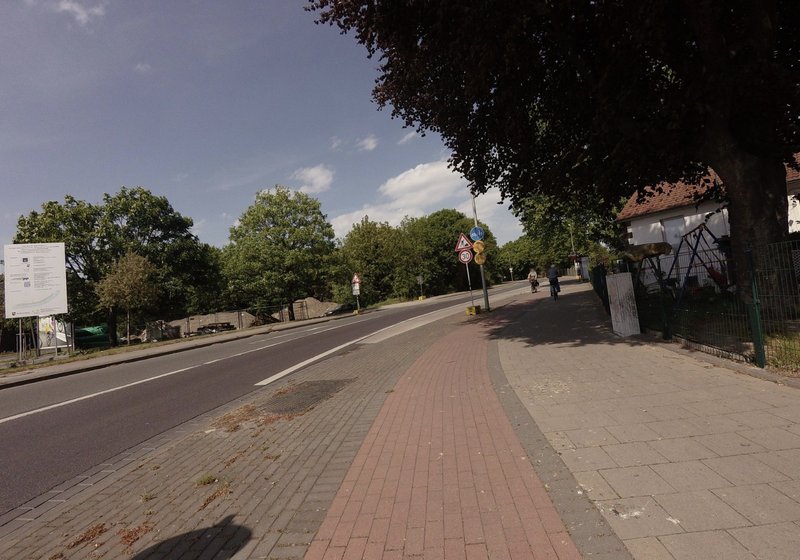}} & \textbf{Address}: Hildesheim, Germany \textbf{Coordinates}: 52.1528, 9.9359 & \textbf{Model: o3}

\textbf{\textit{environment}}: outdoor 
\textbf{\textit{scene\_type}}: street 
\textbf{\textit{setting}}: suburban

\textbf{\textit{landmark\_recognition}}: No globally iconic landmark visible; single-family house with red tile roof and fenced yard on right side.
\textbf{\textit{text\_and\_signage}}: Large white construction-information board on the left shows an EU flag logo and multi-line text typical of EU-funded public works signs in Germany; road signs are European style (red-bordered triangular warning sign, round red-bordered 30 km/h speed limit, blue shared-path sign).
\textbf{\textit{cultural\_indicators}}: Red/grey concrete block sidewalk with integrated cycle lane, metal mesh fencing, and the specific designs of the traffic signs match common German street infrastructure; house architecture (white stucco, steep red clay roof) also typical for central Germany.
\textbf{\textit{spatial\_context}}: Two-lane road with modest traffic, greenery and detached housing indicate a suburban area on the edge of a medium/large German city.

\textbf{\textit{Confidence}}: 2 
\textbf{\textit{Address}}: Berliner Straße, Potsdam, BRANDENBURG, Germany 
\textbf{\textit{Coordinates}}: 52.4000, 13.0400 \\
\hline

\raisebox{-\totalheight}{\includegraphics[width=0.15\textwidth]{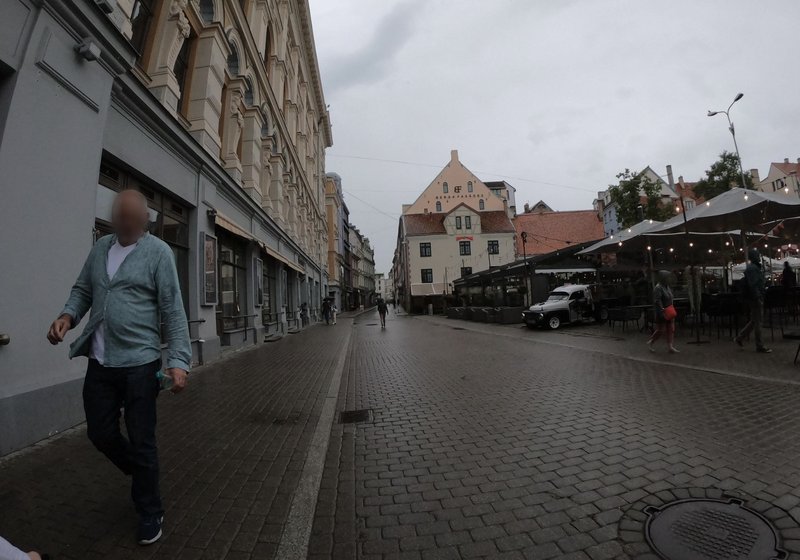}} & \textbf{Address}: Riga, Latvia \textbf{Coordinates}: 56.9492, 24.1100 & \textbf{Model: gemini-2.5-pro}

\textbf{\textit{environment}}: outdoor 
\textbf{\textit{scene\_type}}: street 
\textbf{\textit{setting}}: urban

\textbf{\textit{landmark\_recognition}}: The architecture is characteristic of a Northern or Eastern European old town, specifically the Hanseatic style found in the Baltic states. The pink building at the end of the street with a stepped gable is a distinctive local landmark.
\textbf{\textit{text\_and\_signage}}: A crucial clue is the text 'BERGA FASĀDES' visible on the gable of the pink building. 'Fasādes' is the Latvian word for 'facades'. Below the windows on the same building, the word 'RESTORĀNS' (Latvian for 'restaurant') can be faintly seen. These linguistic cues definitively point to Latvia.
\textbf{\textit{cultural\_indicators}}: The cobblestone street, the style of the buildings, and the general atmosphere are highly indicative of the Old Town (Vecrīga) in Riga, Latvia. The presence of numerous outdoor cafes is also typical for this historic tourist area.
\textbf{\textit{spatial\_context}}: The photograph is taken looking down a narrow, pedestrianized street. By cross-referencing the building with 'BERGA FASĀDES' with maps of Riga, the location is identified as Tirgoņu iela, looking towards the intersection with Šķūņu iela.

\textbf{\textit{Confidence}}: 5 
\textbf{\textit{Address}}: Tirgoņu iela, Rīga, RĪGA, Latvia 
\textbf{\textit{Coordinates}}: 56.9479, 24.1084 \\
\hline

\raisebox{-\totalheight}{\includegraphics[width=0.15\textwidth]{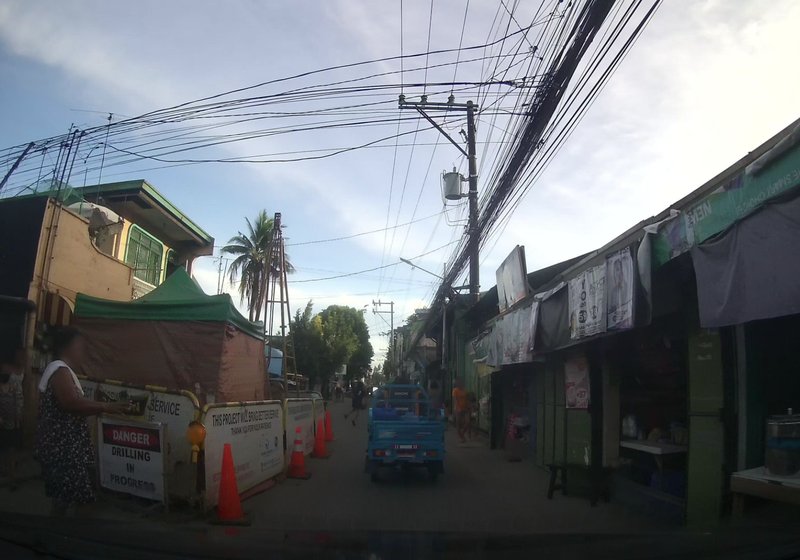}} & \textbf{Address}: Cainta, Philippines \textbf{Coordinates}: 14.5765, 121.1261 & \textbf{Model: o3}

\textbf{\textit{environment}}: outdoor 
\textbf{\textit{scene\_type}}: street 
\textbf{\textit{setting}}: urban

\textbf{\textit{landmark\_recognition}}: No globally iconic landmark visible; scene shows small shops and residential structures typical of Southeast Asian urban neighborhoods.
\textbf{\textit{text\_and\_signage}}: English construction warning signs such as 'DANGER DRILLING IN PROGRESS' and 'THIS PROJECT WILL BENEFIT THE ...' plus campaign-style posters with English names (e.g., 'BEN') suggest the Philippines where English is an official language used in public signage.
\textbf{\textit{cultural\_indicators}}: Dense overhead utility wires, jeepney-sized blue mini-truck, tropical coconut palm, concrete block houses with metal window grilles, and roadside sari-sari-store style stalls strongly resemble streets in Philippine barangays.
\textbf{\textit{spatial\_context}}: Narrow two-way lane lined with stalls and residences, temporary construction barriers, and cluttered electrical infrastructure typical of inner neighborhoods in Metro Manila cities such as Quezon City.

\textbf{\textit{Confidence}}: 3 
\textbf{\textit{Address}}: Quezon City, METRO MANILA, Philippines 
\textbf{\textit{Coordinates}}: 14.6500, 121.0500 \\
\hline





\raisebox{-\totalheight}{\includegraphics[width=0.15\textwidth]{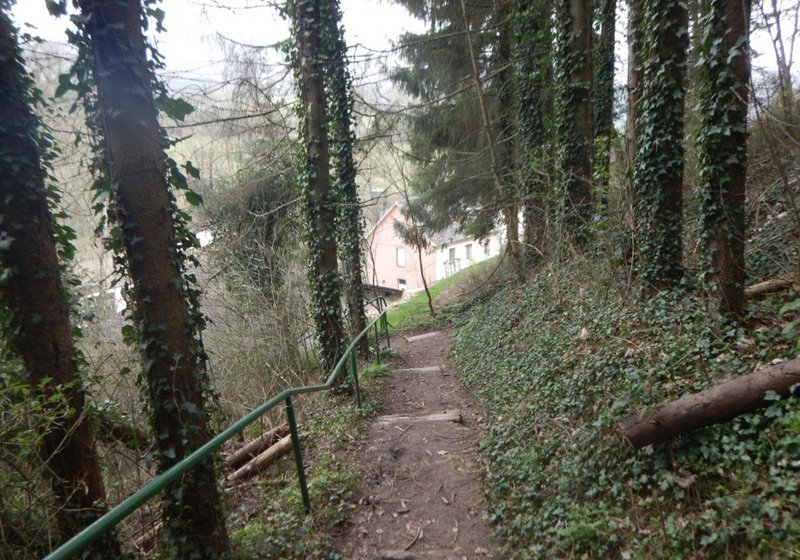}} & \textbf{Address}: Echternach, Luxembourg \textbf{Coordinates}: 49.8355, 6.3285 & \textbf{Model: o3}

\textbf{\textit{environment}}: outdoor 
\textbf{\textit{scene\_type}}: scenery 
\textbf{\textit{setting}}: rural

\textbf{\textit{landmark\_recognition}}: No globally iconic landmark visible; scene shows a forested footpath descending toward a plaster-walled house with a steep roof typical of Central European villages.
\textbf{\textit{text\_and\_signage}}: No readable text or signage present.
\textbf{\textit{cultural\_indicators}}: Green-painted metal handrail, English ivy climbing conifer and deciduous trunks, and pastel stucco house are characteristic of small towns in western Germany's river valleys (e.g., Mosel/Rhine).
\textbf{\textit{spatial\_context}}: Path on a wooded hillside suggests settlements nestled in narrow valleys; common in Rhineland-Palatinate where towns sit between steep, forested slopes and rivers.

\textbf{\textit{Confidence}}: 2 
\textbf{\textit{Address}}: Moselpromenade 1, Cochem, RHINELAND-PALATINATE, Germany 
\textbf{\textit{Coordinates}}: 50.1453, 7.1656 \\
\hline

\end{tabular}
}
\label{tab:sample_gss}
\end{table*}


\begin{table*}[ht]
\centering
\caption{Representative LLM image geolocalization output in Dataset-UPC.}
\label{tab:sample_upc}
{\scriptsize
\begin{tabular}{|p{3cm}|p{2cm}|p{11cm}|}
\hline
\textbf{Image} & \textbf{Ground-Truth} & \textbf{LLM Output} \\
\hline

\raisebox{-\totalheight}{\includegraphics[width=0.15\textwidth]{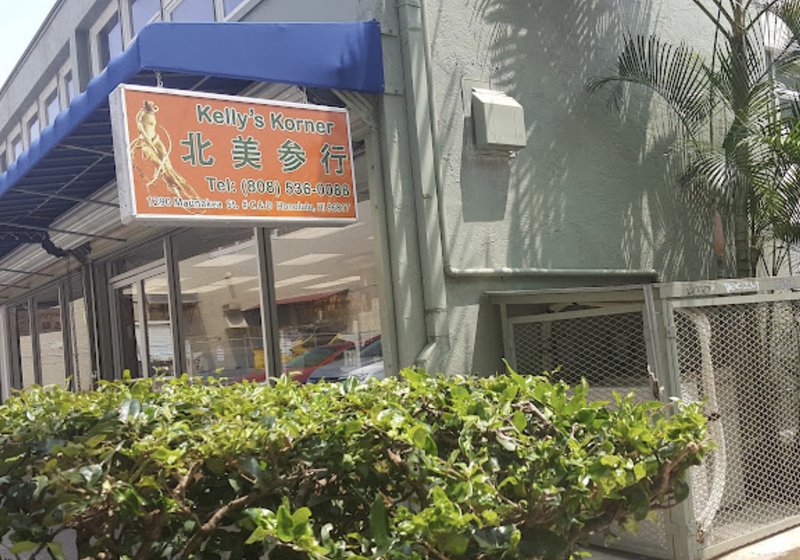}} & \textbf{Address}: Philadelphia, USA \textbf{Coordinates}: 39.9531, -75.1635 & \textbf{Model: o3}

\textbf{\textit{environment}}: outdoor 
\textbf{\textit{scene\_type}}: building 
\textbf{\textit{setting}}: urban

\textbf{\textit{landmark\_recognition}}: Ornate Second-Empire style municipal building with a central clock tower topped by a statue, matching the well-known Philadelphia City Hall and its William Penn statue
\textbf{\textit{text\_and\_signage}}: No readable signage visible but the clock face and tower design are distinctive enough for identification	
\textbf{\textit{cultural\_indicators}}: Presence of both English and Chinese text matches Honolulu Chinatown area; tropical plants and palms fit Hawaiian climate
\textbf{\textit{spatial\_context}}: Street-level retail space in a mid-rise urban block typical of downtown Honolulu near Queen Emma Street corridor

\textbf{\textit{Confidence}}: 4 
\textbf{\textit{Address}}: 1280 Queen Emma Street, Honolulu, HI, USA 
\textbf{\textit{Coordinates}}: 21.3148, -157.8576 \\
\hline

\raisebox{-\totalheight}{\includegraphics[width=0.15\textwidth]{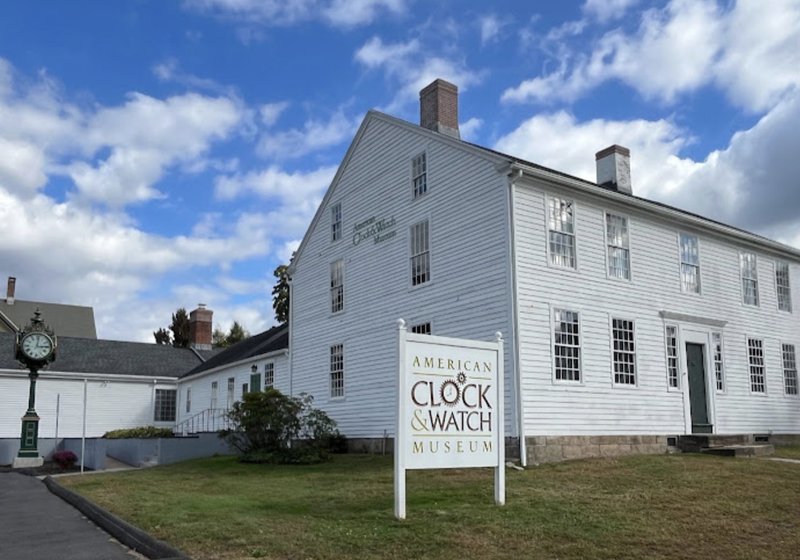}} & \textbf{Address}: Bristol, CT \textbf{Coordinates}: 41.6796, -72.9408 & \textbf{Model: gemini-2.5-pro}

\textbf{\textit{environment}}: outdoor 
\textbf{\textit{scene\_type}}: building 
\textbf{\textit{setting}}: suburban

\textbf{\textit{landmark\_recognition}}: The main building is the American Clock \& Watch Museum, a specific cultural landmark. An ornate street clock is also visible on the left side of the property.
\textbf{\textit{text\_and\_signage}}: A large sign in the foreground and a smaller sign on the building itself clearly read 'AMERICAN CLOCK \& WATCH MUSEUM'.
\textbf{\textit{cultural\_indicators}}: The building exhibits a classic American Colonial architectural style, characterized by its white clapboard siding, symmetrical facade, and multi-pane sash windows, which is very common in the New England region of the United States.
\textbf{\textit{spatial\_context}}: The image shows a large historic house set back from a road with a lawn in front, next to a smaller, attached building and a parking area, which is consistent with the layout of a museum in a suburban or small-town setting.

\textbf{\textit{Confidence}}: 5 
\textbf{\textit{Address}}: 100 Maple St, Bristol, CT, USA 
\textbf{\textit{Coordinates}}: 41.6766, -72.9461 \\
\hline

\raisebox{-\totalheight}{\includegraphics[width=0.15\textwidth]{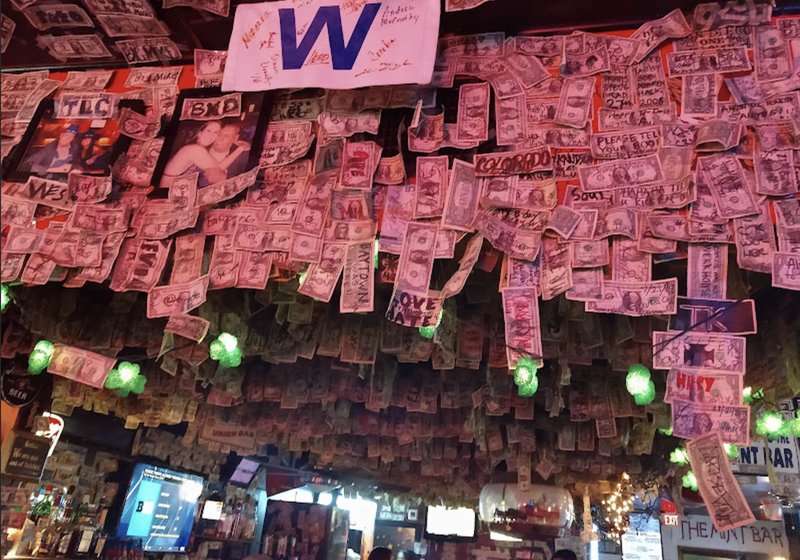}} & \textbf{Address}: Norfolk, NE \textbf{Coordinates}: 42.0158, -97.4102 & \textbf{Model: gemini-2.5-pro}

\textbf{\textit{environment}}: indoor 
\textbf{\textit{scene\_type}}: building 
\textbf{\textit{setting}}: urban

\textbf{\textit{landmark\_recognition}}: The most significant visual feature is the ceiling and upper walls being completely covered in autographed US dollar bills. This is a distinctive tradition of a limited number of bars in the United States, making it a key identifying feature.
\textbf{\textit{text\_and\_signage}}: Several crucial pieces of text are visible. A sign on the lower right, partially obscured, reads 'THE MINT BAR'. A second sign, located under a television on the left side, says 'UNION BAR'. The combination of a bar featuring both 'Union Bar' and 'The Mint Bar' names, along with the dollar bill decor, is highly specific. A large white flag with a blue 'W' is prominently displayed, which is the victory flag for the Chicago Cubs baseball team. While this could be found in many bars, it fits the context of a Midwestern location. Other text on dollar bills like 'COLORADO' and 'BAYTOWN' are likely from visitors and less indicative of the primary location.
\textbf{\textit{cultural\_indicators}}: The tradition of pinning signed currency to the walls is a cultural practice in some American saloons and dive bars, often accumulating over many decades. The Chicago Cubs 'W' flag is a strong element of American sports culture, particularly in the Midwest. The green, shamrock-shaped string lights suggest an Irish pub theme or St. Patrick's Day decorations, common in American bars.
\textbf{\textit{spatial\_context}}: The sheer density of the currency covering almost every available surface suggests this is a very old, long-standing tradition at this establishment. This points towards a historic local bar rather than a new or chain establishment. A search for bars with these specific features (dollar bills, 'Union Bar', 'The Mint Bar') confirms the identity of a well-known location in Nebraska.

\textbf{\textit{Confidence}}: 5 
\textbf{\textit{Address}}: 1401 10th St, Gering, NE, USA 
\textbf{\textit{Coordinates}}: 41.8242, -103.6625 \\
\hline

\raisebox{-\totalheight}{\includegraphics[width=0.15\textwidth]{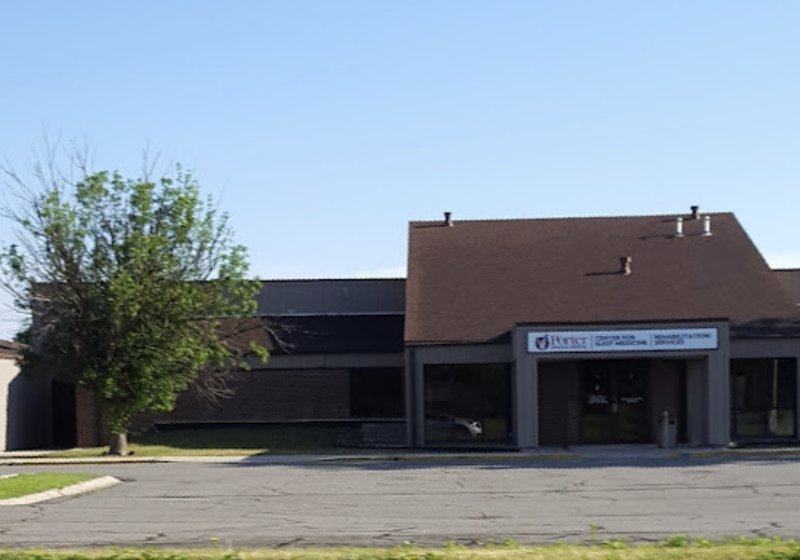}} & \textbf{Address}: Portage, IN \textbf{Coordinates}: 41.5583, -87.1814 & \textbf{Model: gemini-2.5-pro}

\textbf{\textit{environment}}: outdoor 
\textbf{\textit{scene\_type}}: building 
\textbf{\textit{setting}}: suburban

\textbf{\textit{landmark\_recognition}}: The image features a single-story commercial building with a distinct gabled entrance. The architecture, combining brick, siding, and shingled roofing, is common for suburban medical or professional offices in the United States.
\textbf{\textit{text\_and\_signage}}: A prominent sign is visible on the building's facade. The sign includes a logo with the name 'Porter' and text that reads 'CENTER FOR SLEEP MEDICINE' and 'REHABILITATION SERVICES'. This information is key to identifying the specific business and its location.
\textbf{\textit{cultural\_indicators}}: The building style, the presence of a large asphalt parking lot, and the manicured grass strip are all characteristic of suburban development patterns in North America.
\textbf{\textit{spatial\_context}}: A search for 'Porter Center for Sleep Medicine \& Rehabilitation Services' leads to Northwest Health - Porter facilities in Indiana. Using Google Maps and Street View, the building in the image is an exact match for the one located at 2000 Roosevelt Rd in Valparaiso, IN. The building's structure, sign placement, and surrounding features align perfectly with online imagery of this address.

\textbf{\textit{Confidence}}: 5 
\textbf{\textit{Address}}: 2000 Roosevelt Rd,Valparaiso, IN, USA 
\textbf{\textit{Coordinates}}: 41.4880, -87.0545 \\
\hline





\raisebox{-\totalheight}{\includegraphics[width=0.15\textwidth]{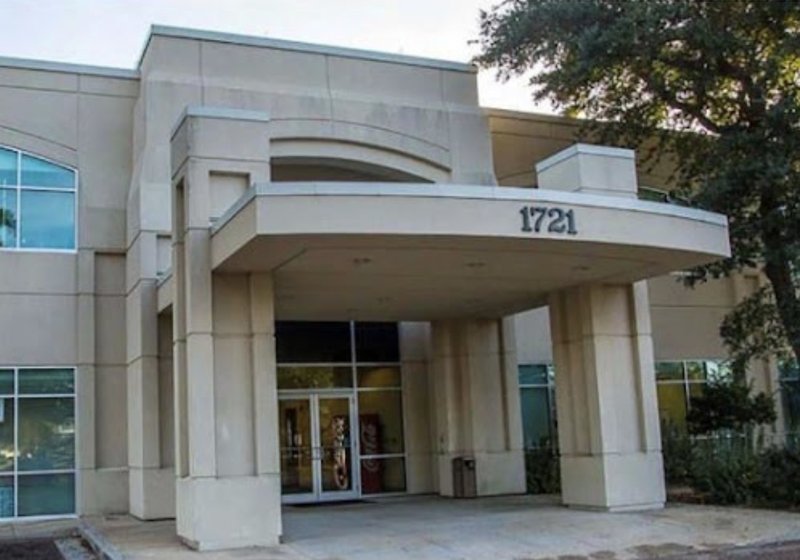}} & \textbf{Address}: Biloxi, MS \textbf{Coordinates}: 30.4461, -88.9394 & \textbf{Model: o3}

\textbf{\textit{environment}}: outdoor 
\textbf{\textit{scene\_type}}: building 
\textbf{\textit{setting}}: suburban

\textbf{\textit{landmark\_recognition}}: No globally iconic landmark, but the facade resembles mid-1990s—2000s U.S. suburban office-park architecture (stucco/EIFS, shallow arch window tops, square columns).
\textbf{\textit{text\_and\_signage}}: The only visible text is the street number "1721" mounted on the entrance canopy.
\textbf{\textit{cultural\_indicators}}: Presence of large live-oak trees, beige stucco, and green-tinted ribbon windows strongly suggest a warm southern-U.S. region, particularly Texas where this design and vegetation are common.
\textbf{\textit{spatial\_context}}: Looks like a stand-alone low-rise in an office park surrounded by mature trees rather than dense urban grid.

\textbf{\textit{Confidence}}: 2 
\textbf{\textit{Address}}: 1721 River Run, Fort Worth, TX, USA 
\textbf{\textit{Coordinates}}: 32.7362, -97.3620 \\
\hline

\raisebox{-\totalheight}{\includegraphics[width=0.15\textwidth]{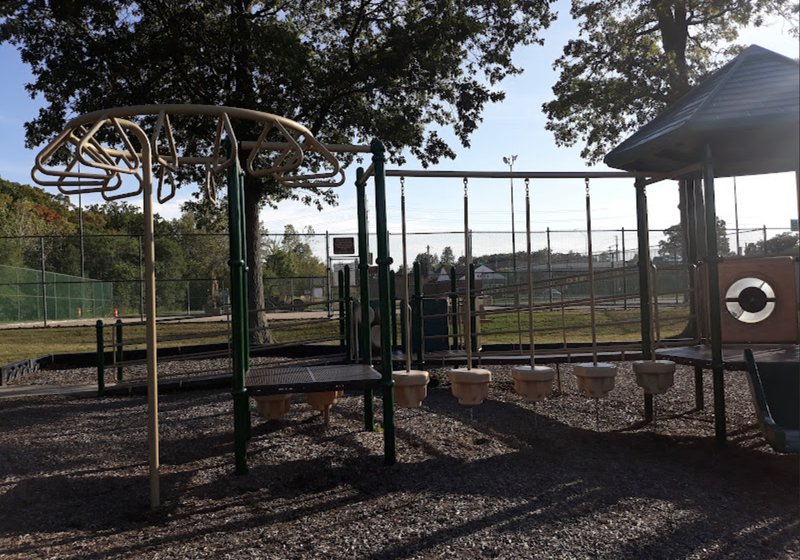}} & \textbf{Address}: Bay Village, OH \textbf{Coordinates}: 41.4794, -81.8815 & \textbf{Model: gemini-2.5-pro}

\textbf{\textit{environment}}: outdoor 
\textbf{\textit{scene\_type}}: mixed 
\textbf{\textit{setting}}: suburban

\textbf{\textit{landmark\_recognition}}: The primary identifying feature is the unique circular monkey bar structure, known as a 'RingTrek' climber. The scene also includes adjacent tennis courts with green surfaces and a chain-link fence, a swing set with distinctive beige bucket-like seats, and a small play structure with a green hexagonal roof. The specific combination and arrangement of this playground equipment are highly distinctive.
\textbf{\textit{text\_and\_signage}}: A brown sign is visible on a post, likely a standard park rules sign, but the text is illegible and does not contribute to the identification.
\textbf{\textit{cultural\_indicators}}: The style of the playground equipment is common in modern North American public parks. The houses visible in the background are of a typical suburban American design.
\textbf{\textit{spatial\_context}}: The layout of a playground directly adjacent to tennis courts, bordered by trees and with residential housing in the near distance, is a common configuration for a community park in the United States. The deciduous trees suggest a temperate climate.

\textbf{\textit{Confidence}}: 5 
\textbf{\textit{Address}}: Continental Pl, Glen Cove, NY, USA 
\textbf{\textit{Coordinates}}: 40.8529, -73.6338 \\
\hline

\raisebox{-\totalheight}{\includegraphics[width=0.15\textwidth]{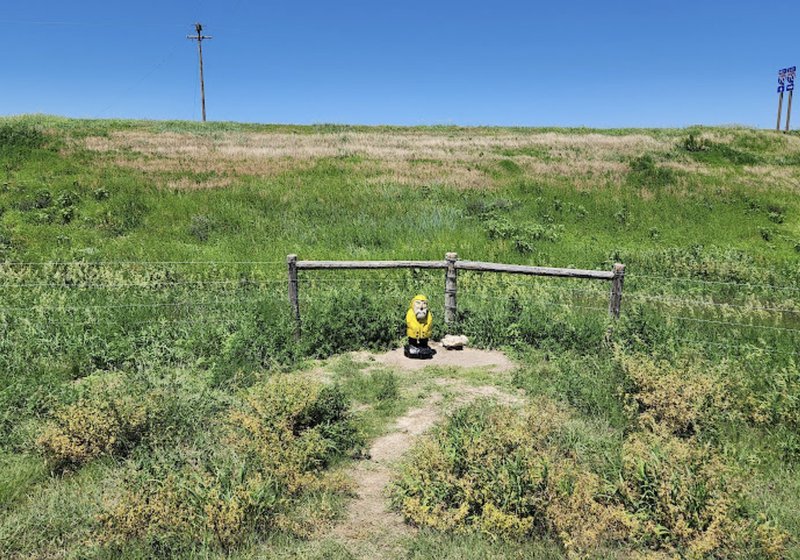}} & \textbf{Address}: Colby, KS \textbf{Coordinates}: 39.2779, -100.9437 & \textbf{Model: gemini-2.5-pro}

\textbf{\textit{environment}}: outdoor 
\textbf{\textit{scene\_type}}: scenery 
\textbf{\textit{setting}}: rural

\textbf{\textit{landmark\_recognition}}: The central feature is a small, yellow statue of a monkey-like figure in a raincoat. This is a known, unique roadside attraction called the 'I-80 Rainslicker Monkey' or 'Monkey Tree', which serves as an informal memorial.
\textbf{\textit{text\_and\_signage}}: In the background, on top of the grassy embankment, there are road signs. A cropped image clearly shows a US Interstate shield for Interstate 80 (I-80) and a US Highway shield for US Route 30 (US-30). These two highways run concurrently in this region.
\textbf{\textit{cultural\_indicators}}: The object is a small, informal roadside memorial, a cultural practice found along highways in the United States, often maintained by locals to commemorate an event or person.
\textbf{\textit{spatial\_context}}: The image is taken from a low vantage point looking up at a grassy embankment next to a major highway. The presence of both I-80 and US-30 signs, combined with the rolling, grassy, and sparsely populated landscape, is characteristic of southern Wyoming.

\textbf{\textit{Confidence}}: 5 
\textbf{\textit{Address}}: Interstate 80, Buford, WY, USA 
\textbf{\textit{Coordinates}}: 41.2251, -105.4242 \\
\hline

\end{tabular}
}
\label{tab:sample_ups}
\end{table*}


\begin{table*}[ht]
\centering
\caption{Representative LLM image geolocalization output in Dataset-PCW.}
{\scriptsize
\begin{tabular}{|p{3cm}|p{2cm}|p{11cm}|}
\hline
\textbf{Image} & \textbf{Ground-Truth} & \textbf{LLM Output} \\
\hline

\raisebox{-\totalheight}{\includegraphics[width=0.15\textwidth]{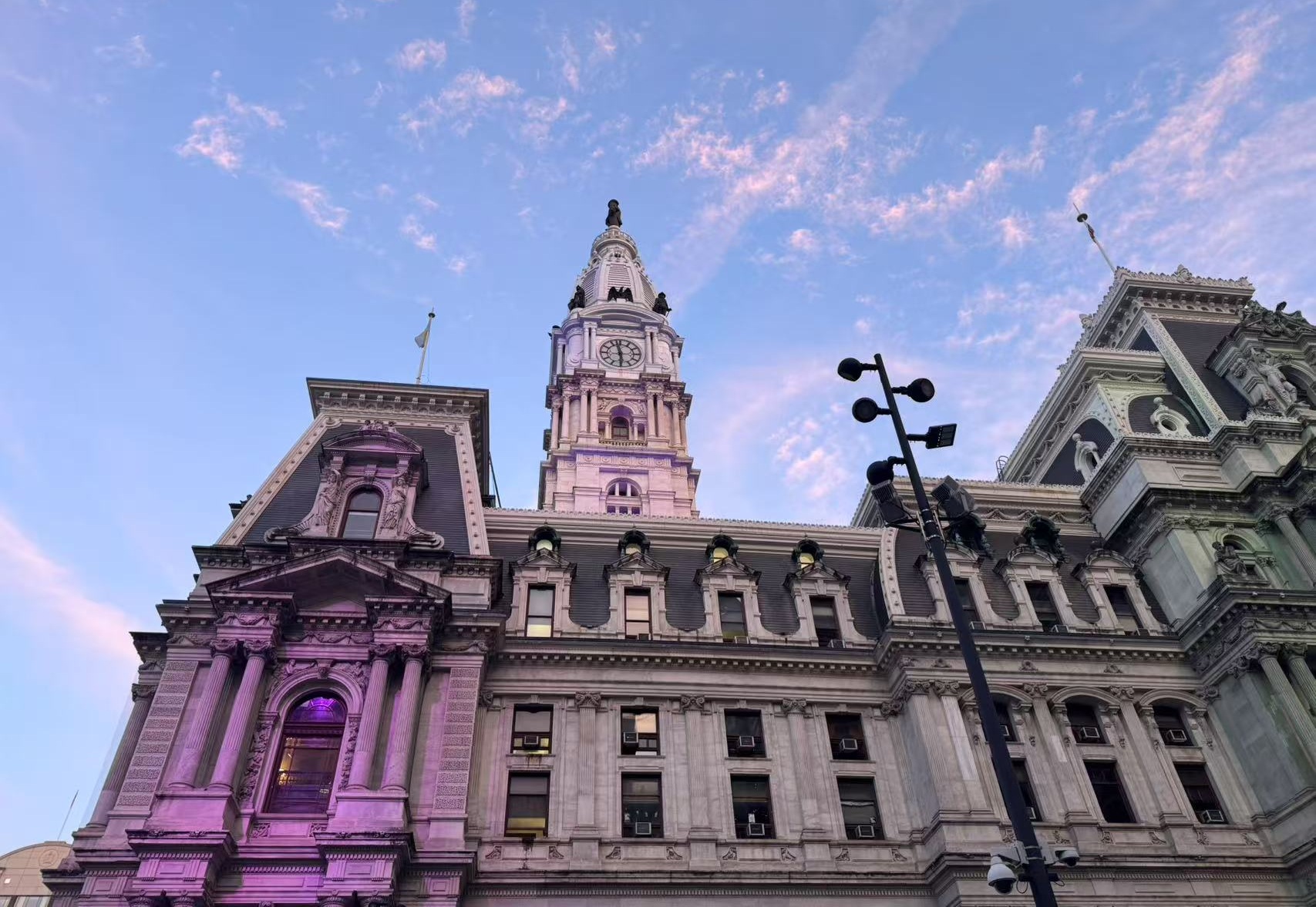}} & \textbf{Address}: Philadelphia, USA \textbf{Coordinates}: 39.9531, -75.1635 & \textbf{Model: o3}

\textbf{\textit{environment}}: outdoor  
\textbf{\textit{scene\_type}}: building  
\textbf{\textit{setting}}: urban   

\textbf{\textit{landmark\_recognition}}: Ornate Second-Empire style municipal building with a central clock tower topped by a statue, matching the well-known Philadelphia City Hall and its William Penn statue  
\textbf{\textit{text\_and\_signage}}: No readable signage visible but the clock face and tower design are distinctive enough for identification   
\textbf{\textit{cultural\_indicators}}: French Second-Empire architecture common to late-19th-century U.S. civic buildings, specifically the gray granite and elaborate sculptural ornamentation characteristic of Philadelphia City Hall  
\textbf{\textit{spatial\_context}}: Large civic structure occupying a city square in the heart of a dense downtown grid, consistent with Center City Philadelphia where City Hall sits at the crossing of Broad and Market Streets  

\textbf{\textit{confidence}}: 5  
\textbf{\textit{address}}: 1400 John F Kennedy Blvd, Philadelphia, PA, USA  
\textbf{\textit{coordinates}}: 39.9526, -75.1636 \\
\hline

\raisebox{-\totalheight}{\includegraphics[width=0.15\textwidth]{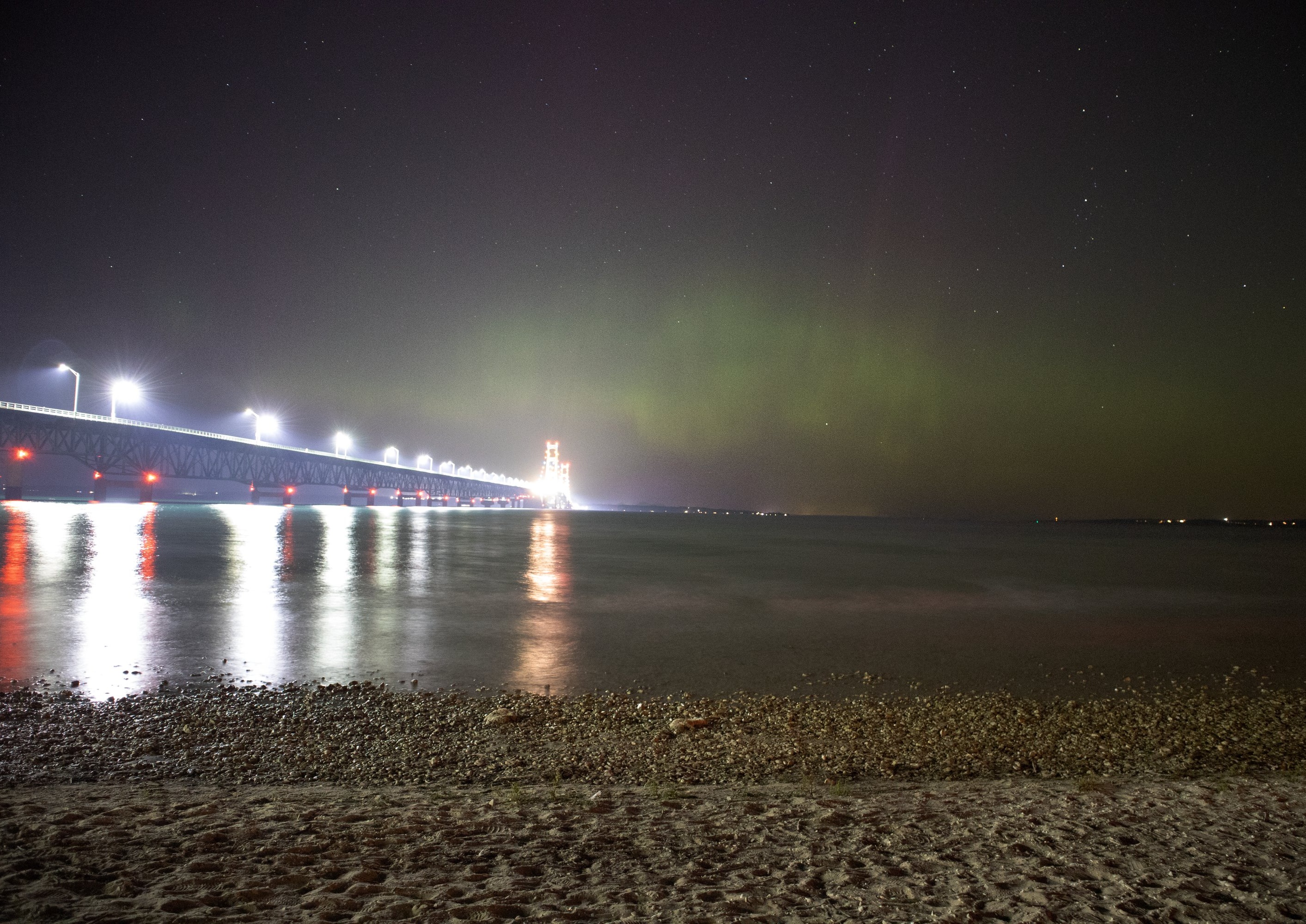}} & \textbf{Address}: Mackinaw City, USA \textbf{Coordinates}: 45.8388, -84.7209 & \textbf{Model: Gemini-2.5-pro}

\textbf{\textit{environment}}: outdoor  
\textbf{\textit{scene\_type}}: scenery  
\textbf{\textit{setting}}: natural   

\textbf{\textit{landmark\_recognition}}: The central feature of the image is a very large, illuminated suspension bridge at night. The design, with its two prominent lighted towers and extensive truss structure, is unmistakably the Mackinac Bridge, which connects the Upper and Lower Peninsulas of the U.S. state of Michigan. Additionally, a faint green glow of the aurora borealis (Northern Lights) is visible in the sky, a phenomenon often observed in this northern region.   
\textbf{\textit{text\_and\_signage}}: There is no legible text or signage visible in the image.  
\textbf{\textit{cultural\_indicators}}: The scale and style of the bridge are characteristic of major mid-20th century American civil engineering projects. The presence of the aurora suggests a location at a high latitude in the Northern Hemisphere.  
\textbf{\textit{spatial\_context}}: The photograph is taken from a sandy and pebbly beach, looking across a large body of water (the Straits of Mackinac) towards the bridge. The angle of the bridge suggests the photo was taken from the southern side, in Mackinaw City, looking north/northeast. A popular and accessible viewing spot with this perspective is Alexander Henry Park, located just west of the bridge's southern approach. 

\textbf{\textit{confidence}}: 5  
\textbf{\textit{address}}: Alexander Henry Park, Mackinaw City, MI, USA  
\textbf{\textit{coordinates}}: 45.782, -84.7334 \\
\hline

\raisebox{-\totalheight}{\includegraphics[width=0.15\textwidth]{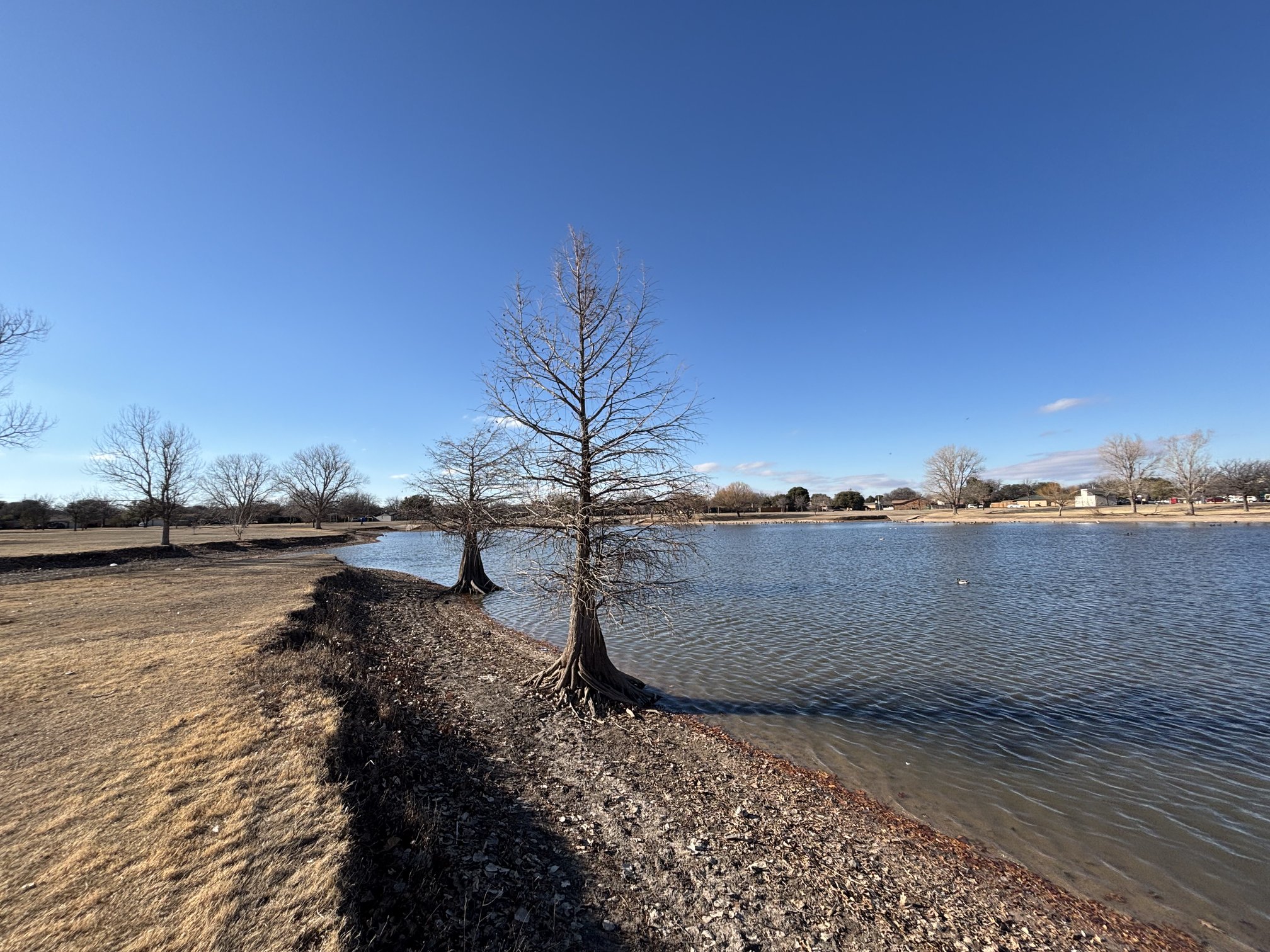}} & \textbf{Address}: Lubbock, USA \textbf{Coordinates}: 33.5162, -101.919 & \textbf{Model: o3}

\textbf{\textit{environment}}: outdoor  
\textbf{\textit{scene\_type}}: scenery  
\textbf{\textit{setting}}: suburban   

\textbf{\textit{landmark\_recognition}}: No globally iconic landmark; scene shows a man-made pond or small lake with bald cypress trees that commonly line park lakes in the southern United States.   
\textbf{\textit{text\_and\_signage}}: No readable text or signage visible in the image.   
\textbf{\textit{cultural\_indicators}}: Dormant warm-season grass and bald cypress suggest a warm climate that still experiences a cool season; this combination is common in North Texas municipal parks.   
\textbf{\textit{spatial\_context}}: Flat terrain, scattered single-story buildings/roofs and a roadway in the distance imply a suburban park setting rather than an isolated rural lake or dense urban waterfront.  

\textbf{\textit{confidence}}: 2  
\textbf{\textit{address}}: 1405 Wilson Creek Pkwy, McKinney, TX, USA  
\textbf{\textit{coordinates}}: 33.1974, -96.6232 \\
\hline

\raisebox{-\totalheight}{\includegraphics[width=0.15\textwidth]{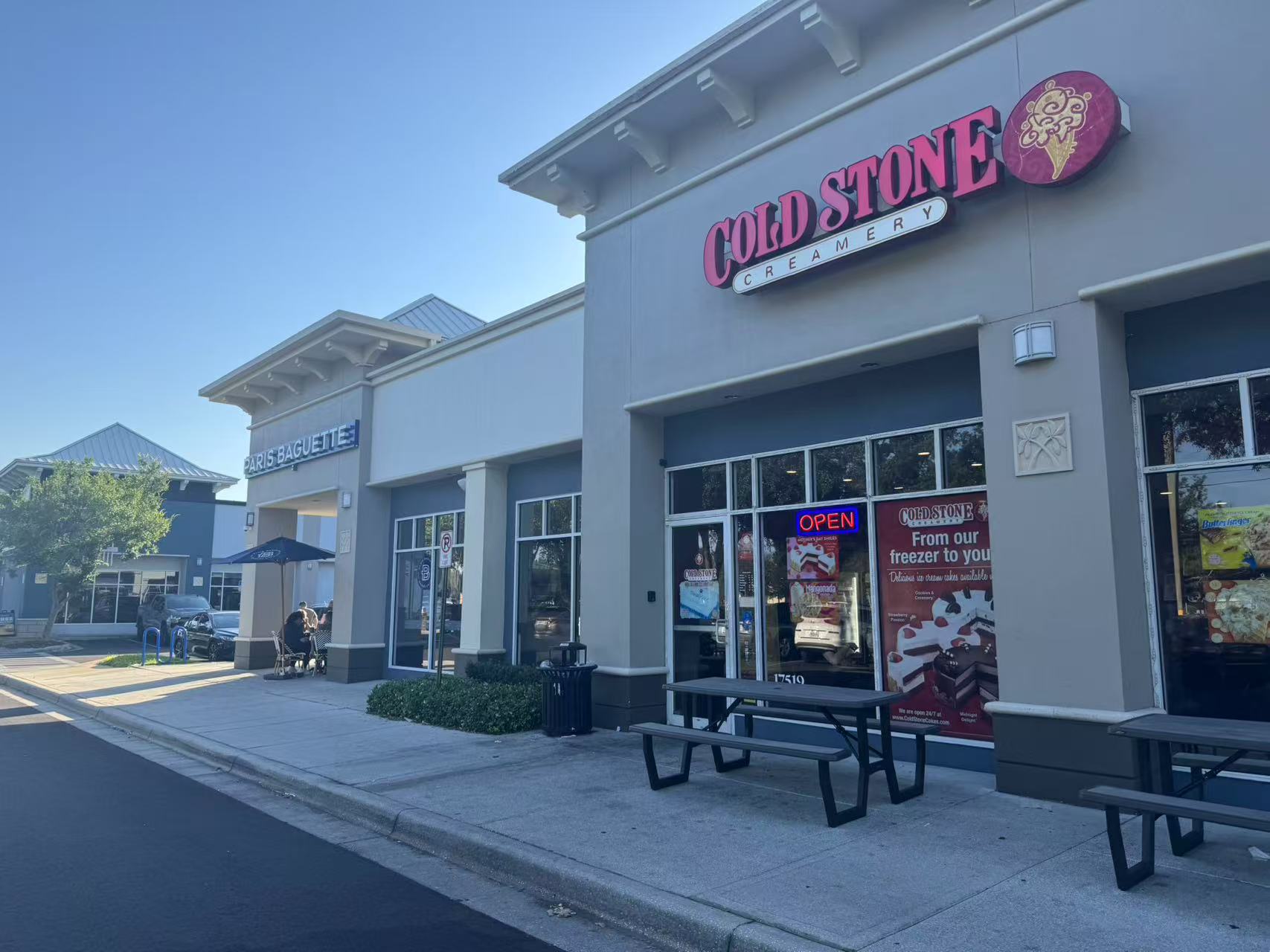}} & \textbf{Address}: Tampa, USA \textbf{Coordinates}: 28.1284, -82.368 & \textbf{Model: Gemini-2.5-pro}

\textbf{\textit{environment}}: outdoor  
\textbf{\textit{scene\_type}}: street  
\textbf{\textit{setting}}: suburban   

\textbf{\textit{landmark\_recognition}}: The image features storefronts in a modern suburban shopping plaza, not a unique landmark, but characteristic of this type of development.  
\textbf{\textit{text\_and\_signage}}: Multiple signs are clearly visible. The main storefronts are for 'COLD STONE CREAMERY' and 'PARIS BAGUETTE'. Crucially, the street number '17519' is displayed on a pillar at the entrance of the Cold Stone Creamery. A poster in the window also provides a URL: www.ColdStoneCakes.com, which is a US-based domain.   
\textbf{\textit{cultural\_indicators}}: The architectural style of the buildings is typical of contemporary American suburban retail centers or strip malls. The combination of these specific international food chains (Cold Stone Creamery and Paris Baguette) is common in more affluent suburban areas in the United States.   
\textbf{\textit{spatial\_context}}: The image shows two businesses located directly next to each other in a shopping center. The combination of Cold Stone Creamery, Paris Baguette, and the specific address number '17519' provides a unique set of data points to query for a specific location.  

\textbf{\textit{confidence}}: 5  
\textbf{\textit{address}}: 17519 W Colonial Dr, Winter Garden, FL, USA  
\textbf{\textit{coordinates}}: 28.552942, -81.564633 \\
\hline

\raisebox{-\totalheight}{\includegraphics[width=0.15\textwidth]{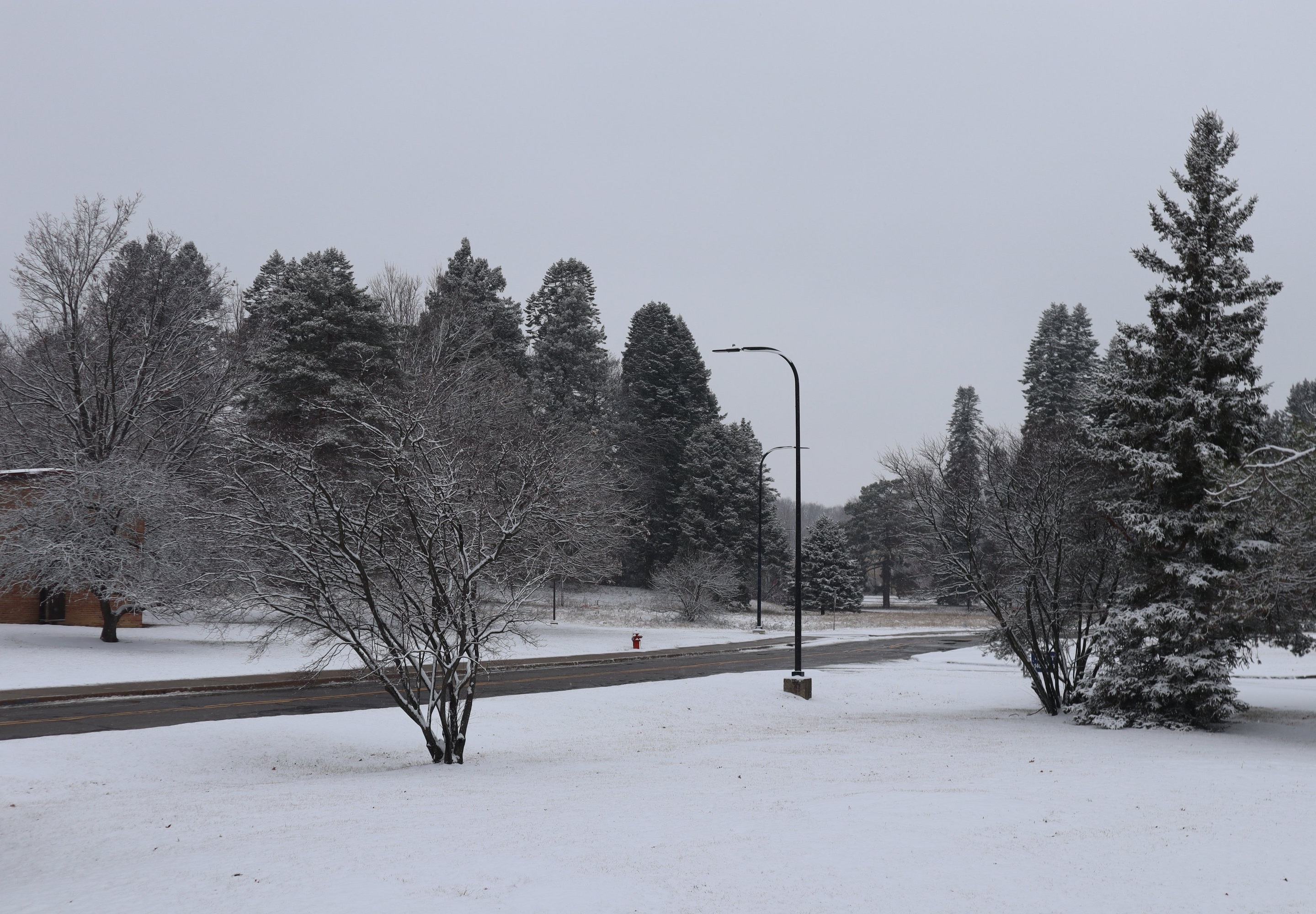}} & \textbf{Address}: Ann Arbor, USA \textbf{Coordinates}: 42.2991, -83.7129 & \textbf{Model: o3}

\textbf{\textit{environment}}: outdoor, 
\textbf{\textit{scene\_type}}: scenery, 
\textbf{\textit{setting}}: suburban   

\textbf{\textit{landmark\_recognition}}: No clearly identifiable landmark; landscape of mixed deciduous and coniferous trees with a small glimpse of a red-brick building typical of Midwestern college campuses   
\textbf{\textit{text\_and\_signage}}: No readable street signs, business names, or license plates visible   
\textbf{\textit{cultural\_indicators}}: Streetlights with simple black curved arms and a red fire-hydrant are common municipal furnishings in many U.S. northern cities; red-brick institutional building hints at a school or campus built mid-20th-century, often found in the Upper Midwest   
\textbf{\textit{spatial\_context}}: Lightly trafficked two-lane road inside a landscaped area, snow cover, and dense spruce/pine mix are consistent with cities along Lake Superior such as Duluth, Minnesota  

\textbf{\textit{confidence}}: 2 
\textbf{\textit{address}}: University Drive, Duluth, MN, USA  
\textbf{\textit{coordinates}}: 46.8185, -92.0844 \\
\hline

\raisebox{-\totalheight}{\includegraphics[width=0.15\textwidth]{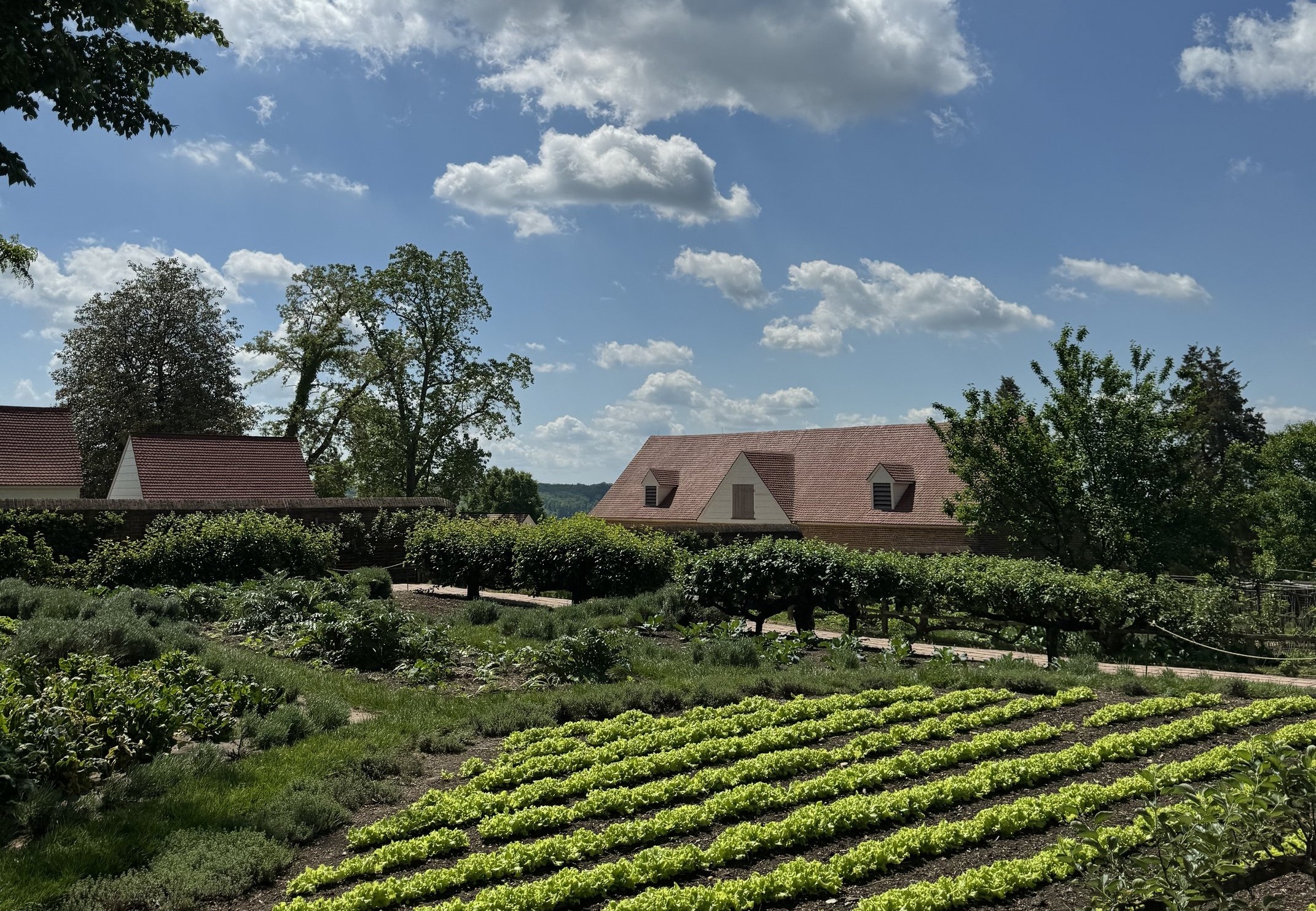}} & \textbf{Address}: Mount Vernon, USA \textbf{Coordinates}: 38.7082, -77.087 & \textbf{Model: o3}

\textbf{\textit{environment}}: outdoor  
\textbf{\textit{scene\_type}}: mixed  
\textbf{\textit{setting}}: rural   

\textbf{\textit{landmark\_recognition}}: Formal, walled kitchen garden (potager) with espaliered fruit trees and neat vegetable beds; small ancillary buildings with steep red-tile roofs and dormer windows suggest 17th-18th-century French estate outbuildings   
\textbf{\textit{text\_and\_signage}}: No readable signage or text visible   
\textbf{\textit{cultural\_indicators}}: Clay-tile roofing, cream masonry walls, symmetrical dormers and the very structured ‘potager’ layout are typical of historic French châteaux estates, especially the Potager du Roi built for Louis XIV near Versailles   
\textbf{\textit{spatial\_context}}: Garden enclosed by low walls, outbuildings just inside, wide sky and distant tree line hint at location on slightly elevated plateau much like the Royal Kitchen Garden site south of the Palace of Versailles  

\textbf{\textit{confidence}}: 2  
\textbf{\textit{address}}: 10 Rue du Maréchal Joffre, Versailles, ÎLE-DE-FRANCE, France 
\textbf{\textit{coordinates}}: 48.8015, 2.133 \\
\hline

\raisebox{-\totalheight}{\includegraphics[width=0.15\textwidth]{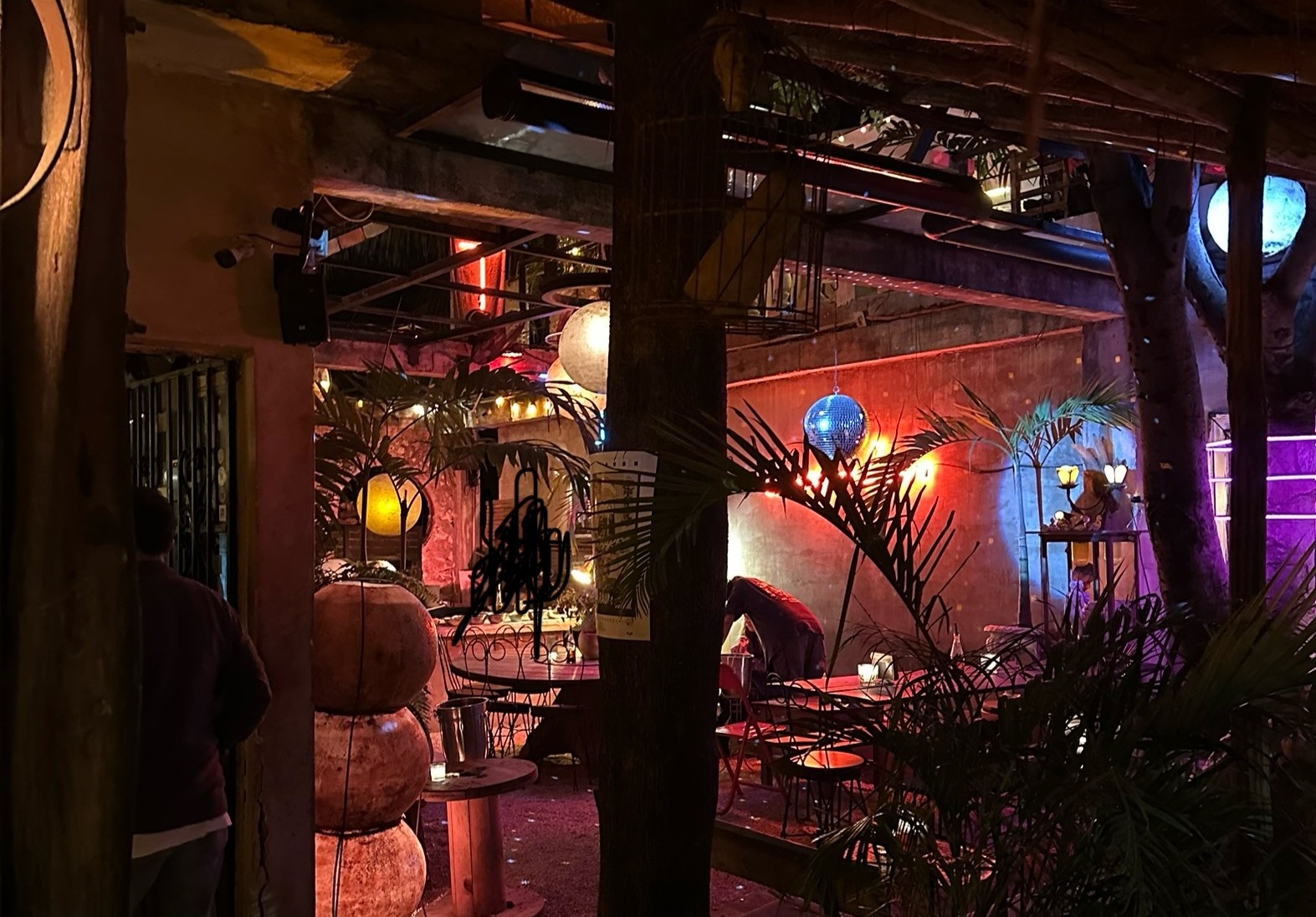}} & \textbf{Address}: Tulum, Mexico \textbf{Coordinates}: 20.2029, -87.4734 & \textbf{Model: Gemini-2.5-pro}

\textbf{\textit{environment}}: outdoor 
\textbf{\textit{scene\_type}}: building  
\textbf{\textit{setting}}: urban   

\textbf{\textit{landmark\_recognition}}: The location is identified as GITANO Garden of Love in New York City. Distinctive features include a thatched roof over a patio, real tree trunks used as support columns, a prominent disco ball, a gravel floor, and an overall 'jungle chic' or 'Tulum-inspired' aesthetic.   
\textbf{\textit{text\_and\_signage}}: No legible text or signage is clearly visible that aids in identification. A poster on a column is intentionally obscured.   
\textbf{\textit{cultural\_indicators}}: The decor is highly characteristic of the Gitano brand, which blends a bohemian, jungle-like atmosphere (often associated with Tulum, Mexico) with urban nightlife elements like disco balls and moody lighting. This specific style is a strong cultural indicator for this particular chain of venues.   
\textbf{\textit{spatial\_context}}: The scene depicts a trendy outdoor bar/restaurant patio, a type of venue common in major urban centers. The specific combination of design elements, including the layout of the patio, the stacked terracotta pots, and the wooden spool table, was matched via comparison with online photos of the identified venue in New York City.  

\textbf{\textit{confidence}}: 5  
\textbf{\textit{address}}: 76 Varick Street, New York, NY, USA  
\textbf{\textit{coordinates}}: 40.72234, -74.00681 \\
\hline

\raisebox{-\totalheight}{\includegraphics[width=0.15\textwidth]{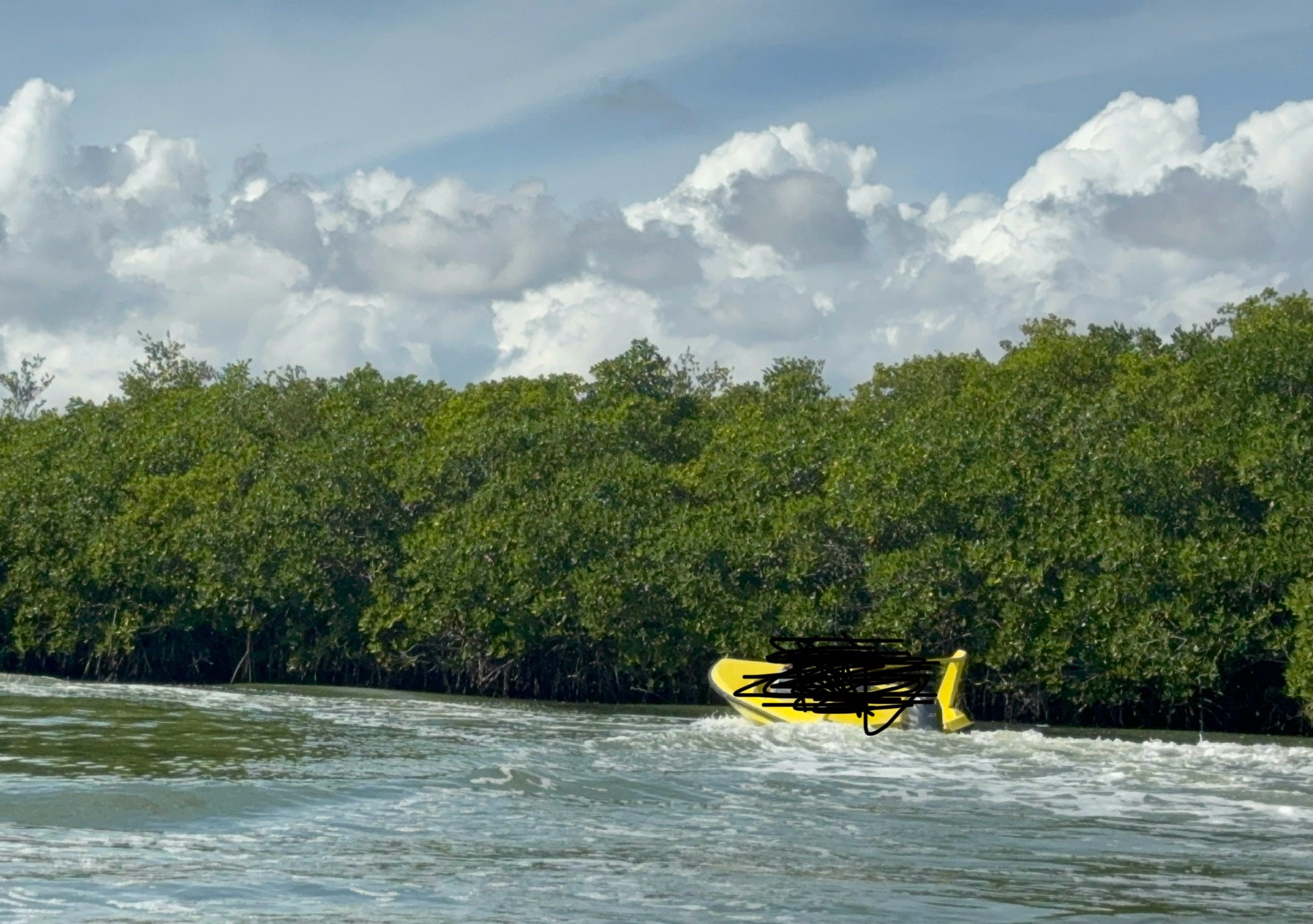}} & \textbf{Address}: Mexico \textbf{Coordinates}: 21.0466, -86.793 & \textbf{Model: Gemini-2.5-pro}

\textbf{\textit{environment}}: outdoor  
\textbf{\textit{scene\_type}}: scenery  
\textbf{\textit{setting}}: natural   

\textbf{\textit{landmark\_recognition}}: The dominant natural feature is a dense mangrove forest, characterized by trees with visible prop roots growing directly at the edge of a saltwater or brackish waterway. This type of ecosystem is specific to tropical and subtropical coastlines.  
\textbf{\textit{text\_and\_signage}}: No text or identifiable signage is visible in the image.   
\textbf{\textit{cultural\_indicators}}: The presence of a recreational watercraft (partially obscured yellow boat, possibly a jet ski or small motorboat) combined with the mangrove environment is very common in the coastal regions of Florida, which has extensive mangrove forests and a large boating culture.   
\textbf{\textit{spatial\_context}}: The image depicts a coastal inlet or intracoastal waterway. The choppy water and the wake from a boat suggest an active channel used for recreation. This scenery is highly representative of the Intracoastal Waterway in South Florida, particularly in areas like Fort Lauderdale or Miami.  

\textbf{\textit{confidence}}: 3 
\textbf{\textit{address}}: Intracoastal Waterway, Fort Lauderdale, FL, USA  
\textbf{\textit{coordinates}}: 26.099167, -80.116389 \\
\hline

\end{tabular}
}
\label{tab:sample_pcw}
\end{table*}

\end{document}